\documentclass[sigconf]{acmart}

\AtBeginDocument{%
  \providecommand\BibTeX{{%
    \normalfont B\kern-0.5em{\scshape i\kern-0.25em b}\kern-0.8em\TeX}}}

\copyrightyear{2023} 
\acmYear{2023} 
\setcopyright{acmlicensed}\acmConference[WWW '23 Companion]{Companion
Proceedings of the ACM Web Conference 2023}{April 30-May 4, 2023}{Austin,
TX, USA}
\acmBooktitle{Companion Proceedings of the ACM Web Conference 2023 (WWW
'23 Companion), April 30-May 4, 2023, Austin, TX, USA}
\acmPrice{15.00}
\acmDOI{10.1145/3543873.3587623}
\acmISBN{978-1-4503-9419-2/23/04}

\usepackage{mwe}
\usepackage{graphbox} 

\begin{document}

\title{Psychotherapy AI Companion with Reinforcement Learning Recommendations and Interpretable Policy Dynamics}

\author{Baihan Lin}
\email{baihan.lin@columbia.edu}
\authornote{Corresponding author: Baihan Lin (baihan.lin@columbia.edu)}
\orcid{0000-0002-7979-5509}
\affiliation{%
  \institution{Columbia University}
  \streetaddress{}
  \city{New York}
  \state{NY}
  \country{USA}
  \postcode{10027}
}

\author{Guillermo Cecchi}
\email{gcecchi@us.ibm.com}
\affiliation{%
  \institution{IBM TJ Watson Research Center}
  \streetaddress{}
  \city{Yorktown Heights}
\state{NY}
  \country{USA}
  \postcode{10598}}

\author{Djallel Bouneffouf}
\email{djallel.bouneffouf@ibm.com}
\affiliation{%
  \institution{IBM TJ Watson Research Center}
  \streetaddress{}
  \city{Yorktown Heights}
\state{NY}
  \country{USA}
  \postcode{10598}}

\renewcommand{\shortauthors}{Lin et al.}

\begin{abstract}
We introduce a Reinforcement Learning Psychotherapy AI Companion that generates topic recommendations for therapists based on patient responses. The system uses Deep Reinforcement Learning (DRL) to generate multi-objective policies for four different psychiatric conditions: anxiety, depression, schizophrenia, and suicidal cases. We present our experimental results on the accuracy of recommended topics using three different scales of working alliance ratings: task, bond, and goal. We show that the system is able to capture the real data (historical topics discussed by the therapists) relatively well, and that the best performing models vary by disorder and rating scale. To gain interpretable insights into the learned policies, we visualize policy trajectories in a 2D principal component analysis space and transition matrices. These visualizations reveal distinct patterns in the policies trained with different reward signals and trained on different clinical diagnoses. Our system's success in generating DIsorder-Specific Multi-Objective Policies (DISMOP) and interpretable policy dynamics demonstrates the potential of DRL in providing personalized and efficient therapeutic recommendations.
\end{abstract}



\begin{teaserfigure}
  \includegraphics[width=\textwidth]{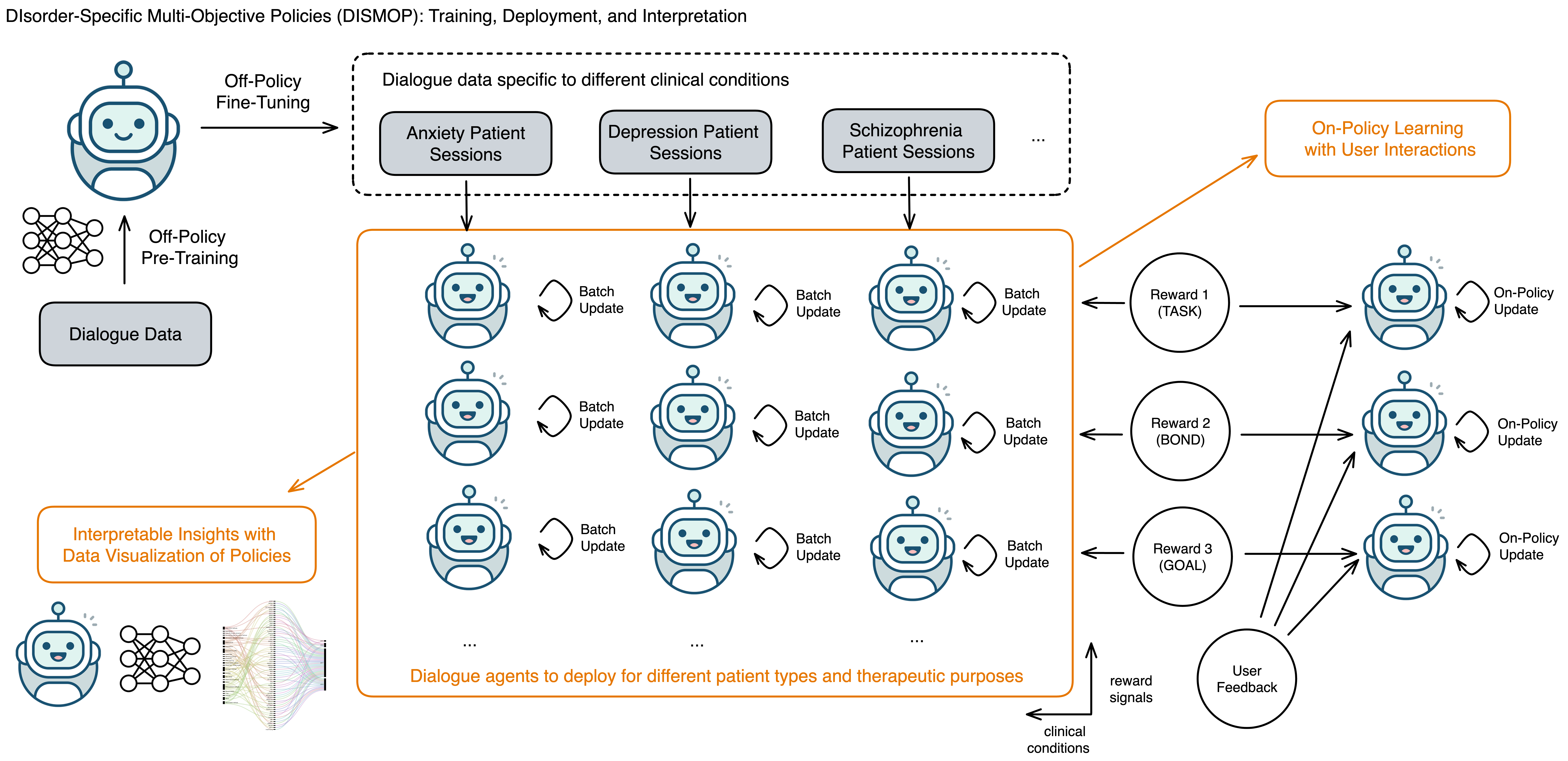}
  \caption{Reinforcement Learning Psychotherapy AI Companion with Disorder-Specific Multi-Objective Policies (DISMOP)}
  \label{fig:dismop}
\end{teaserfigure}


\maketitle

\begin{figure*}[tb]
\centering
    \includegraphics[width=\linewidth]{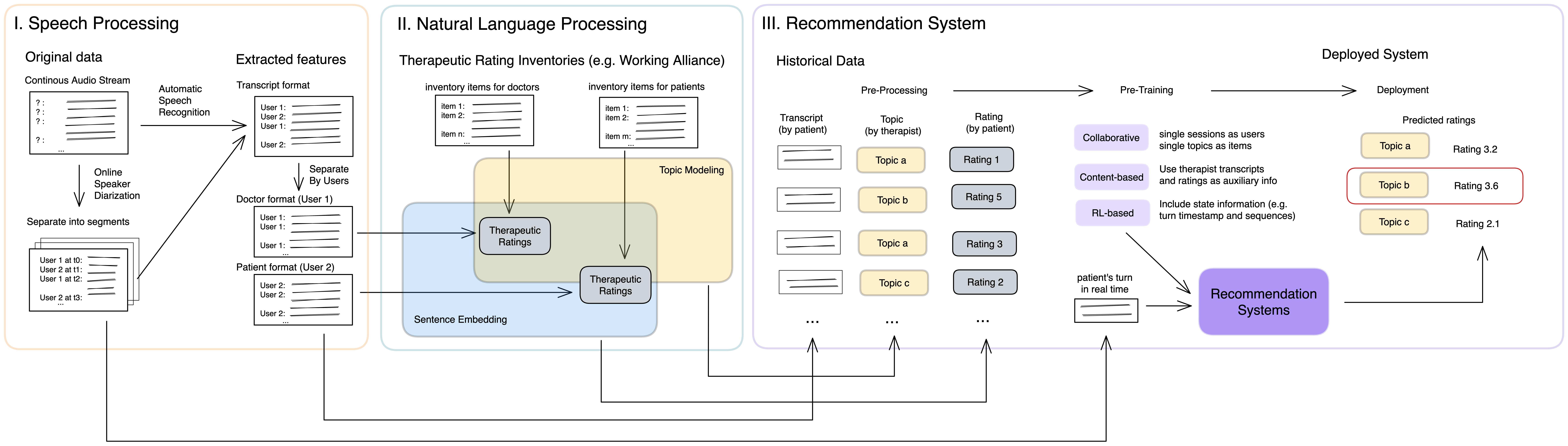} 
\par\caption{The speech, NLP and recommendation system components of the Psychotherapy AI Companion.
}\label{fig:pipeline1}
\end{figure*}

\section{Introduction}

Mental illness is a severe healthcare problem that affects a significant proportion of the global population \cite{patel2018lancet}. 
Despite the high prevalence of mental health conditions, many countries, including the US, face a severe shortage of mental health practitioners, such as psychiatrists and clinical psychologists \cite{satiani2018projected}. The COVID-19 pandemic has further amplified this demand gap by taking a significant toll on people's mental health \cite{wang2020investigating}.

The current education and training systems for mental health practitioners cannot keep up with this trend because each licensed therapist requires years of continual learning and supervised training. Even when a therapist is ripe for independent practice, many still seek weekly supervision from more senior therapists who serve as crucial mentors in navigating the difficulties that arise during psychotherapy training \cite{watkins2013being}.

To address this gap, we propose a virtual psychotherapy AI companion that provides real-time feedback and recommends treatment strategies to therapists while they are conducting psychotherapy. Like a supervisor, our AI companion offers feedback and guidance that are case-dependent and has learned from thousands of historical therapy sessions and case studies.

The base of our recommendation system relies on a rating system that evaluates the effectiveness of a treatment strategy. As characterizing a patient's mental state can be complicated, we focus our approach on well-defined clinical outcomes. One such outcome is the working alliance, a psychological concept that is highly predictive of the success of psychotherapy in a clinical setting \cite{Wampold2015}. It describes important cognitive and emotional components of the relationship between the therapist and patient, including the agreement on goals and tasks and the establishment of a bond, trust, and respect over the course of the dialogue \cite{Bordin79}. In our previous work \cite{lin2022deep}, we developed a natural language processing (NLP) approach to infer this quantity as real-time ratings of the therapist's treatment progress within the patient's entire program.

We also proposed the Reinforced Recommendation model for Dialogue topics in psychiatric Disorders (R2D2) \cite{lin2022supervisor,lin2023help}, which is the first-ever recommendation system of dialogue topics proposed for the psychotherapy setting. R2D2 transcribes the session in real-time, predicts the therapeutic outcome as a turn-level rating, and recommends a treatment strategy that is best suited for the current context and state of the psychotherapy. This framework is a critical step towards addressing the global issue of mental health by augmenting the treatment and education of clinical practitioners with a recommendation system of therapeutic strategies.

In this work, we extend this framework to offer interpretable insights by visualizing the policies fine-tuned for different clinical conditions and therapeutic emphases. We propose the Disorder-Specific Multi-Objective Policies (DISMOP), as shown in Figure \ref{fig:dismop}. These reinforcement learning agents can be pre-trained in a large text corpus of dialogues and then fine-tuned with off-policy training given historical data of psychotherapy sessions for patients with different clinical conditions. In addition to learning from these historical state transitions, we can compute post hoc reward signals using the computational inference defined in \cite{lin2022deep,lin2022unsupervised}, such that each policy has both a disorder specificity and a therapeutic emphasis (e.g., to achieve better bonding between the therapist and the patient). These fine-tuned policies can then be inspected for interpretable insights with policy visualizations and other types of explainable AI approaches or deployed as a psychotherapy AI companion. These autonomous agents can be adaptively updated given certain therapeutic rewards, such as the computational inferred working alliance scores in task, bond, or goal scales, as in \cite{lin2022deep,lin2022unsupervised}, or updated given implicit or explicit user feedback (e.g., the therapist can rate the recommended topics given by our AI companion) or a mixture of reward signals as our on-policy training objectives.

In the following sections, we will introduce this interpretable reinforcement learning AI therapy companion framework and provide empirical evaluations and policy visualizations. We will also discuss the ethical considerations and future steps. The contributions of this work are twofold: first, we propose an interpretable AI framework that can recommend treatment strategies to therapists during psychotherapy sessions. Second, we provide insights into the policies fine-tuned for different clinical conditions and therapeutic emphases, which can help therapists gain a better understanding of the treatment process and improve patient outcomes. Our framework has the potential to significantly impact mental healthcare by addressing the shortage of mental health practitioners and improving the quality and accessibility of psychotherapy.

\section{Methods}

In this section, we outline our analytic framework (Figure \ref{fig:pipeline1}). The framework includes speaker diarization, therapeutic quality rating, transcription and real-time rating assessment, topic modeling as recommendation items, recommendation system settings, deep reinforcement learning recommendation approaches, disorder-specific multi-objective policies (DISMOP), and the three levels of our Psychotherapy AI Companion.

\subsection{Speech Processing and Transcription}

The continuous audio stream is fed into the system, and we perform speaker diarization using real-time solutions such as \cite{lin2021speaker,lin2020voiceid,lin2020speaker,lin2023rldiarization}. This process separates audio into dyads of doctor-patient, which are then transcribed into linguistic turns for downstream analyses.

\subsection{Therapeutic Quality Ratings}

After obtaining a well diarization result, we configure the quality assessment setting by specifying a proper inventory. In this system, we use the Working Alliance Inventory (WAI) \cite{horvath1981exploratory,tracey1989factor,martin2000relation}, a set of self-report measurement questionnaires that quantifies the therapeutic bond, task agreement, and goal agreement. The goal is to derive three alliance scales: the task scale, the bond scale, and the goal scale, which measure the three major themes of psychotherapy outcomes. The score corresponding to the three scales comes from a key table that specifies the positivity or the sign weight to be applied to the questionnaire answer when summing in the end.

\subsection{Real-time Rating Assessment}

After we transcribe the diarized audio stream with a standard automatic speech recognition module \cite{adorf2013web}, we embed both the dialogue turns and WAI items with deep sentence or paragraph embeddings (in this case, Doc2Vec \cite{le2014distributed}), following the approach proposed in \cite{lin2022deep,lin2022working,lin2022unsupervised,lin2022voice}. We then compute the cosine similarity between the embedding vectors of the turn and its corresponding inventory vectors. With that, we obtain a 36-dimension working alliance score for each turn (either by patient or by therapist), which we may save in a bidirectional relational database as in \cite{lin2022knowledge}, or visualize in real-time as a conversation guide \cite{lin2022voice}.

\subsection{Topic Modeling as Recommendation Items}

In our recommendation system, the items the system recommends are treatment strategies, represented as topics that the therapist should initiate or continue for the next turn. Given a large text corpus of many psychotherapy sessions, we can first perform topic modeling to extract the main concepts discussed in the psychotherapy \cite{lin2022neural}, which can also be directly visualized for interpretable insights \cite{lin2023therapyview}. We use the Embedded Topic Model (ETM) \cite{dieng2020topic}, which was shown to create the most diverse concepts in psychological corpus as in this systematic analysis \cite{lin2022neural}. In this study, we annotate each turn with their most likely topic and identify seven unique topics: Topic 0 is about figuring out, self-discovery and reminiscence; Topic 1 is about play; Topic 2 is about anger, scare and sadness; Topic 3 is about counts; Topic 6 is about explicit ways to deal with stress, such as keeping busy and reaching out for help; Topic 7 is about numbers, and Topic 8 is about continuation.

\subsection{Recommendation System Setting}

In this subsection, we describe how we set up the recommendation system for psychotherapy treatment strategies. We define the ``items'', ``users'', ``contents'' and ``ratings'' in our recommendation system. In our case, the ``items'' the system recommends are treatment strategies, which we represent as a topic that the therapist should initiate or continue for the next turn.

We pair these ``items'' with the ``users'' and ``contents'', which, in our case, would be the patientID, their previous turns, their aggregated formats, and other metadata. For instance, we know that within each session, there exist many pairs of turns, and they would belong to the same user''. However, one can also assign all turns belonging to one clinical label or all turns related to a certain topic as one ``user''. In this example, we choose the session IDs as users. Lastly, the ``ratings'' would be patients' inferred alliance scores predictive of the therapeutic outcomes.

Creating this database from historical data, we can train our system. Since we have defined our users, items, contents, and ratings, the recommendation engine can be easily crafted with content-based \cite{pazzani2007content,basu1998recommendation,aggarwal2016recommender} and collaborative filtering \cite{sarwar2001item,he2017neural,koren2022advances,su2009survey} methods. Since our session turns are sequential and can specify a state or timestamp, it might be suitable for RL \cite{zheng2018drn,wang2014exploration,zou2020pseudo} and session-based methods \cite{li2017neural,wu2019session,ludewig2018evaluation}, which can provide further interpretable clinical insights.

During deployment, our system registers our session as a new ``user'' if we adopt a session-based item, providing punctuated rater evaluations as inference anchors \cite{lin2022supervisor,lin2023help}. Next steps include predicting these inference anchors as states (like \cite{lin2022neural,lin2022predicting,lin2022predicting2}) and training chatbots as reinforcement learning agents given these states and neuroscience inspirations \cite{lin2020story,lin2021models,lin2020unified}.

\subsection{Deep RL Recommendation Approaches}

Reinforcement learning approaches have been effectively applied in language and speech tasks \cite{lin2022rl4lang,lin2022voice,lin2022ispeak}, including recommendation systems. Here we evaluate three popular deep RL algorithms: Deep Deterministic Policy Gradients (DDPG) \cite{lillicrap2015continuous}, Twin Delayed DDPG (TD3) \cite{fujimoto2018addressing}, and Batch Constrained Q-Learning (BCQ) \cite{fujimoto2019off}.

DDPG is a model-free algorithm for continuous action spaces based on the deterministic policy gradient in an actor-critic architecture. It is one of the first successful algorithms to learn policies end-to-end. Building upon the Double Q-Learning, TD3 is a similar solution that is proposed to correct for the overestimated value issue, and yields more competitive results in various game settings. BCQ is the first continuous control deep RL algorithm with competitive results in off-policy evaluations by restricting the agent's exploration in action space.

As the online data collection of RL models is usually time-consuming, in real-world industrial settings, these models are usually trained using previously collected data. Offline reinforcement learning methods \cite{levine2020offline} have become popular for this purpose.

\subsection{Disorder-Specific Multi-Objective Policies (DISMOP)}

To further enhance the performance of our recommendation system, we introduce the Disorder-Specific Multi-Objective Policies (DISMOP) framework, based on our previous R2D2 model \cite{lin2022supervisor,lin2023help}. DISMOP is designed to improve the generalizability of our policies across different psychiatric disorders by training on disorder-specific datasets. The framework consists of a pretraining step and an off-policy batch training process, which uses disorder-specific historical data to learn policies that maximize multiple objectives, such as the therapeutic bond, task agreement, and goal agreement. These policies can then be deployed in suitable settings and incorporate user feedback as an additional reward signal for on-policy updates and real-time improvements.

For the recommendation systems, we identify each session as a user, and the states are frames of dialogues that can be labeled with their topics in real-time and their ratings with a working alliance (WA) inference module. The reinforcement learning core, powered by deep RL, predicts the best action represented by an embedding for the items (topics). Treating the actions, i.e. predicted topics, as a continuous action space can utilize the innate structure and relationships between the topics, comparing to a discrete action space. This embedding can be pre-computed, for instance, using dimension reduction techniques to find clusters of different topics in a low-dimensional space. We use the Doc2Vec embedding of the original dialogue turns, averaged by their topic labels, such that each action (i.e. the topic ID) has an averaged representation in the sentence embedding space. This action representation can be translated into a topic label with nearest neighbor, and a given dialogue response will be selected from the historical dialogue data to continue the conversation. The reward can then be computed using the working alliance rate or other types of therapeutic signals, including user feedback from doctors or patients.

\subsection{Three Levels of Psychotherapy AI Companion}
\label{sec:threelevels}

As shown in Figure \ref{fig:pipeline1}, our recommendation systems can be extended into three levels. The first level, our backbone, is reinforcement learning-based, which considers the stateful nature of dialogue data. The flexibility of reward signals, i.e., using any rewards, pseudo-rewards, multiple rewards, hybrid rewards, or even inferred rewards, makes our policies adaptable to a versatile suite of clinical settings. 

The second level is to use additional context, as in content-based recommendation systems. This involves treating the patient turns before the current turns, or all the previous turns up to now, as a feature in our deep reinforcement learning models, by concatenating their sentence embeddings to our states. This provides more context for in-context learning of our generalized models, which can be a foundation model in future work.

In the third level, if we are given the patient ID and doctor ID, we can create personalized policies with collaborative filtering-type recommendation systems, which can potentially improve the compositionality and generalizability of our models for a wide range of populations. 

In this work, we evaluated the performance of the first level, but our models should be able to extend to the other two as long as we have access to the available meta-information. We will further investigate the full spectrum in our future work.

\begin{table}[tb]
      \caption{Turn-level test accuracy of predicted topics
      }
      \label{tab:acc} 
      \centering
      \resizebox{\linewidth}{!}{
     \begin{tabular}{l | c | c | c | c | c }
 & Anxi & Depr & Schi & Suic & All \\ \hline
 DISMOP-DDPG-TASK & \textbf{0.6406} & 0.0002 & 0.1183 & 0.0010 & 0.0232 \\ 
DISMOP-DDPG-BOND & 0.0168 & 0.0001 & 0.0382 & 0.0232 & 0.0001 \\ 
DISMOP-DDPG-GOAL & 0.0000 & 0.0010 & 0.0083 & 0.0232 & 0.0011 \\ \hline
DISMOP-TD3-TASK & 0.0232 & 0.0202 & 0.0232 & 0.0051 & 0.0000 \\ 
DISMOP-TD3-BOND & 0.0232 & 0.0001 & 0.0001 & \textbf{0.3143} & 0.0232 \\ 
DISMOP-TD3-GOAL & 0.0131 & 0.0001 & \textbf{0.3144} & 0.3138 & 0.0232 \\ \hline
DISMOP-BCQ-TASK & 0.0001 & \textbf{0.0539} & \textbf{0.3144} & 0.0232 & 0.0000 \\ 
DISMOP-BCQ-BOND & 0.0232 & 0.0000 & 0.0010 & 0.0000 & 0.0019 \\ 
DISMOP-BCQ-GOAL & 0.0232 & 0.0010 & 0.2783 & 0.0000 & \textbf{0.6424} \\ 
\end{tabular}
 }
\end{table}

\begin{table*}[tb]
      \caption{The average topics trajectories of policy trained for different disorders and therapeutic purposes (rewards)
      }
      \label{tab:traj_viz} 
      \centering
     \begin{tabular}{c  c  c  c  c  c }
 & Anxiety & Depression & Schizophrenia & Suicidal & All Sessions \\ 
DISMOP-DDPG 
& \includegraphics[align=c,width=.15\textwidth]{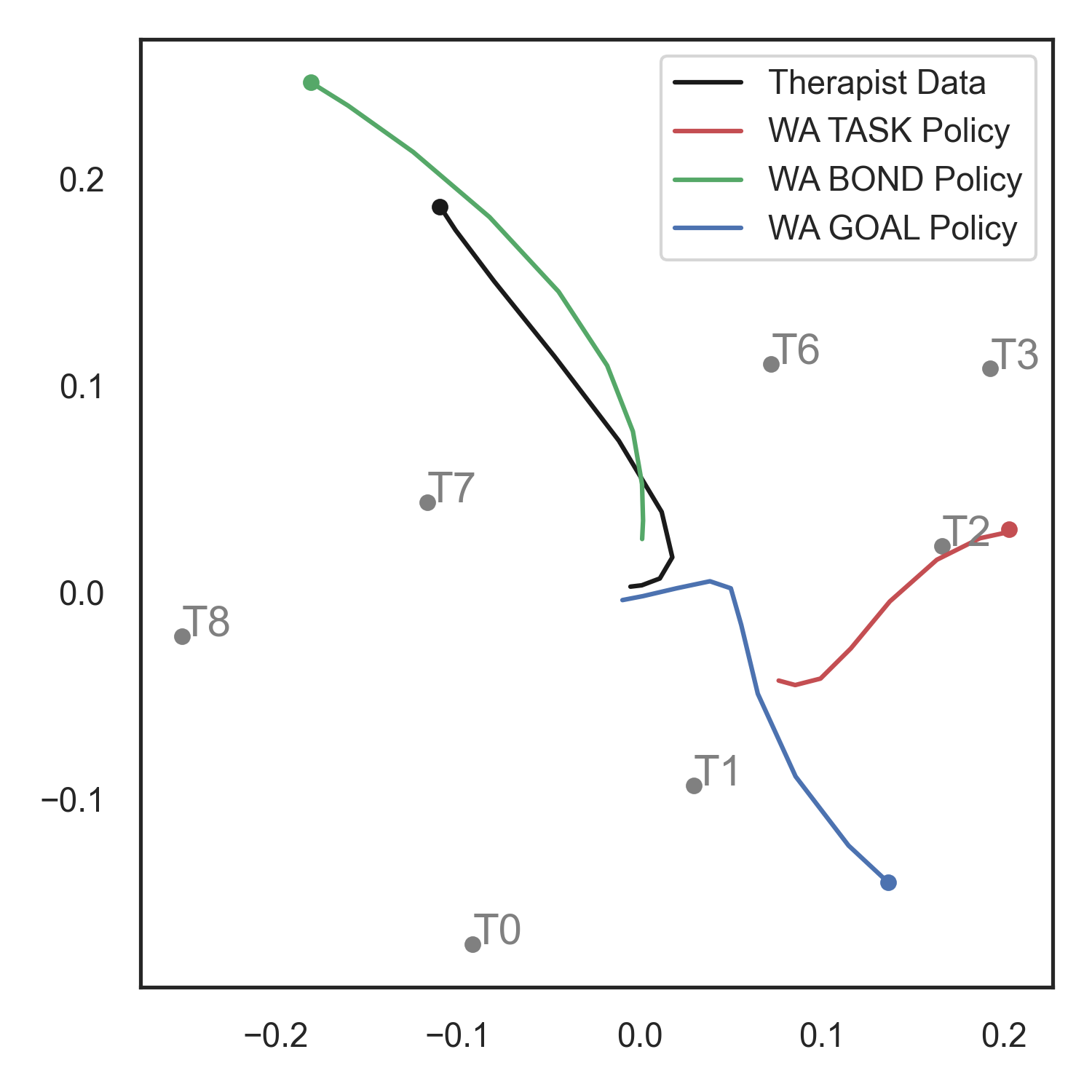}
& \includegraphics[align=c,width=.15\textwidth]{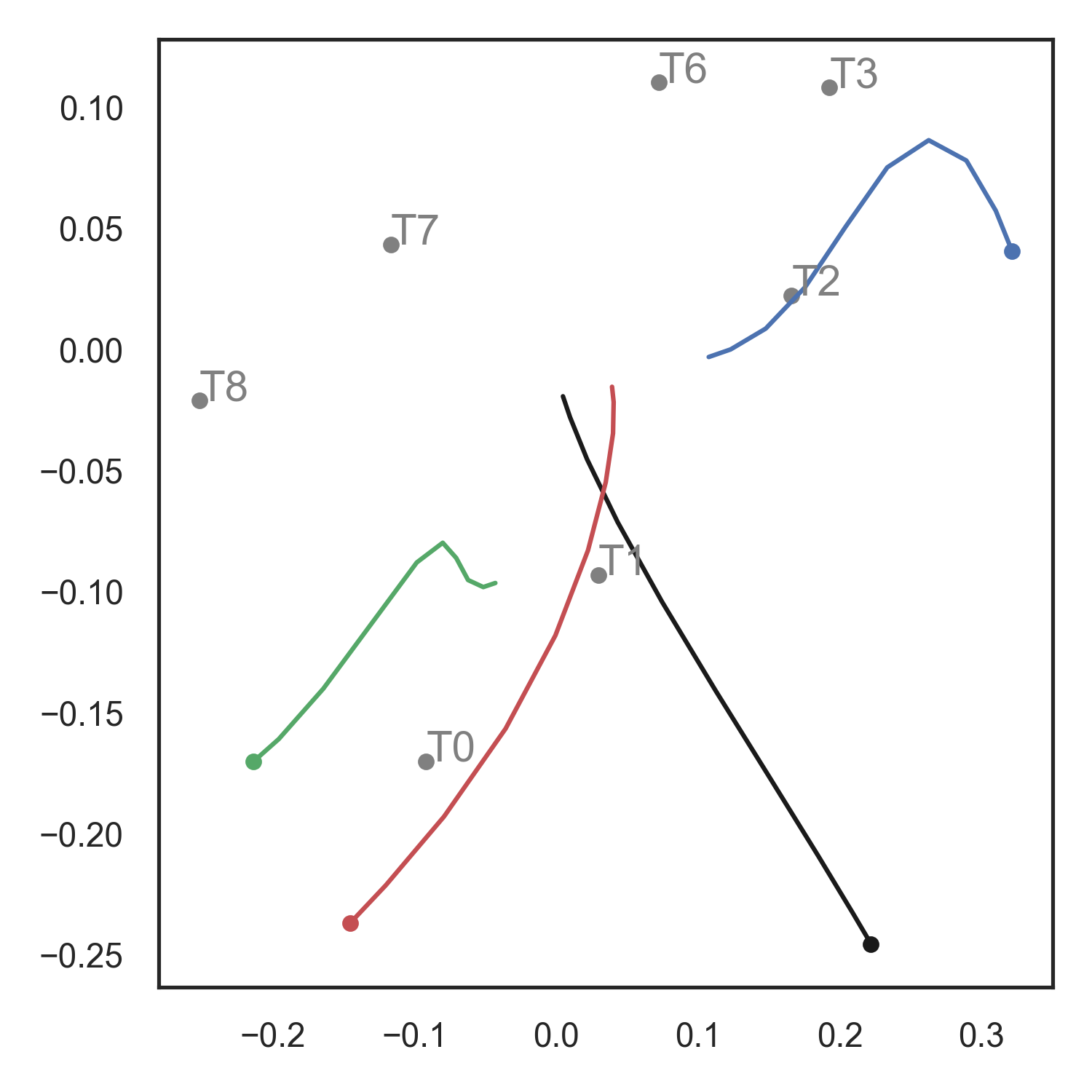}
& \includegraphics[align=c,width=.15\textwidth]{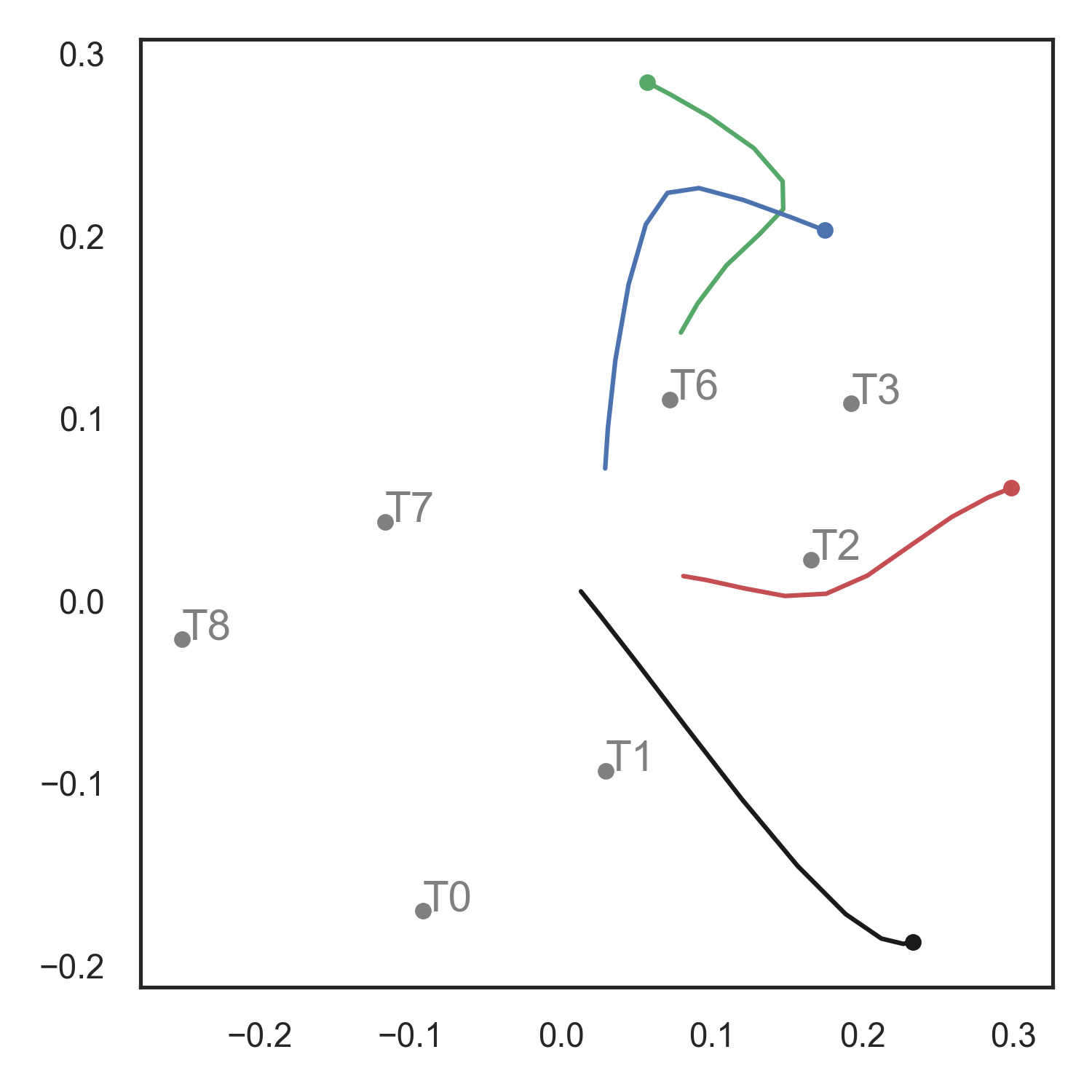}
& \includegraphics[align=c,width=.15\textwidth]{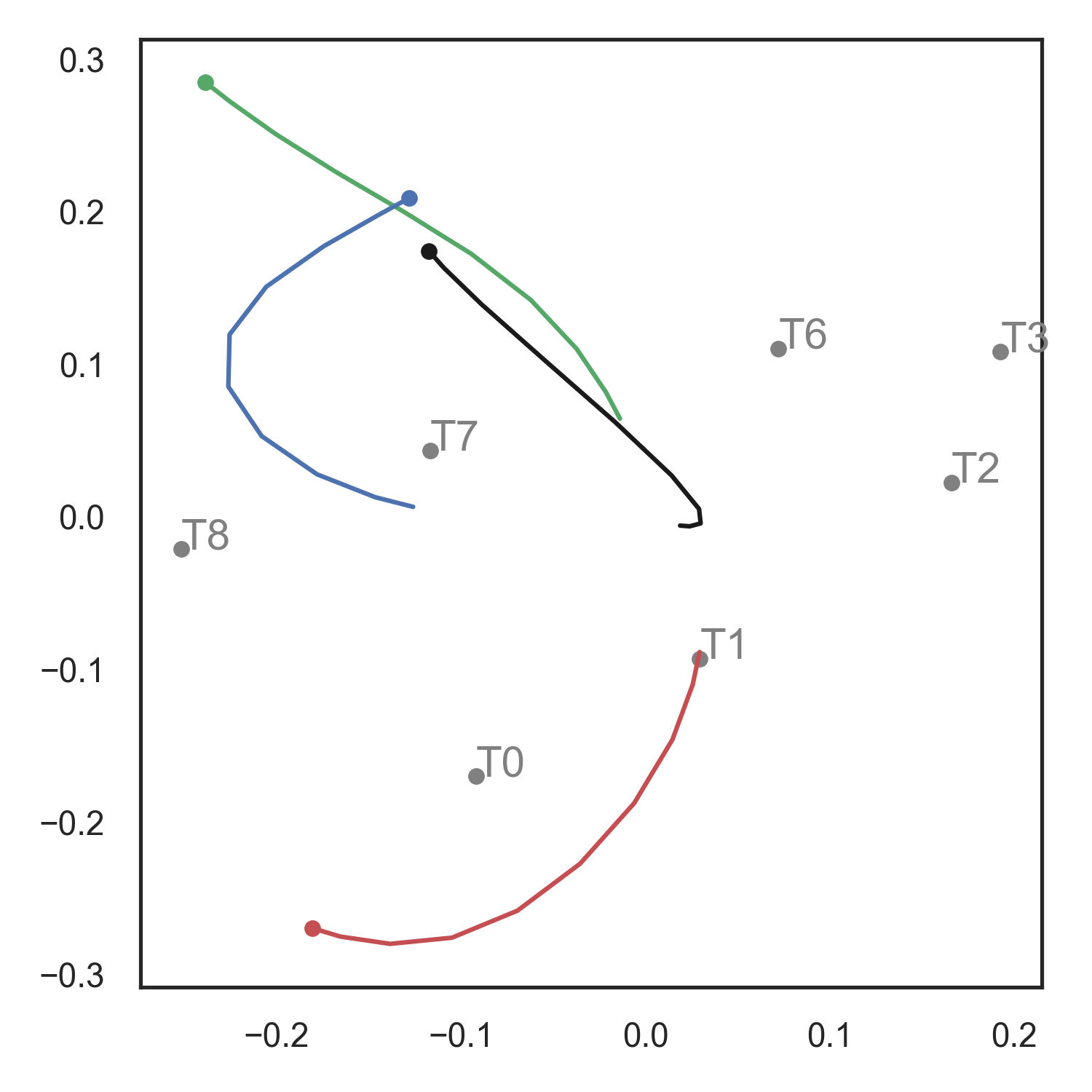}
& \includegraphics[align=c,width=.15\textwidth]{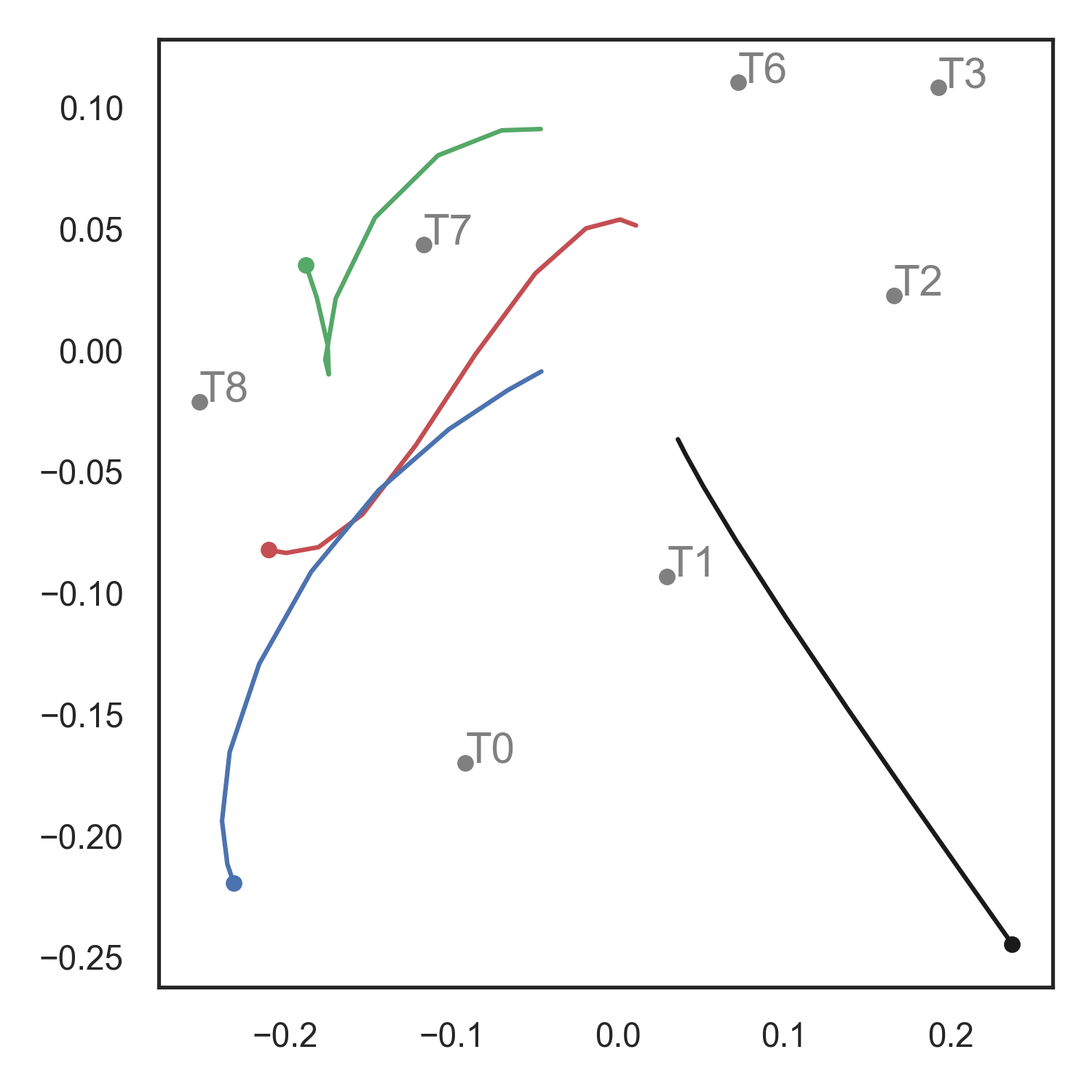}
\\ 
DISMOP-TD3   
& \includegraphics[align=c,width=.15\textwidth]{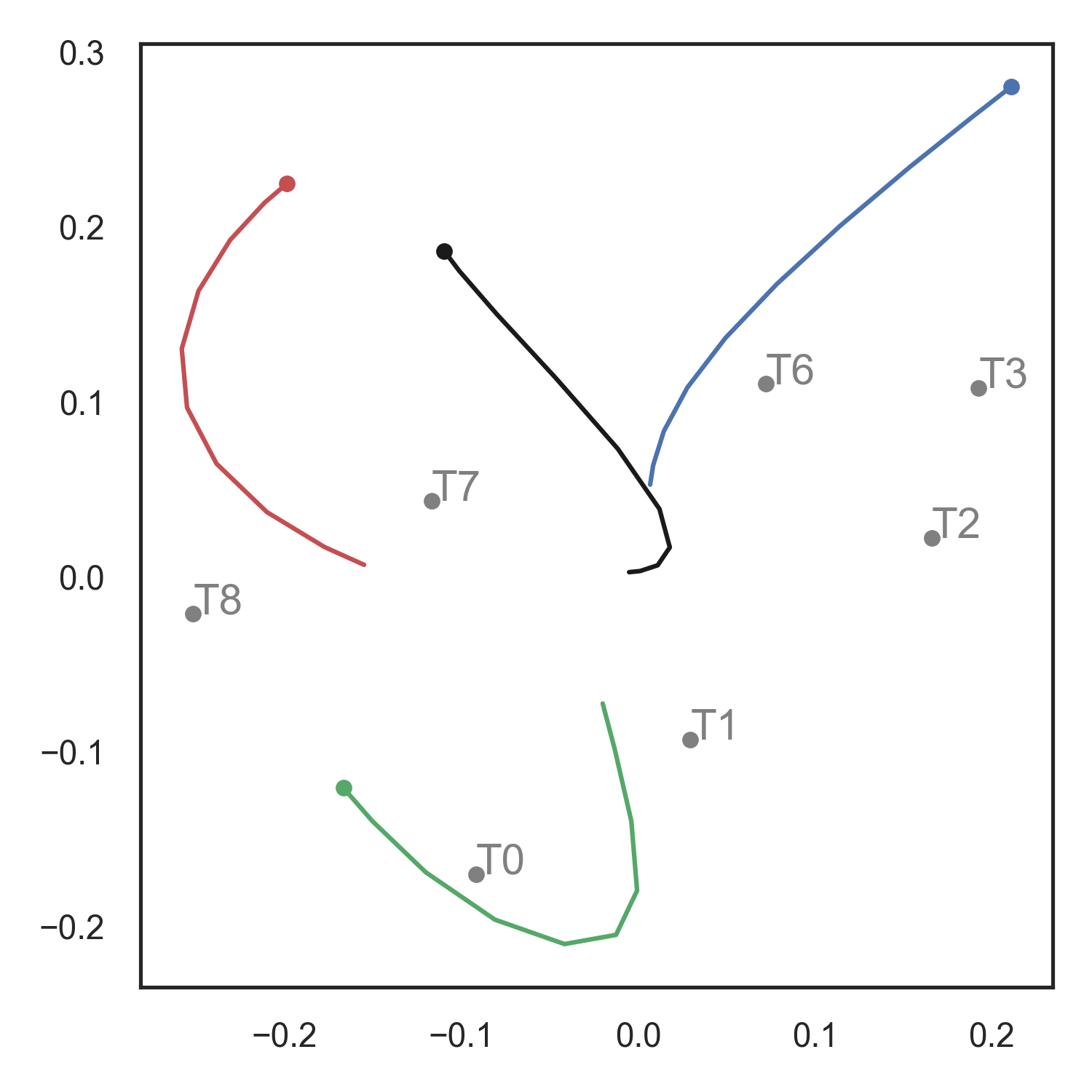}
& \includegraphics[align=c,width=.15\textwidth]{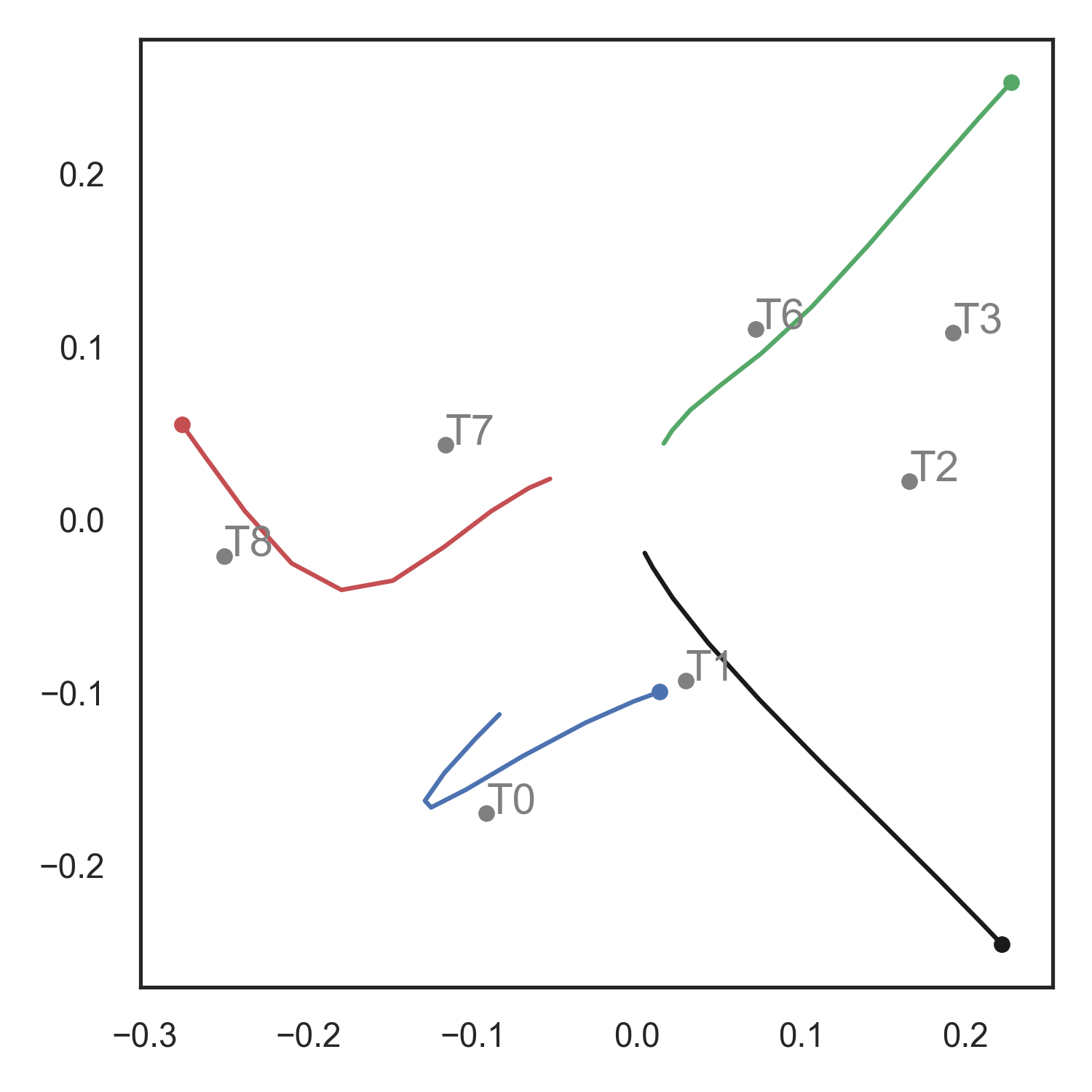}
& \includegraphics[align=c,width=.15\textwidth]{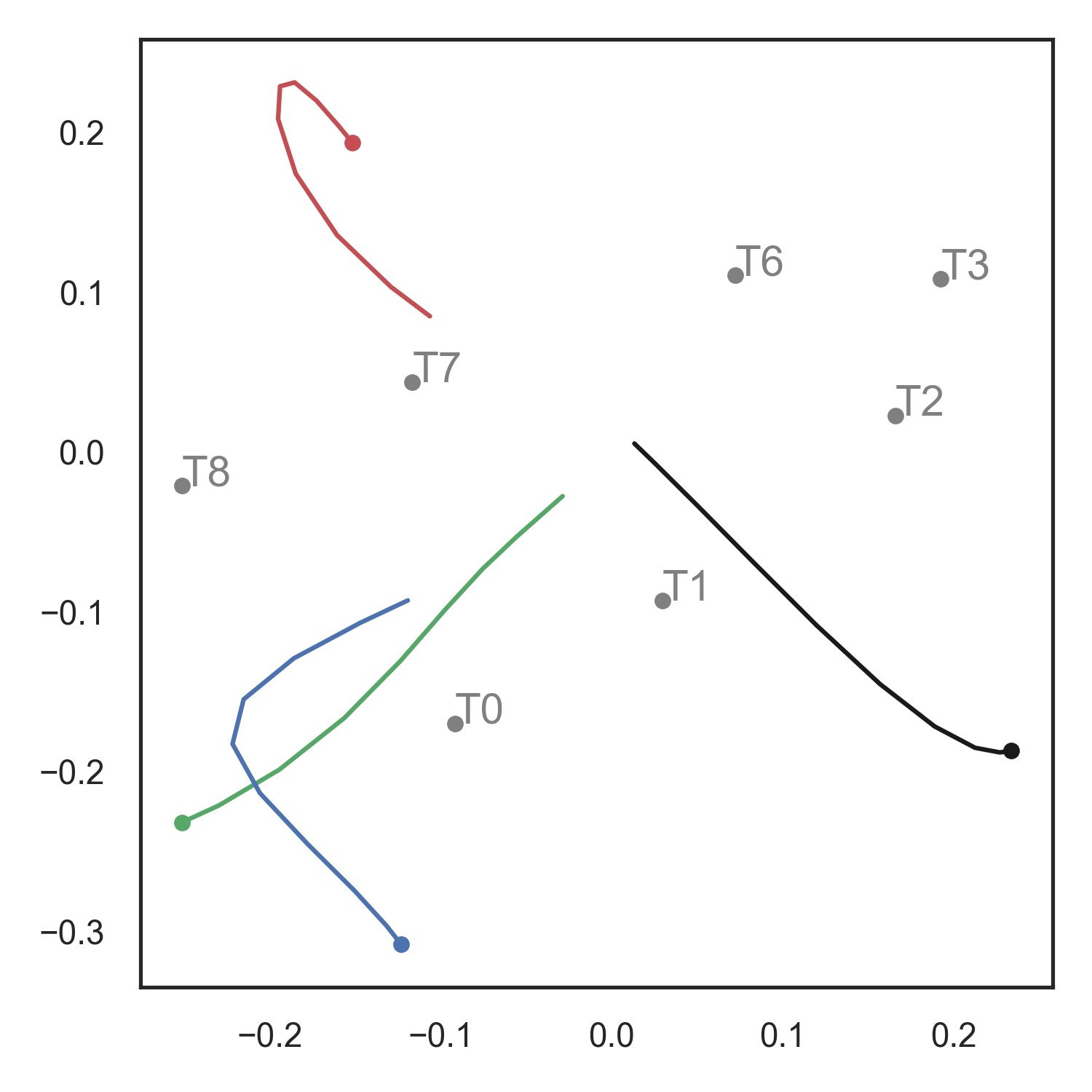}
& \includegraphics[align=c,width=.15\textwidth]{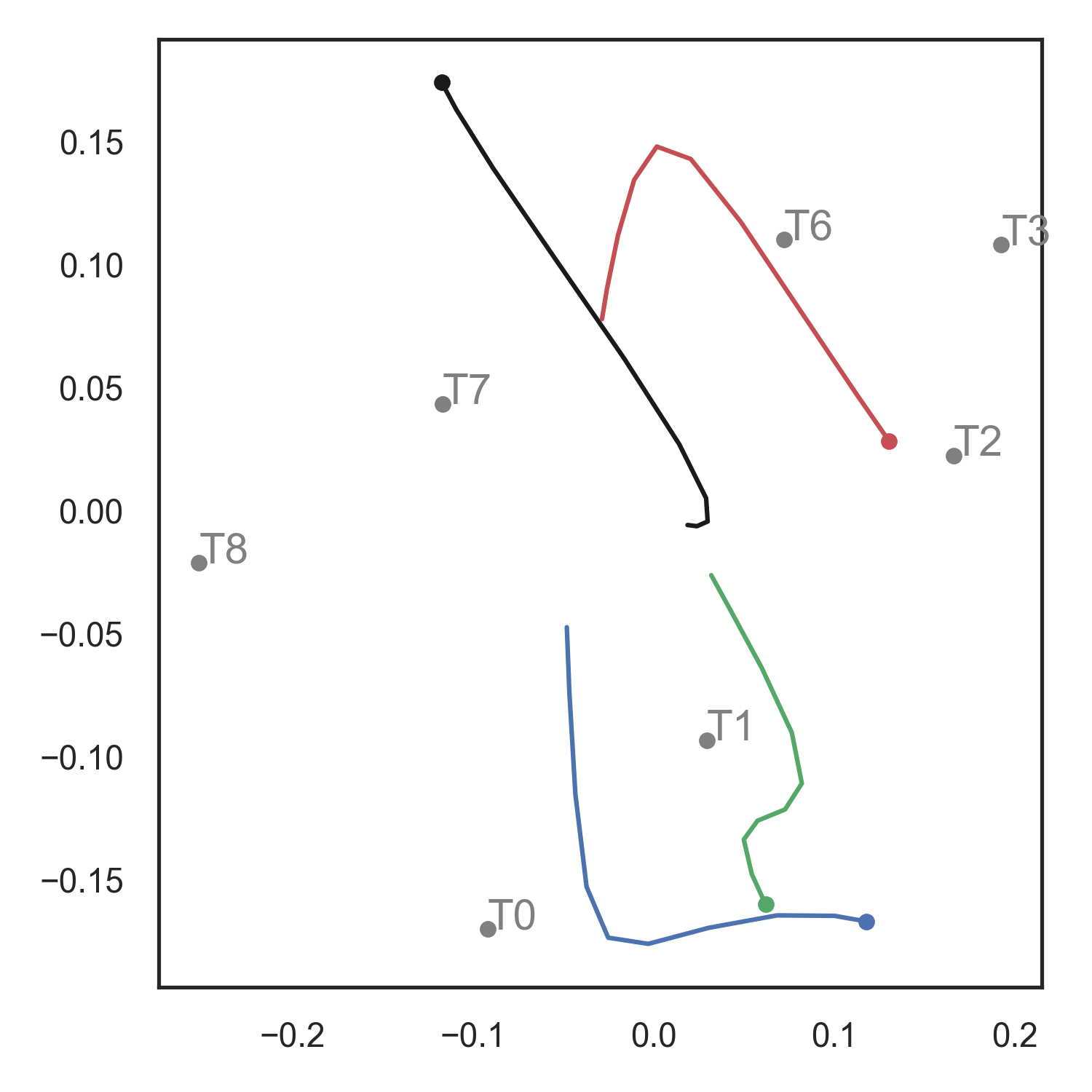}
& \includegraphics[align=c,width=.15\textwidth]{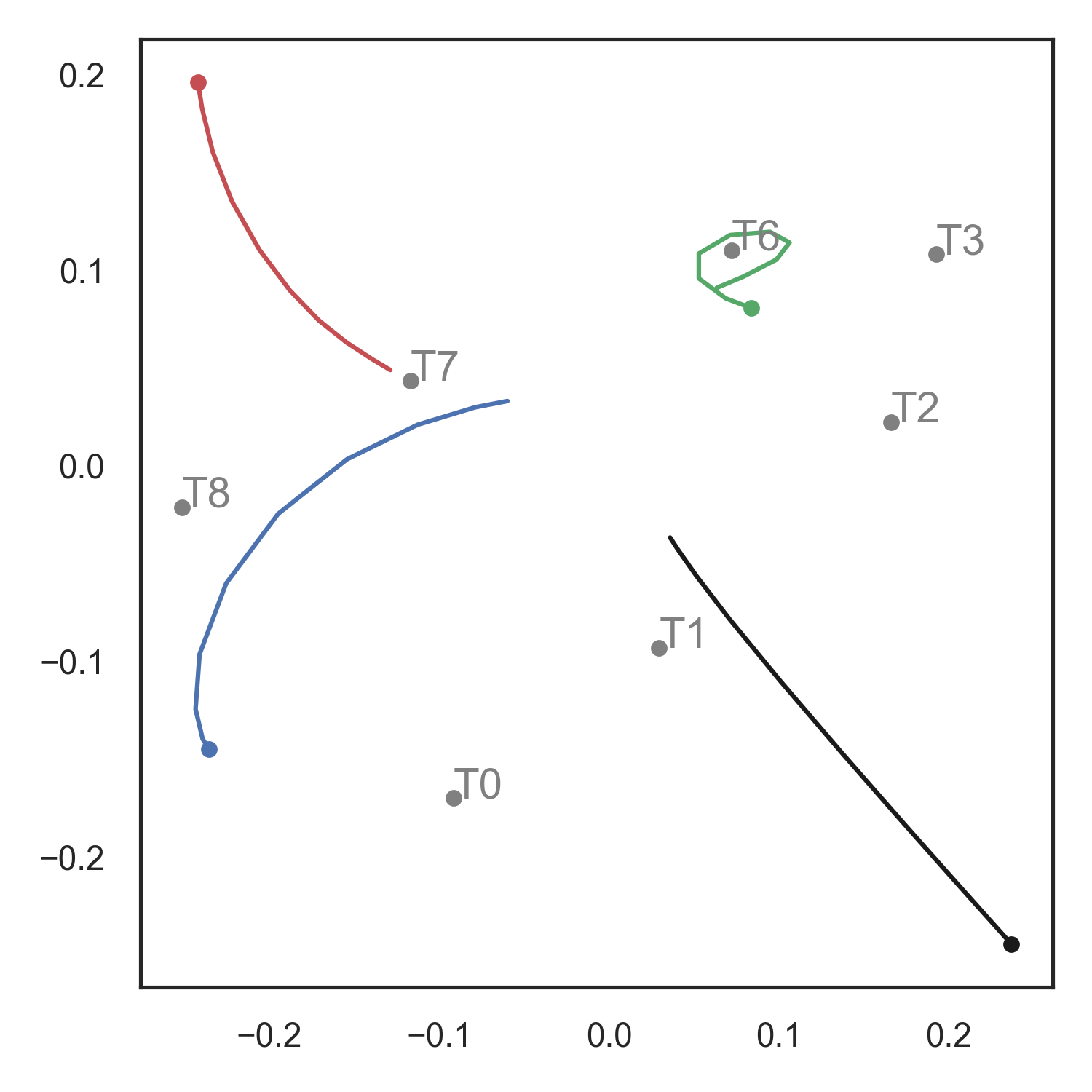}
\\ 
DISMOP-BCQ  
& \includegraphics[align=c,width=.15\textwidth]{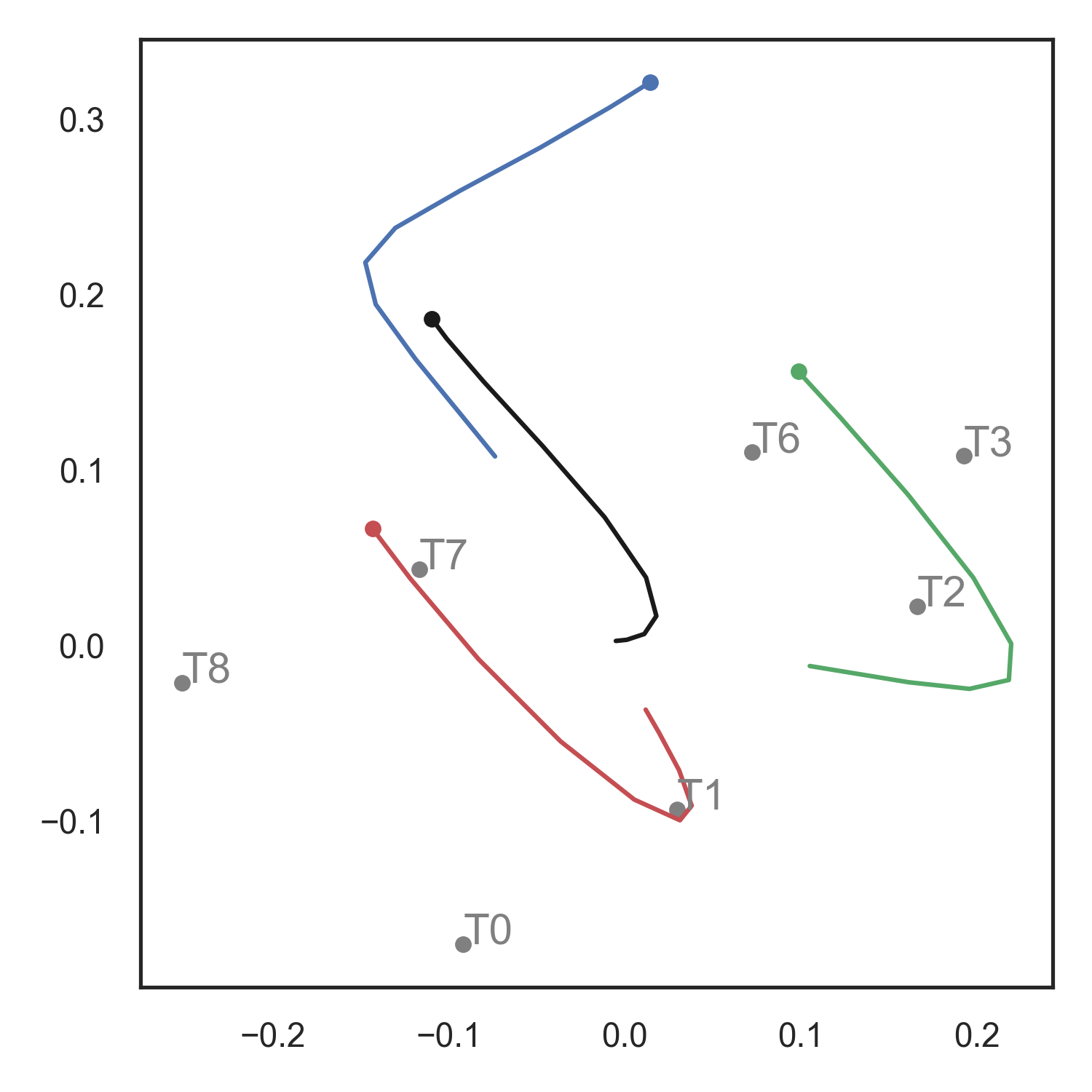}
& \includegraphics[align=c,width=.15\textwidth]{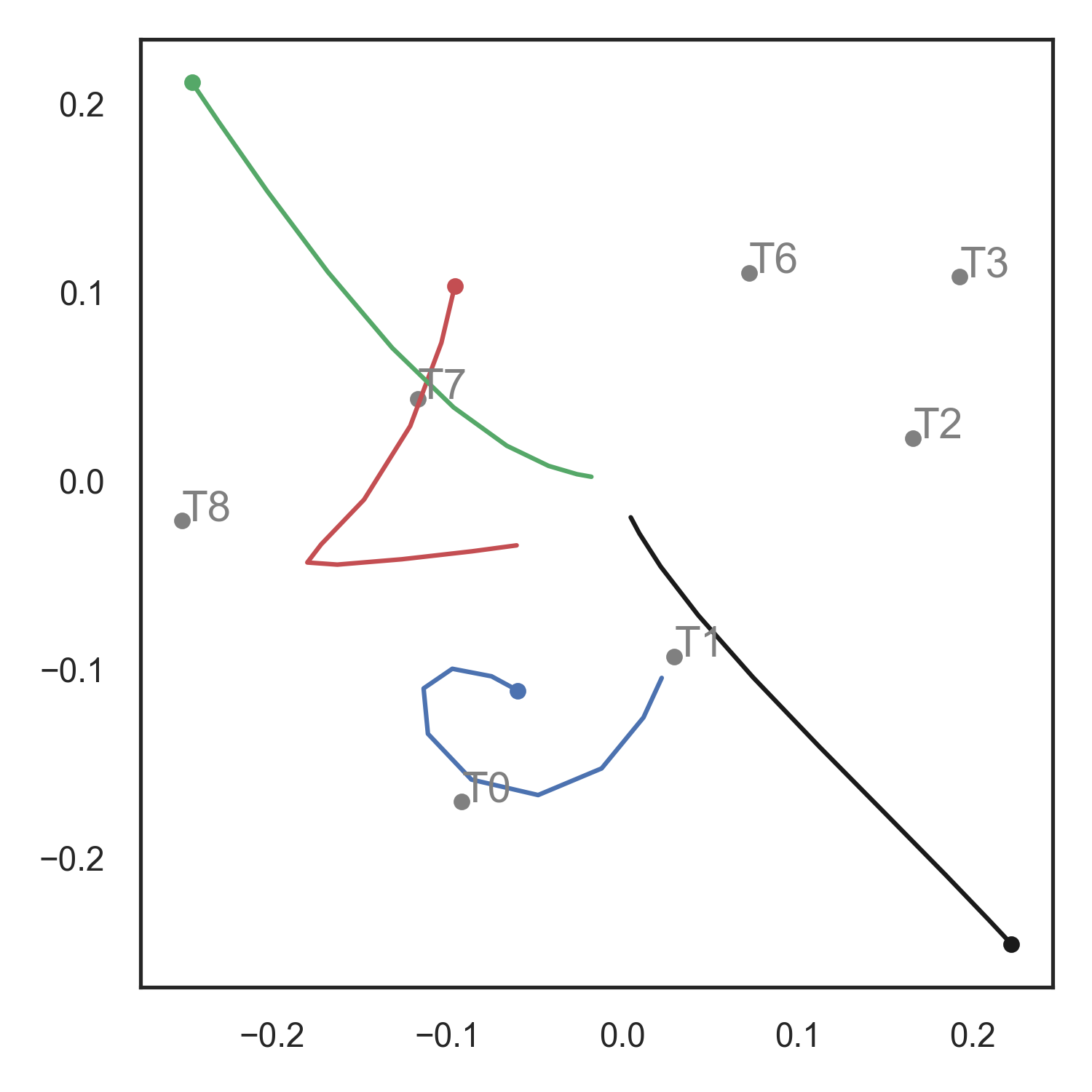}
& \includegraphics[align=c,width=.15\textwidth]{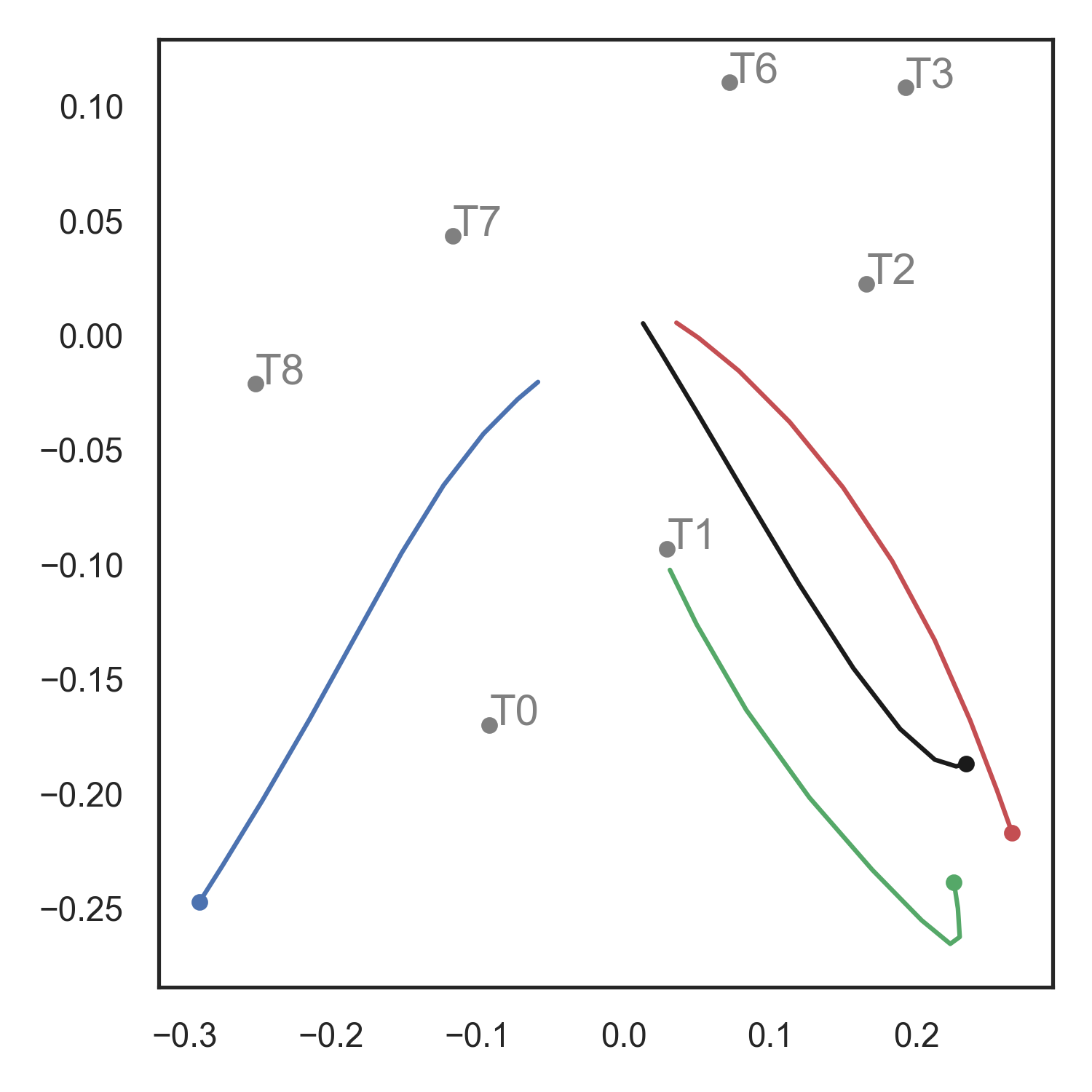}
& \includegraphics[align=c,width=.15\textwidth]{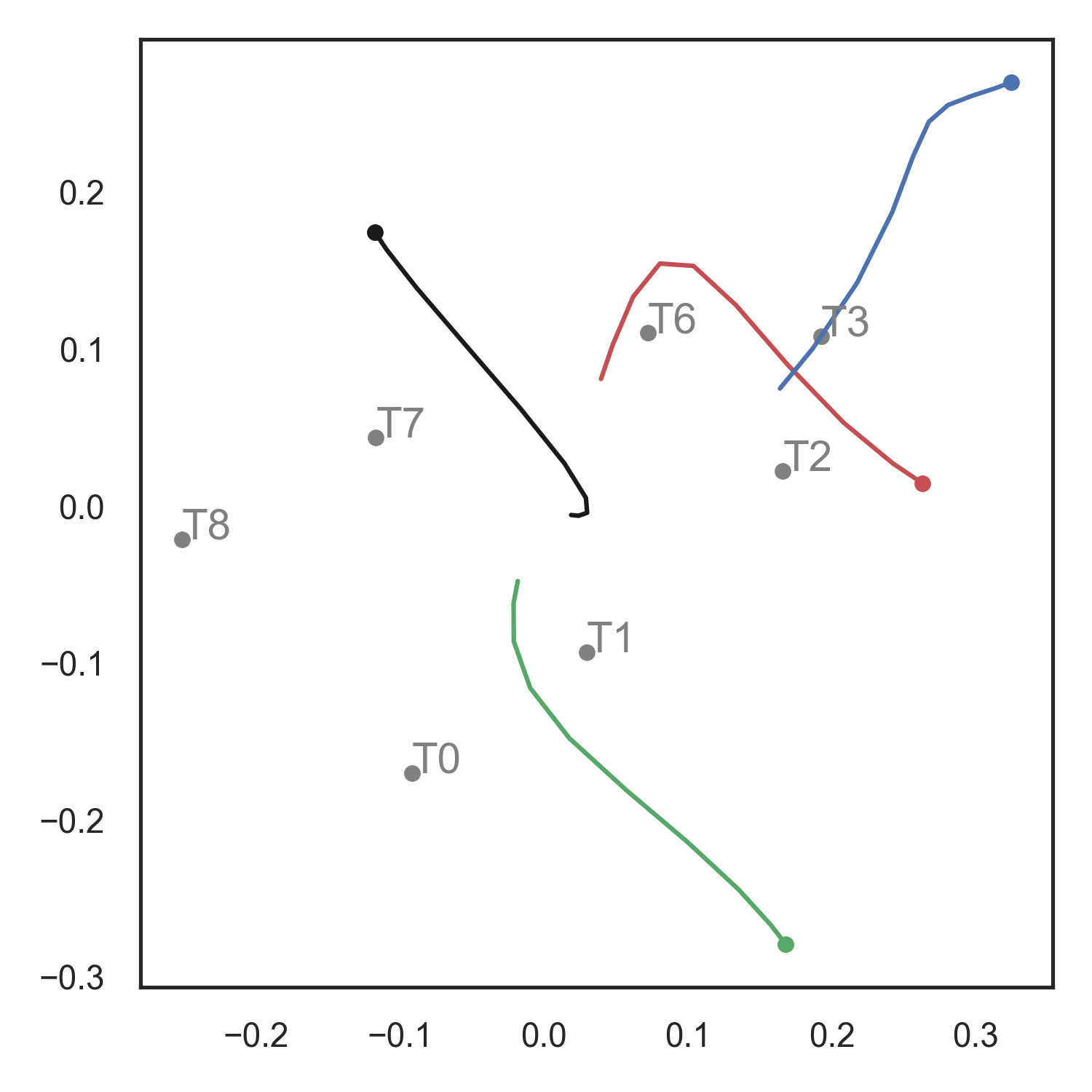}
& \includegraphics[align=c,width=.15\textwidth]{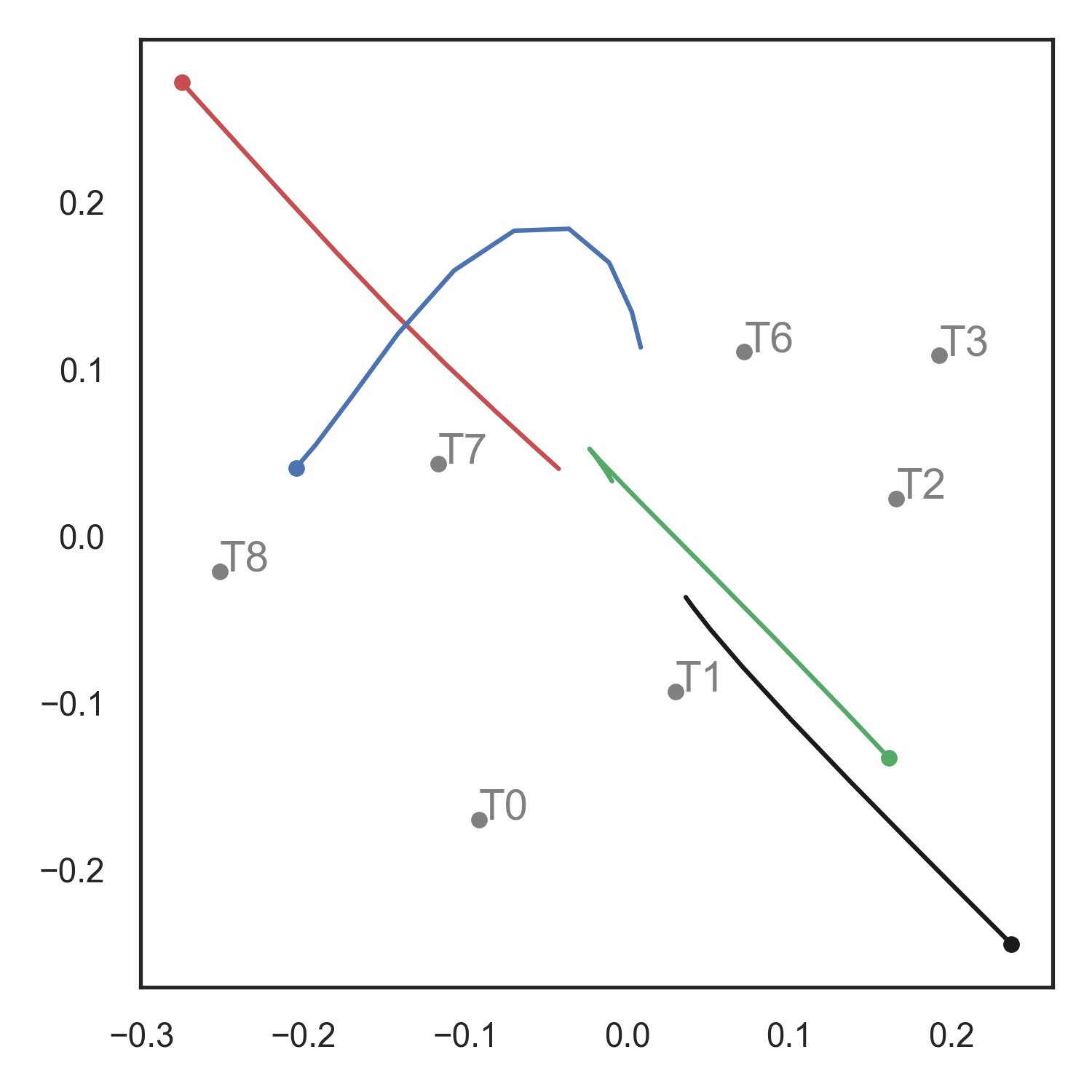} \\ 
\end{tabular}
\end{table*}

\begin{table*}[tb]
      \caption{The 1-step transition matrices of the trained policies
      }
      \label{tab:policy_viz} 
      \centering
     \begin{tabular}{c | c  c  c  c  c }
 & Anxiety & Depression & Schizophrenia & Suicidal & All Sessions \\ \hline
DISMOP-DDPG-TASK 
& \includegraphics[align=c,width=.15\textwidth]{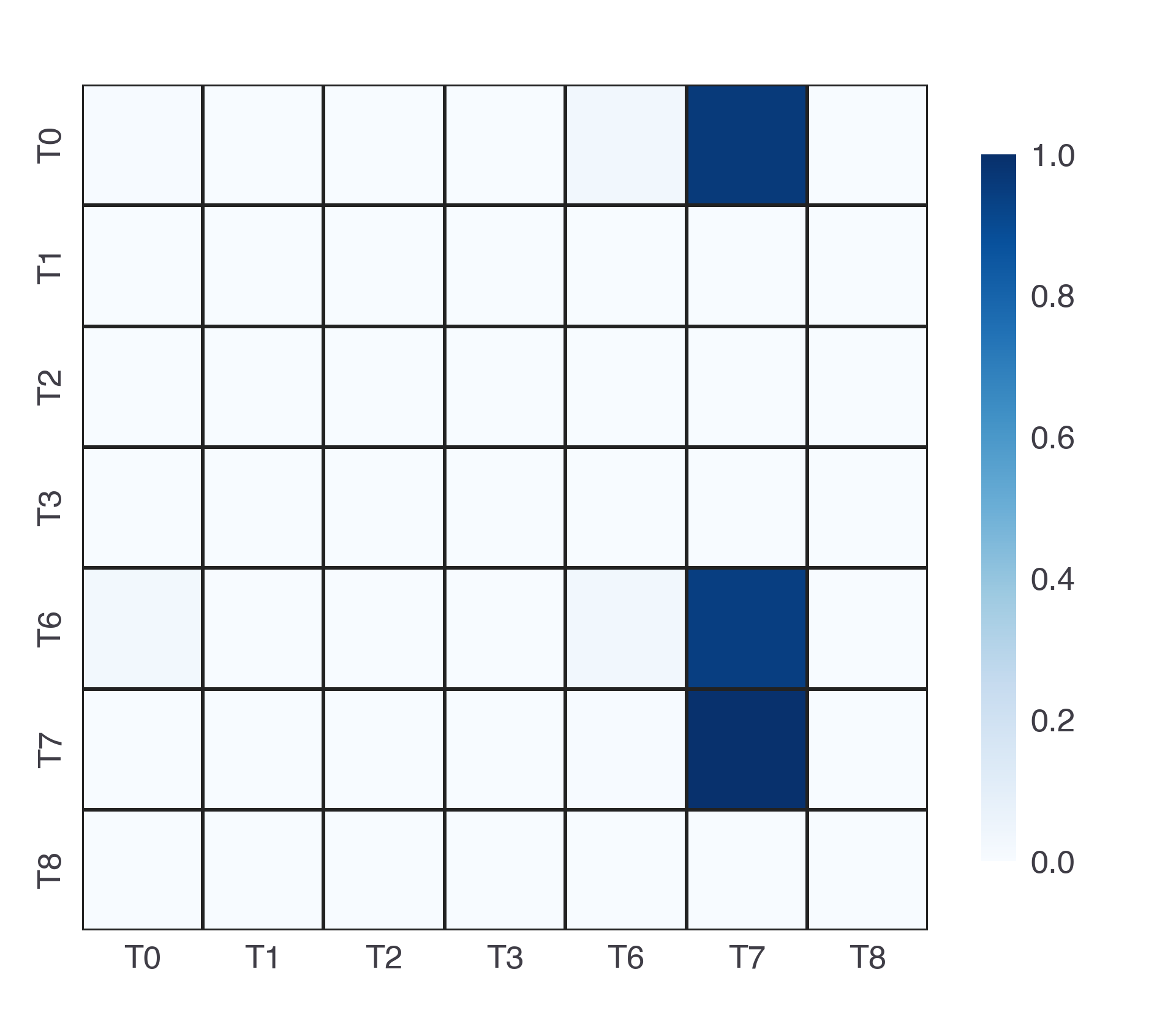}
& \includegraphics[align=c,width=.15\textwidth]{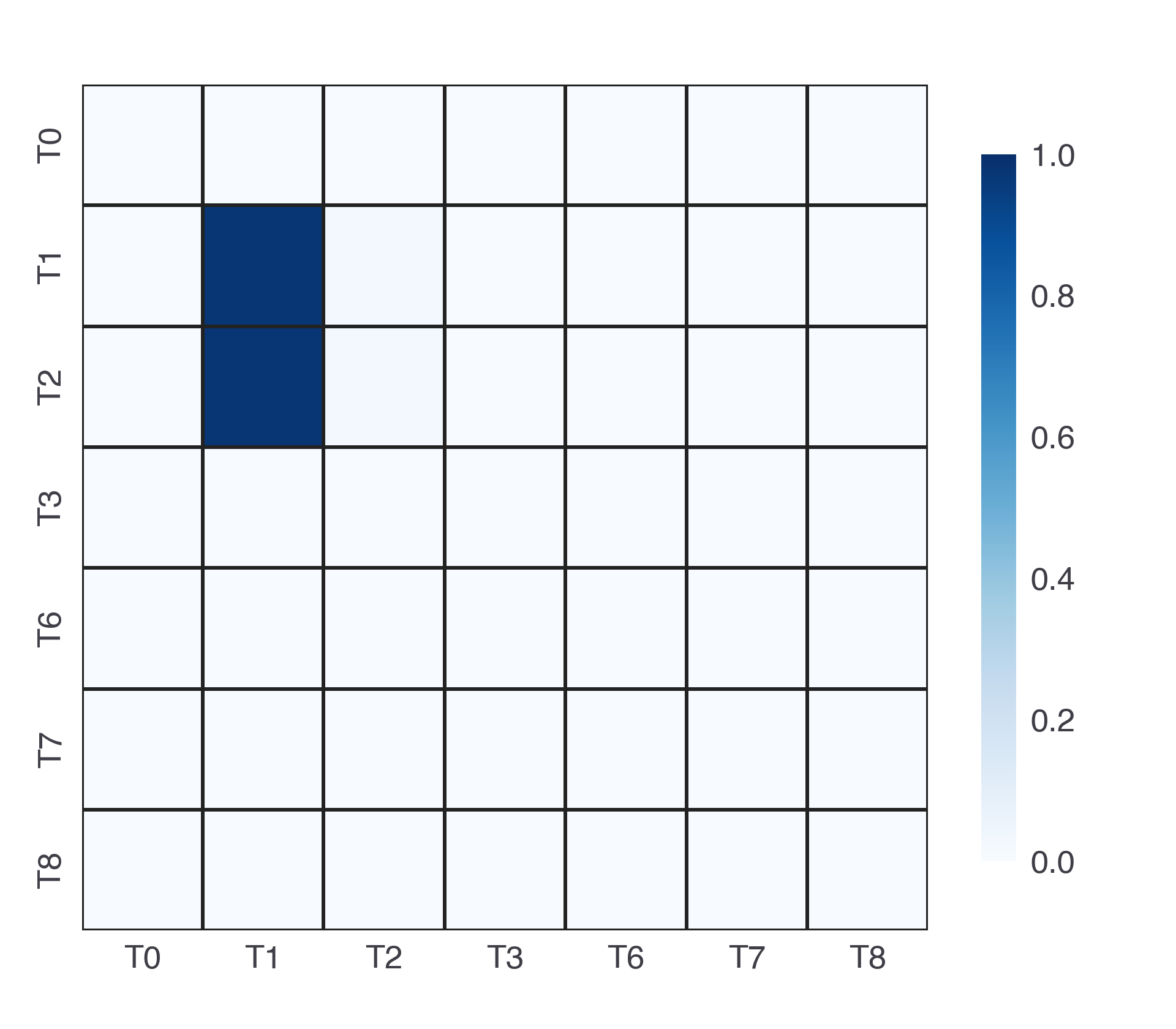}
& \includegraphics[align=c,width=.15\textwidth]{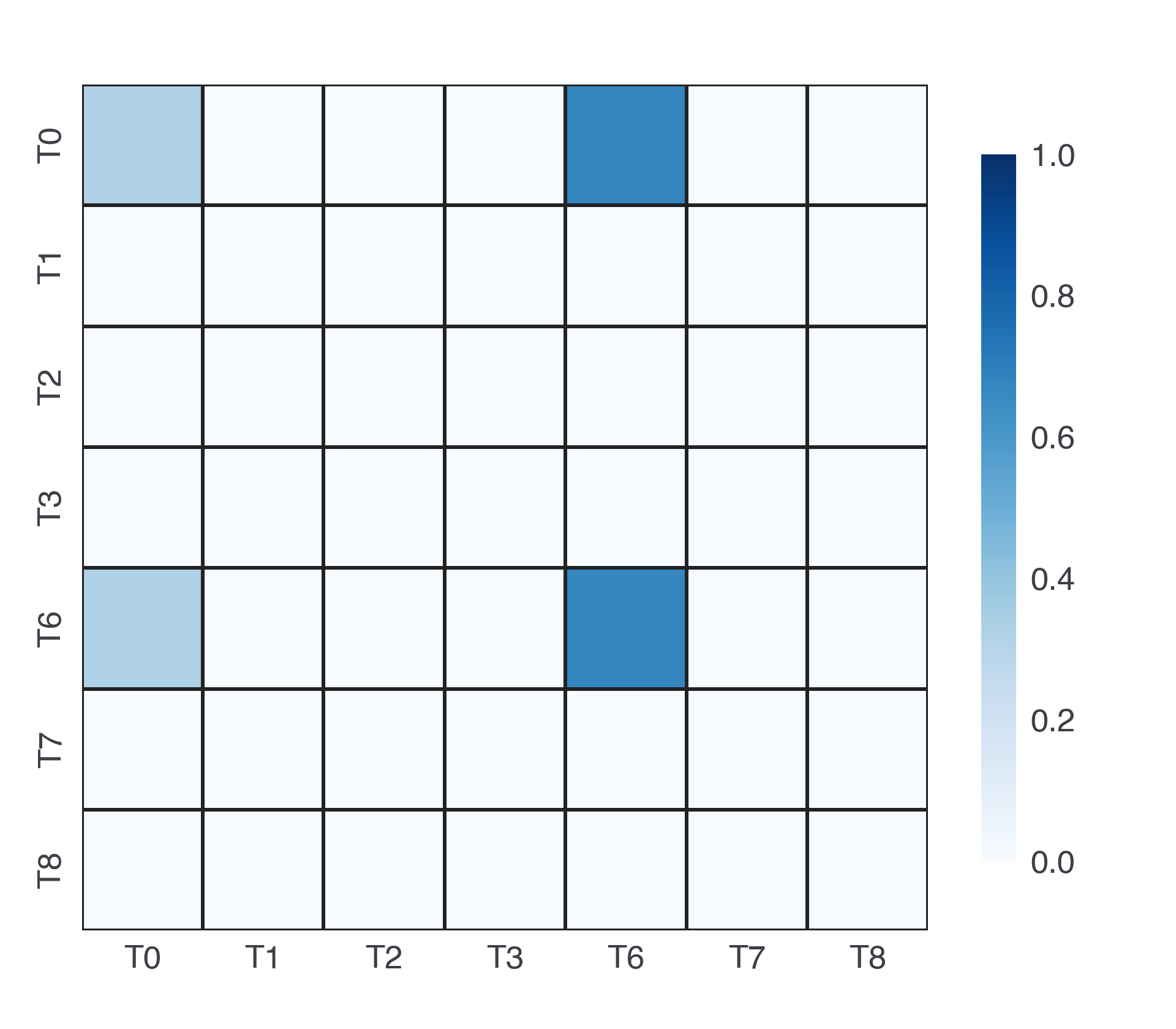}
& \includegraphics[align=c,width=.15\textwidth]{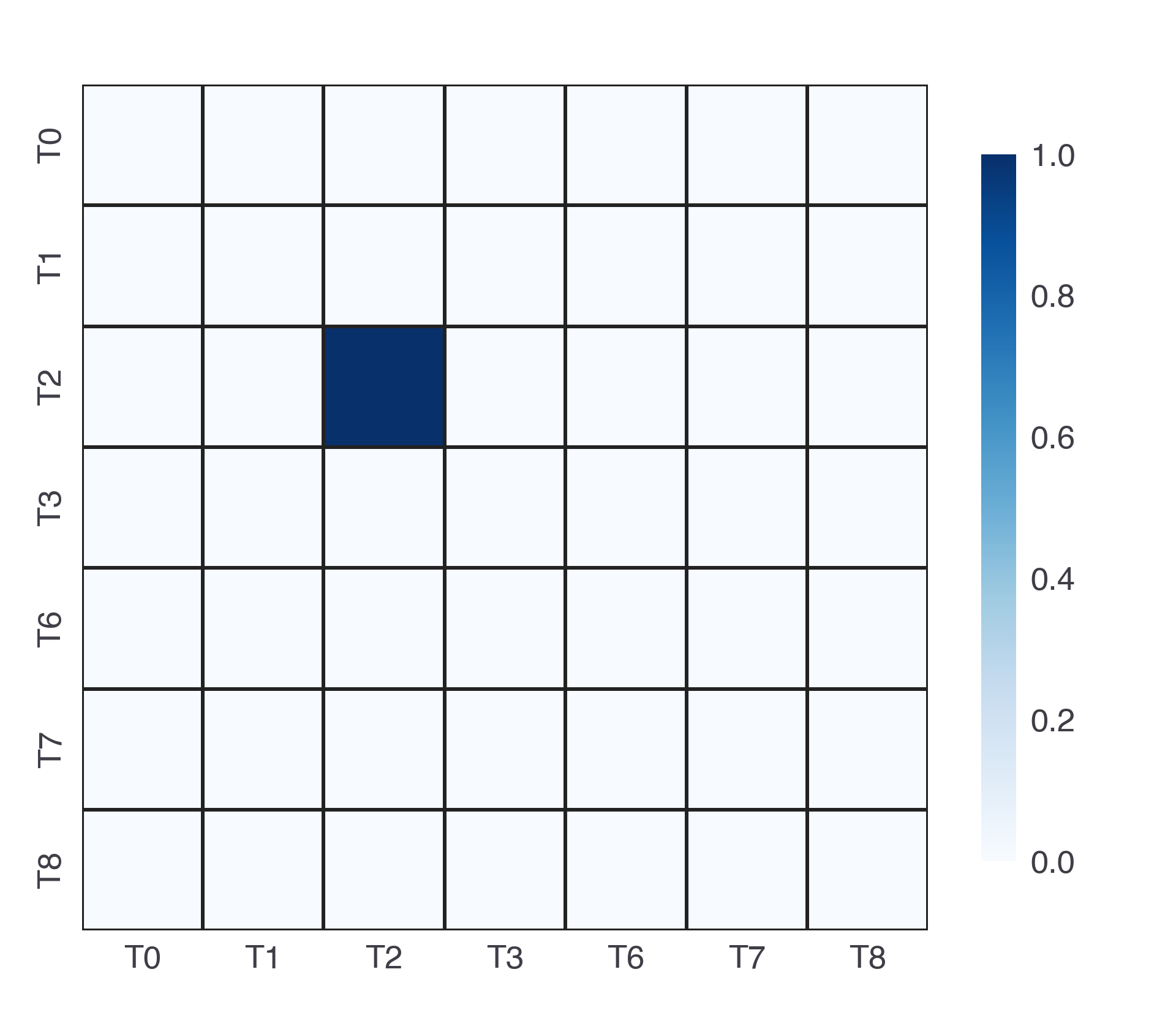}
& \includegraphics[align=c,width=.15\textwidth]{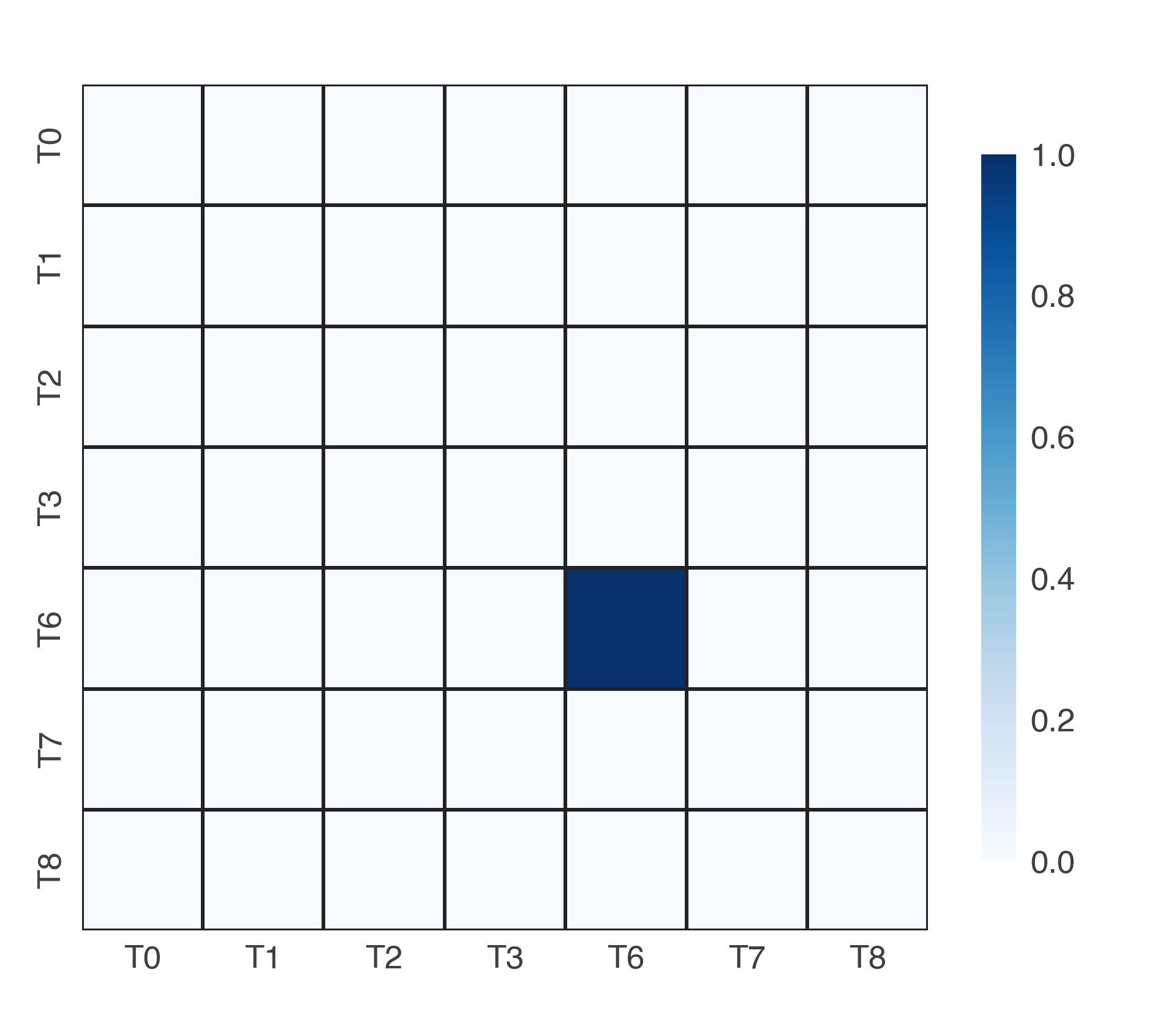}
\\ 
DISMOP-DDPG-BOND  
& \includegraphics[align=c,width=.15\textwidth]{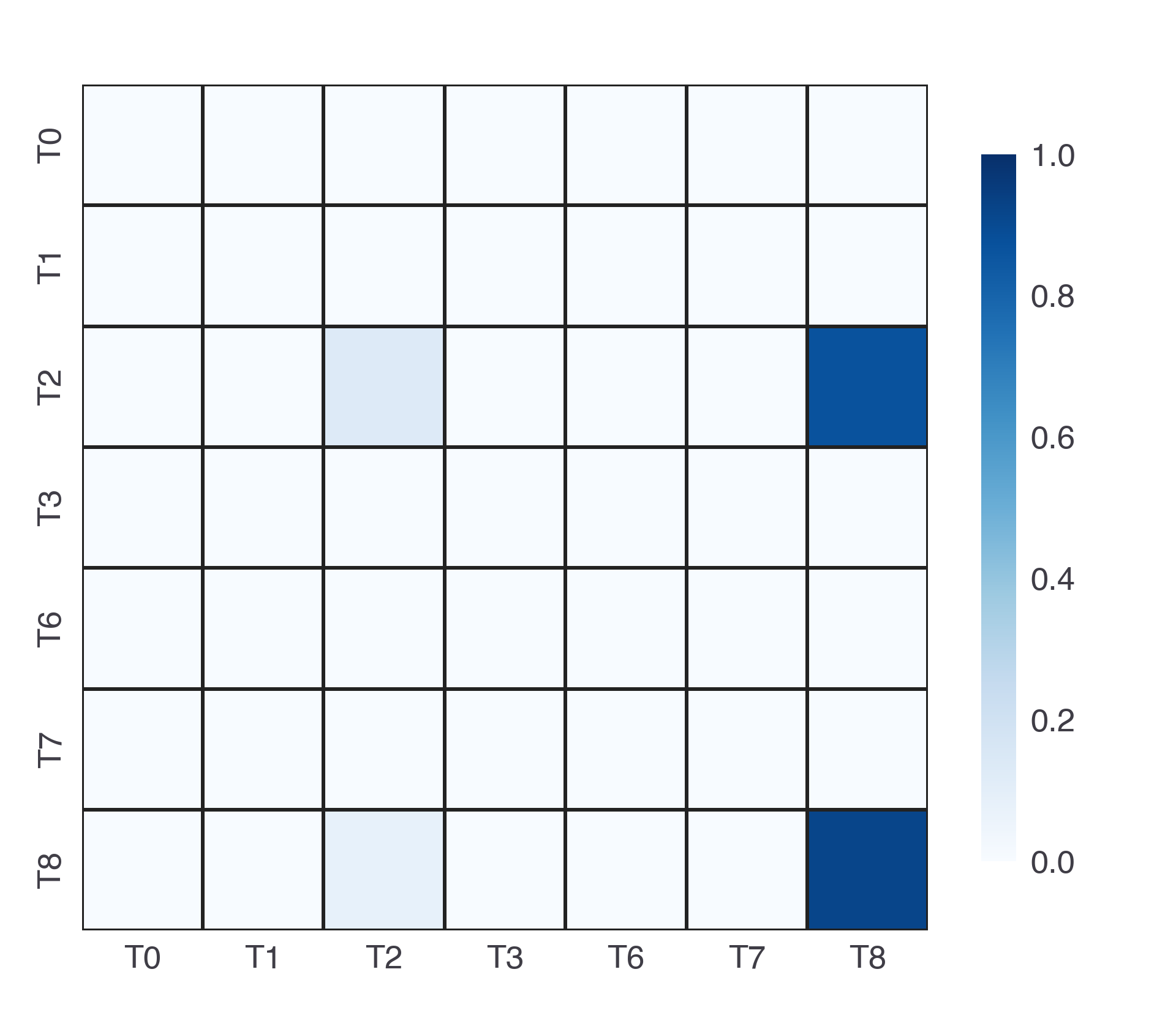}
& \includegraphics[align=c,width=.15\textwidth]{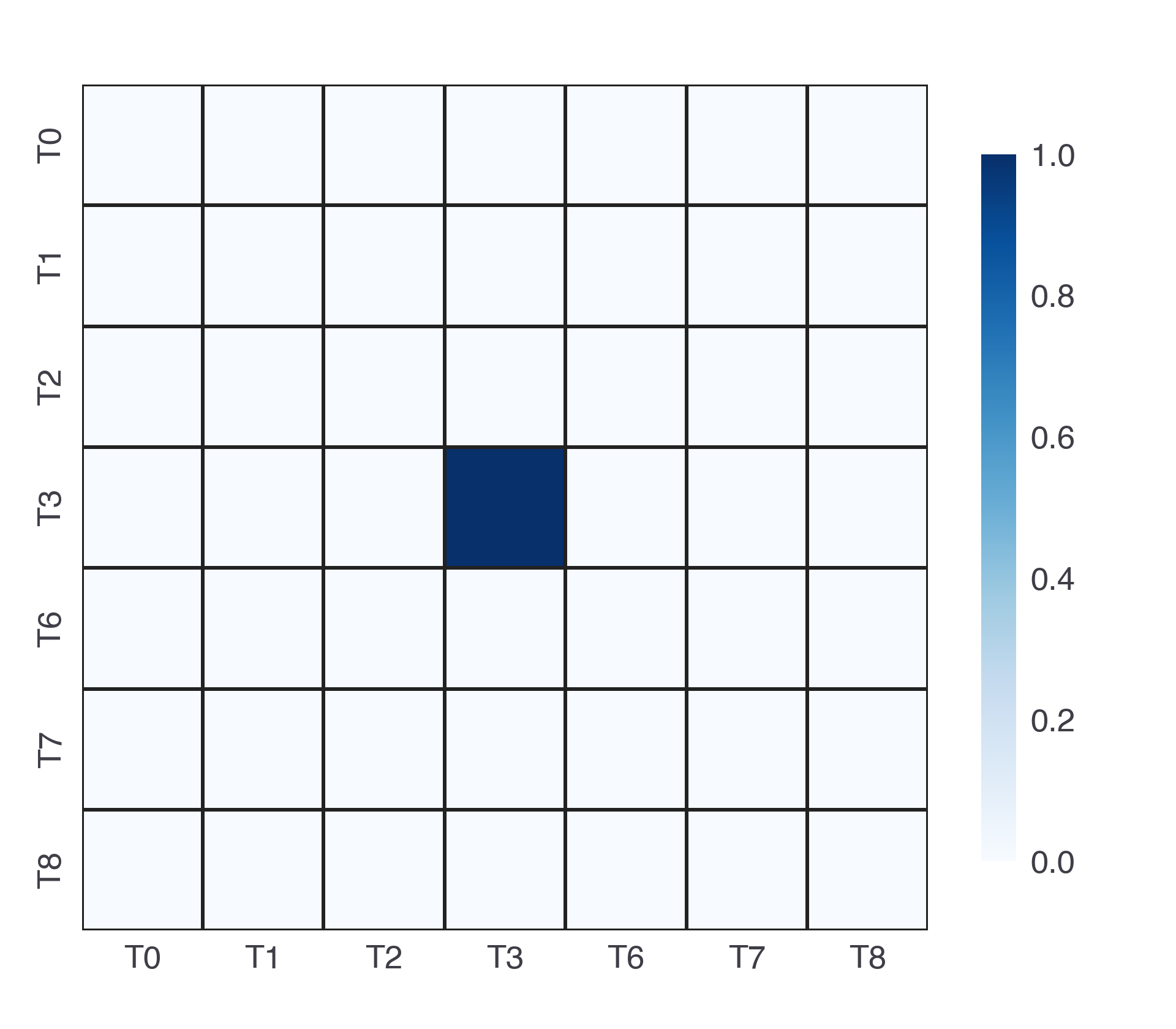}
& \includegraphics[align=c,width=.15\textwidth]{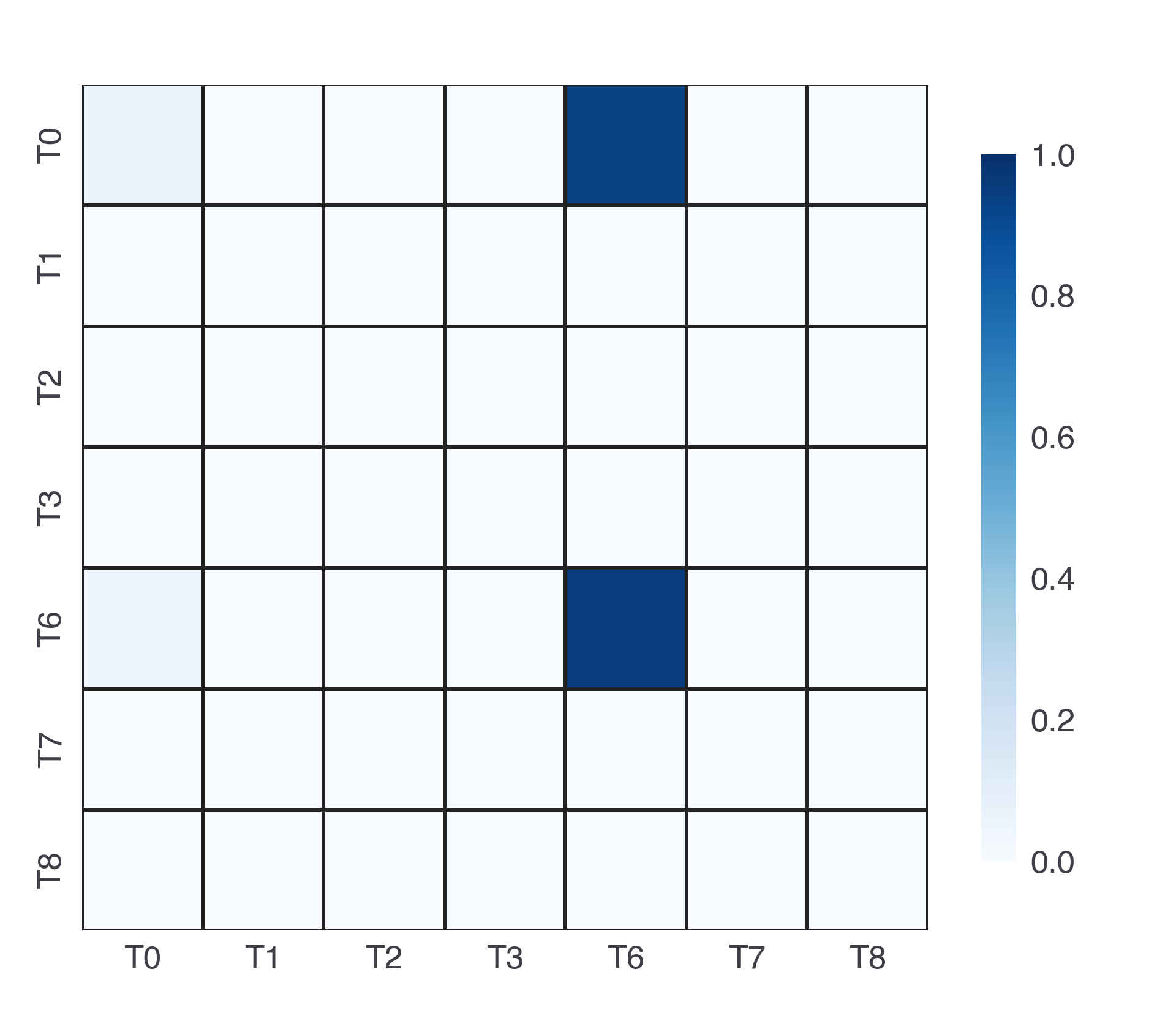}
& \includegraphics[align=c,width=.15\textwidth]{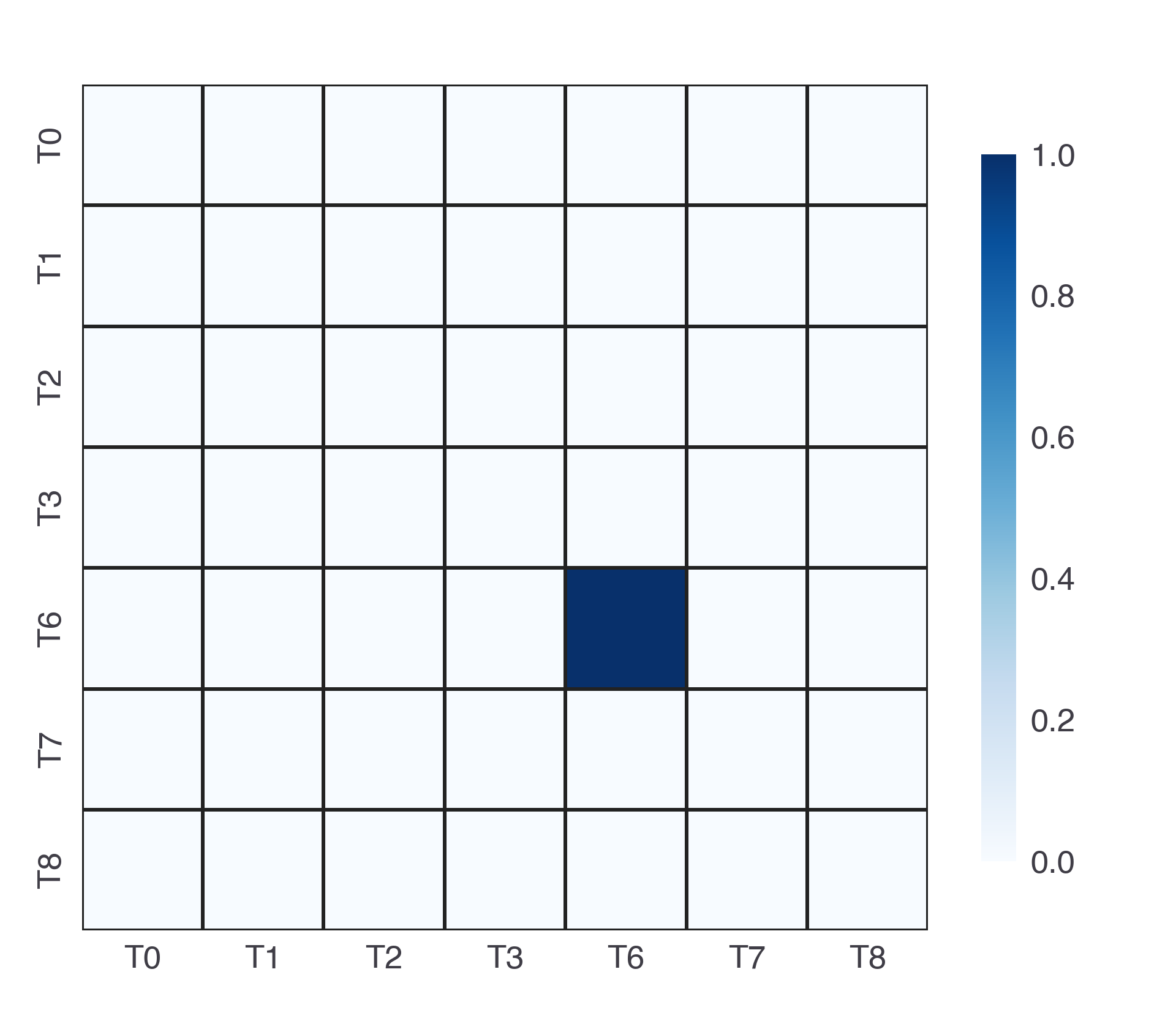}
& \includegraphics[align=c,width=.15\textwidth]{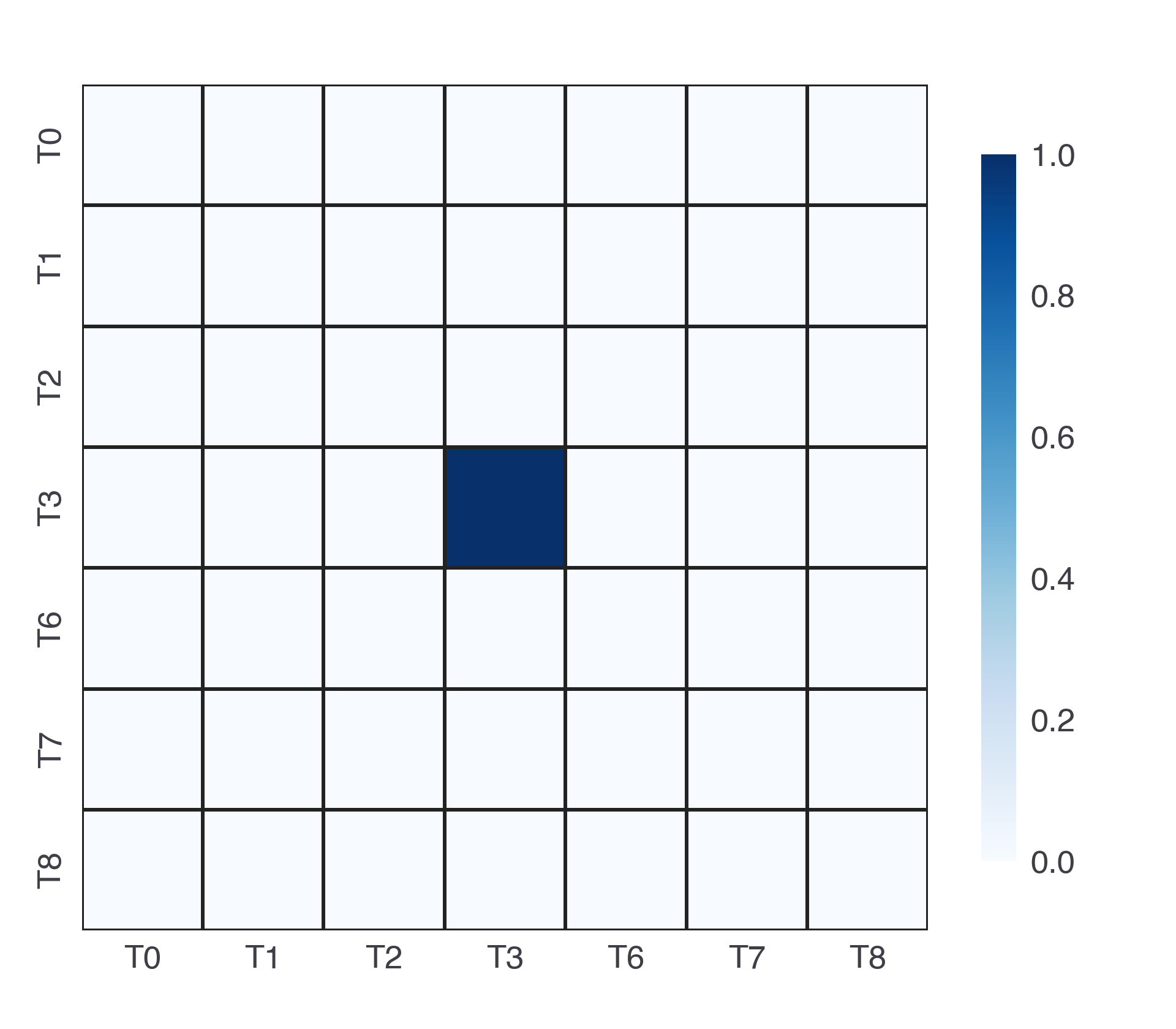}
\\ 
DISMOP-DDPG-GOAL  
& \includegraphics[align=c,width=.15\textwidth]{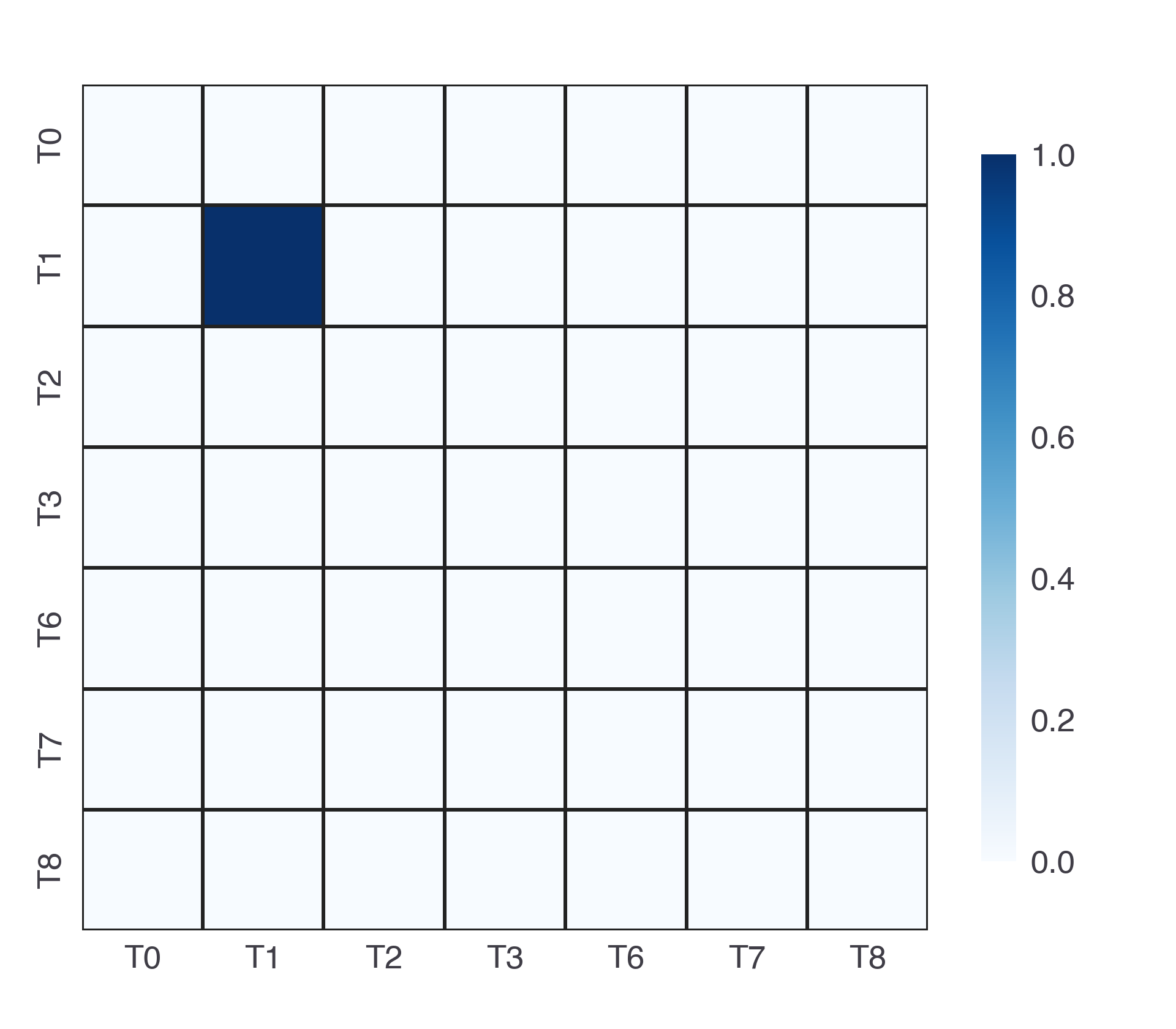}
& \includegraphics[align=c,width=.15\textwidth]{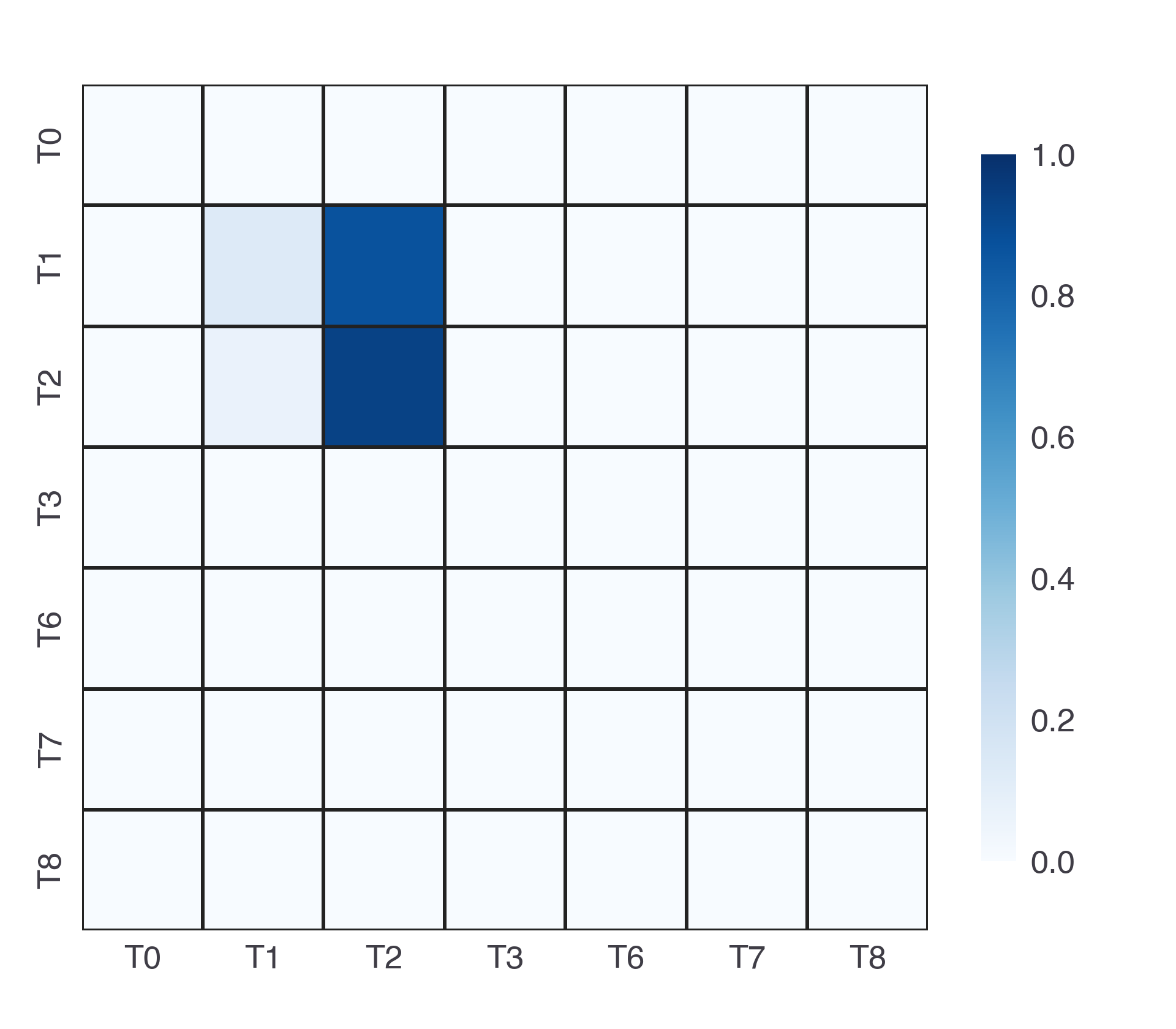}
& \includegraphics[align=c,width=.15\textwidth]{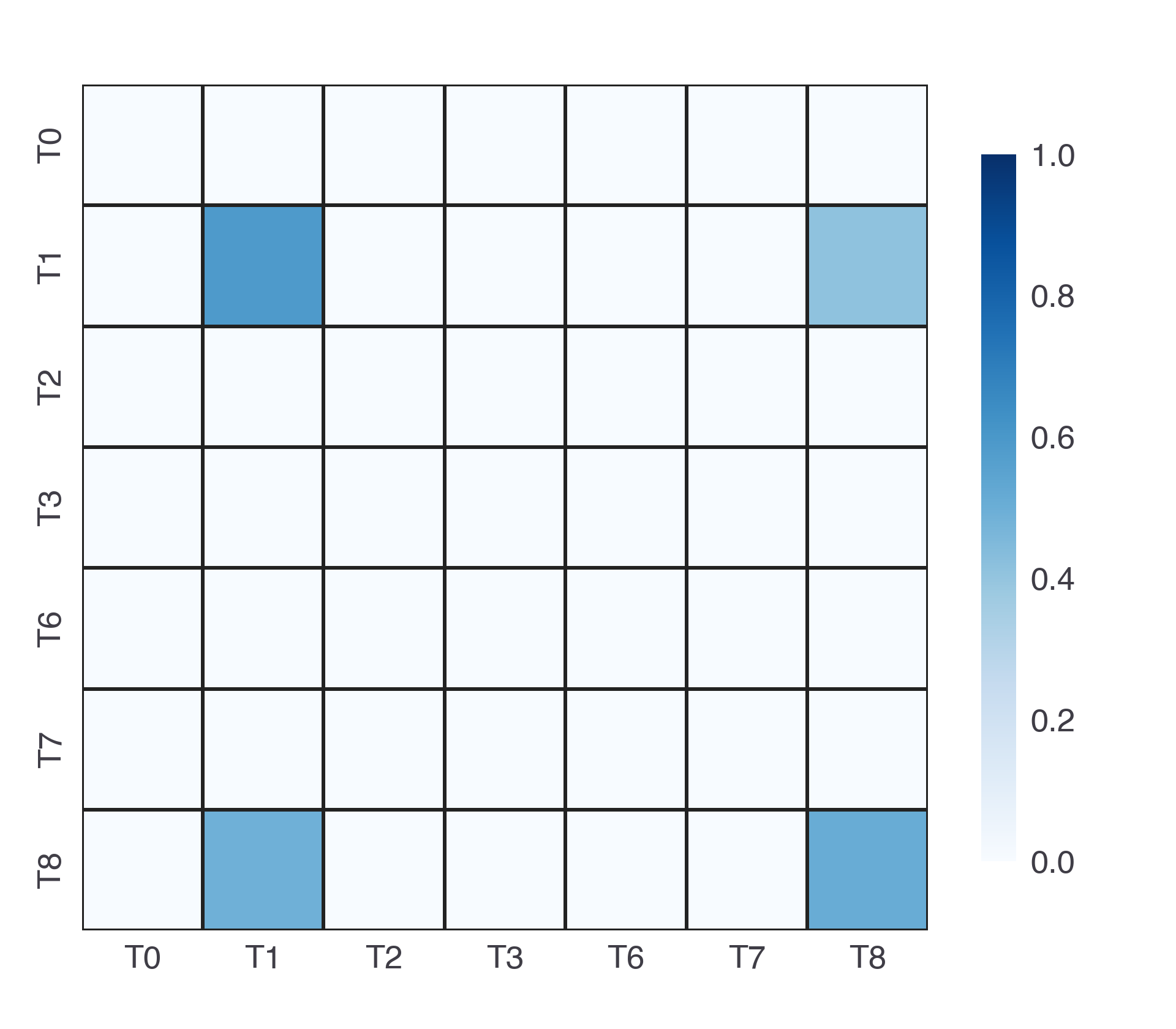}
& \includegraphics[align=c,width=.15\textwidth]{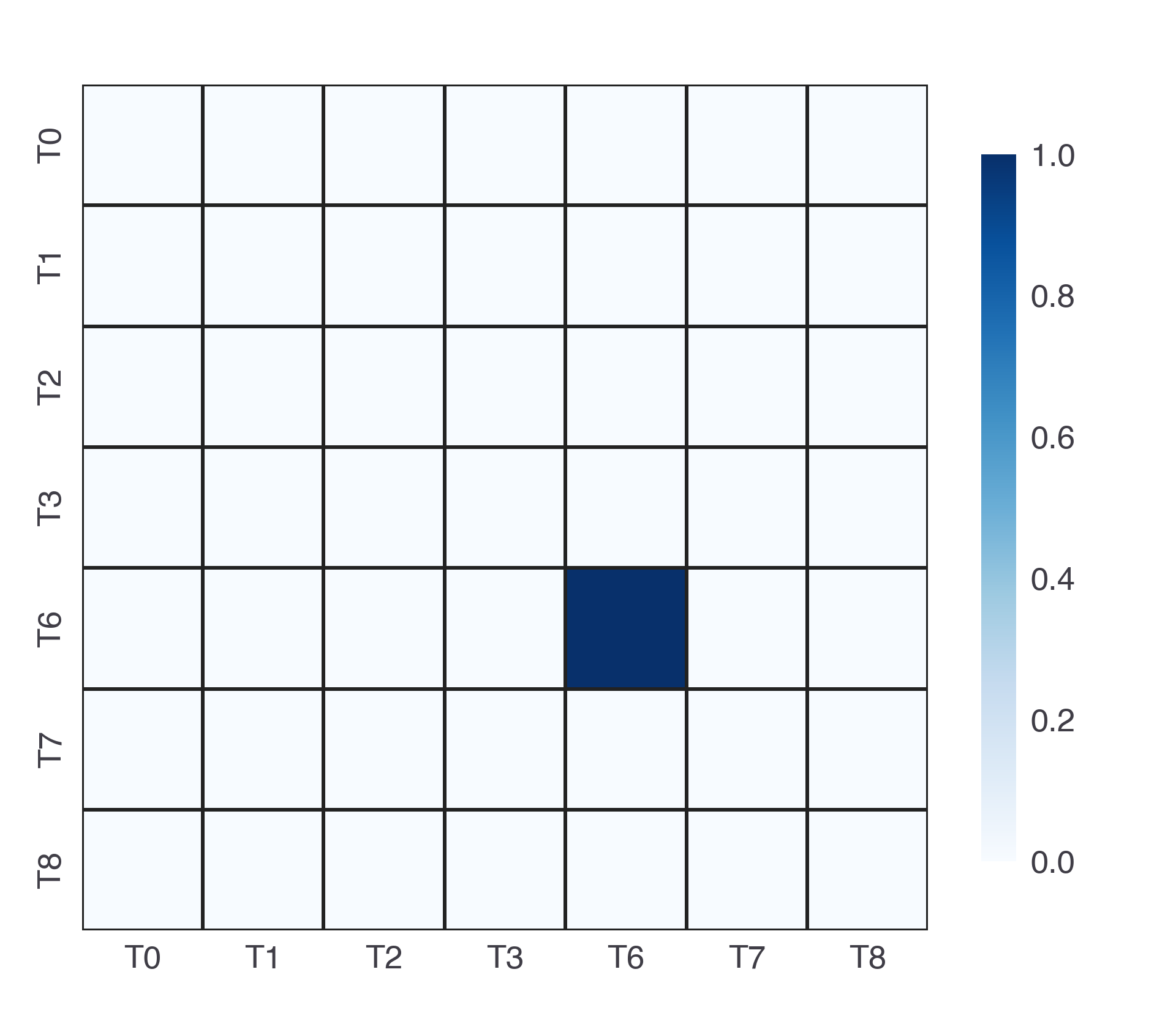}
& \includegraphics[align=c,width=.15\textwidth]{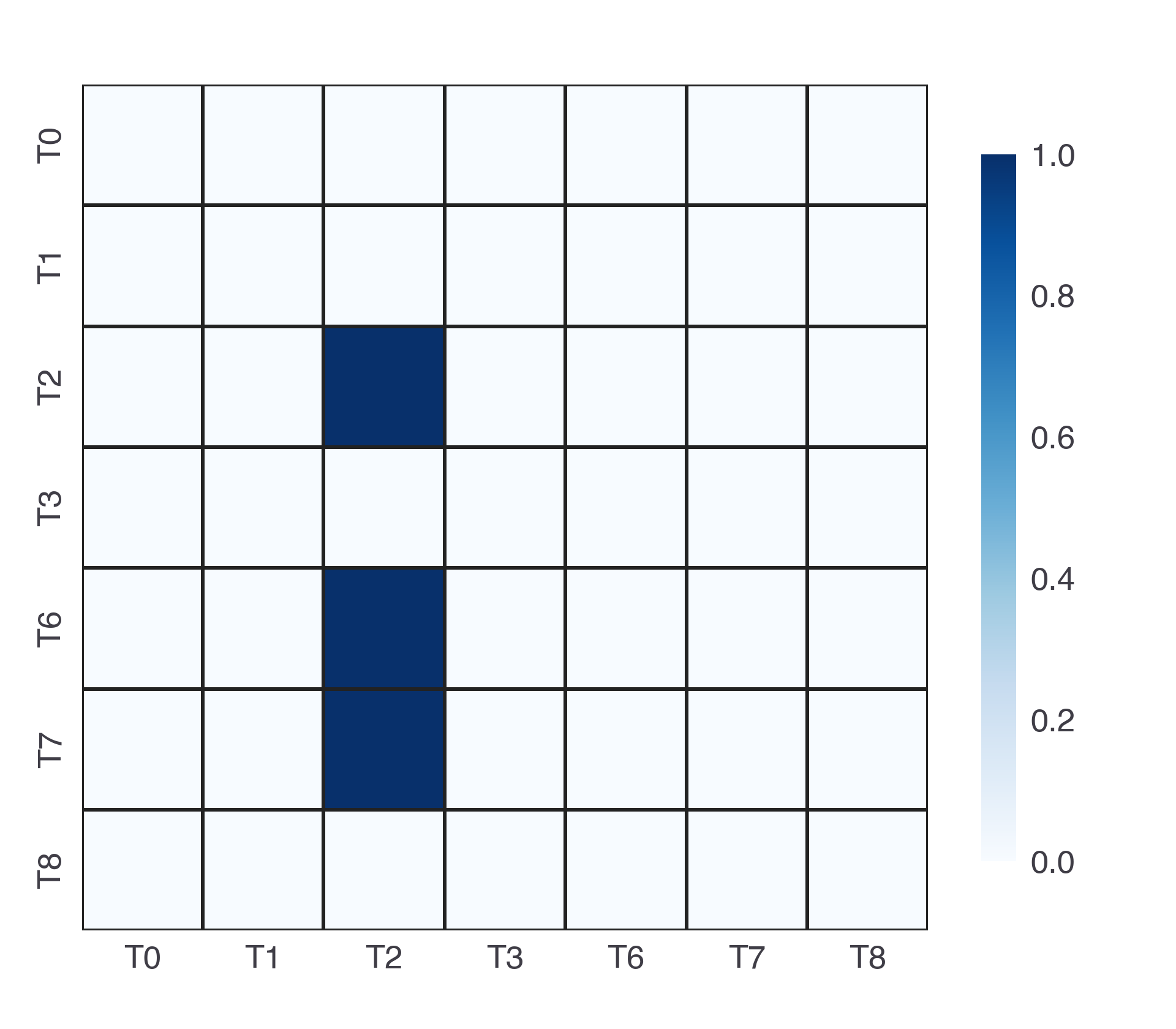}
\\ \hline
DISMOP-TD3-TASK   
& \includegraphics[align=c,width=.15\textwidth]{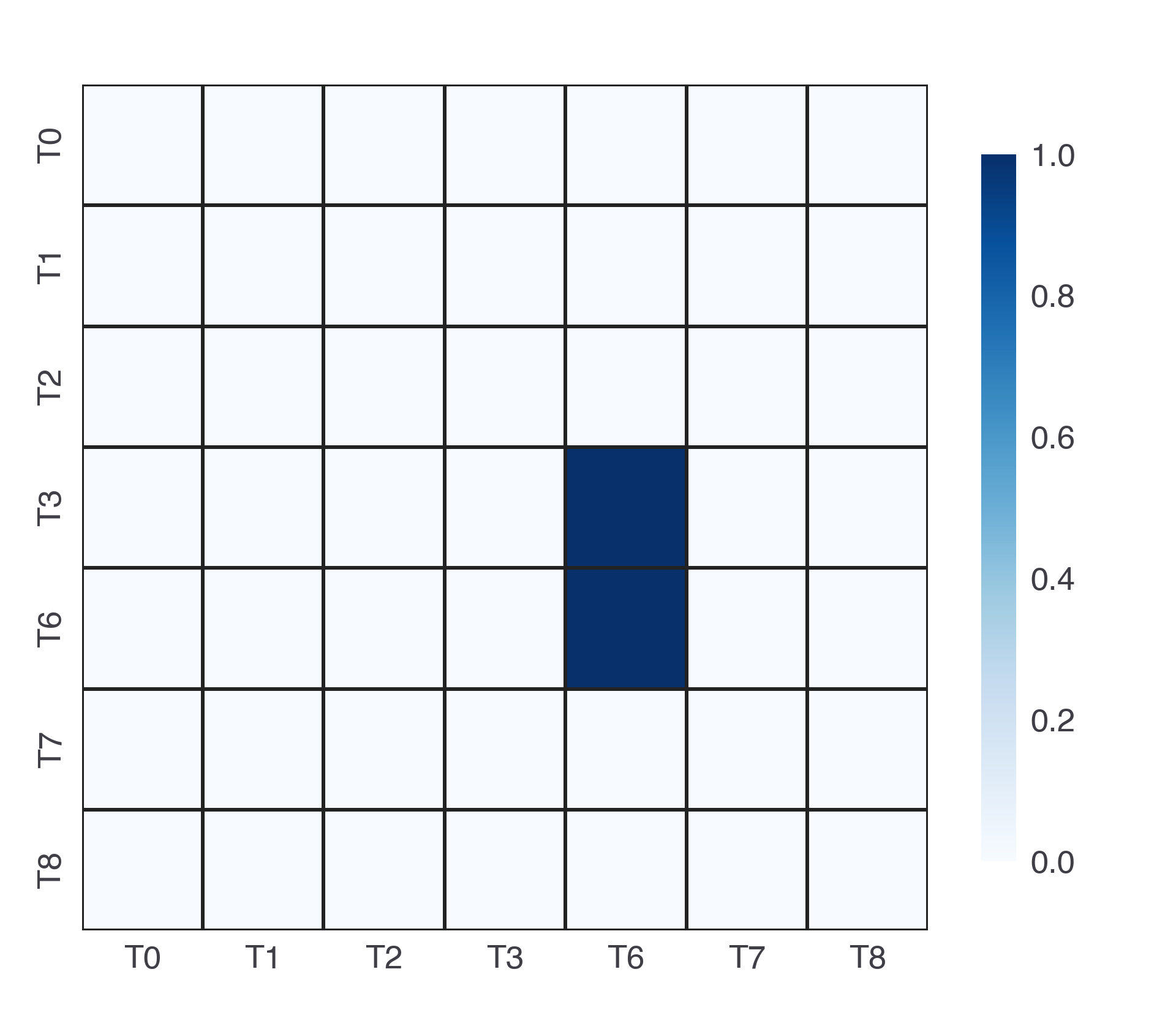}
& \includegraphics[align=c,width=.15\textwidth]{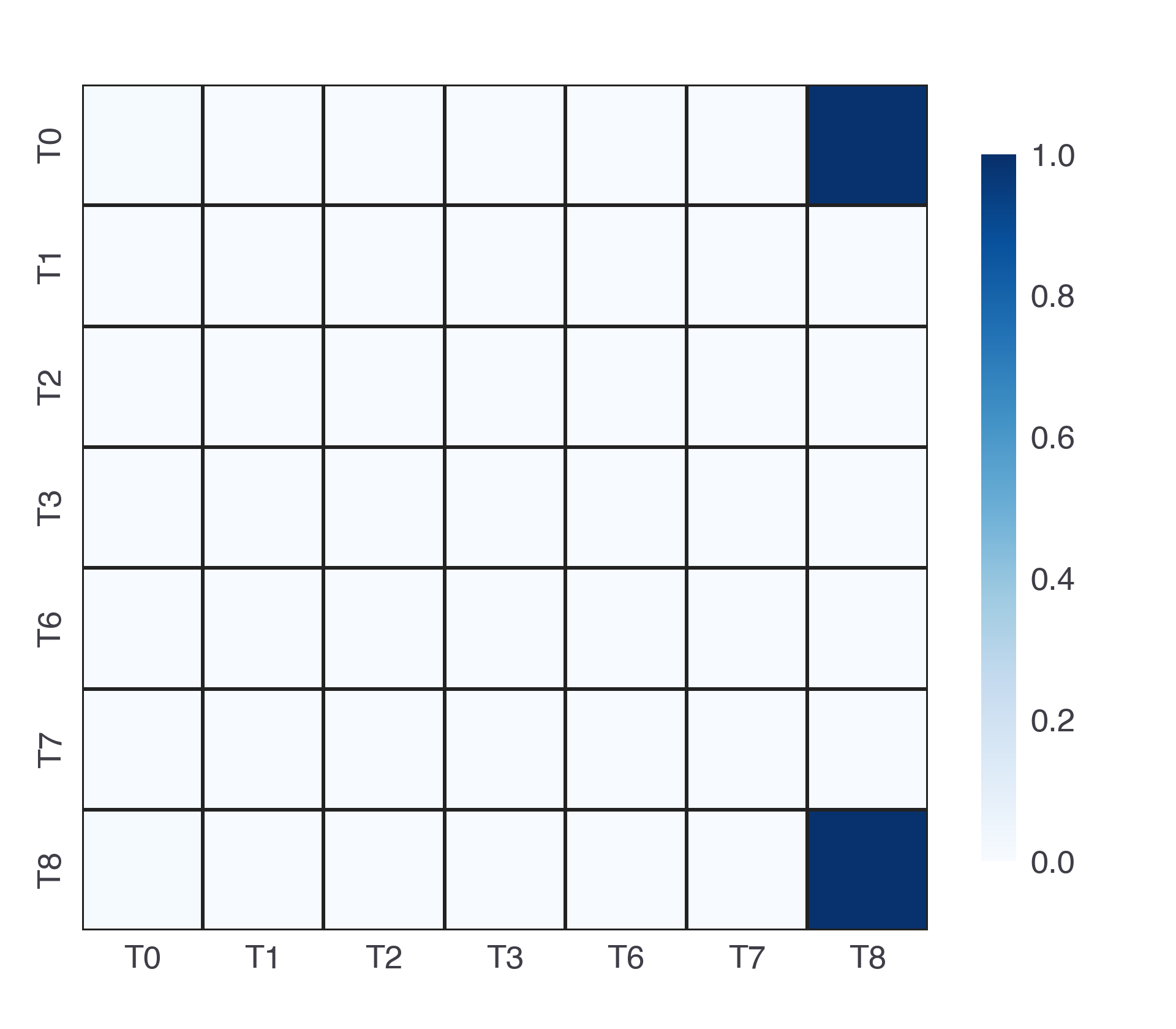}
& \includegraphics[align=c,width=.15\textwidth]{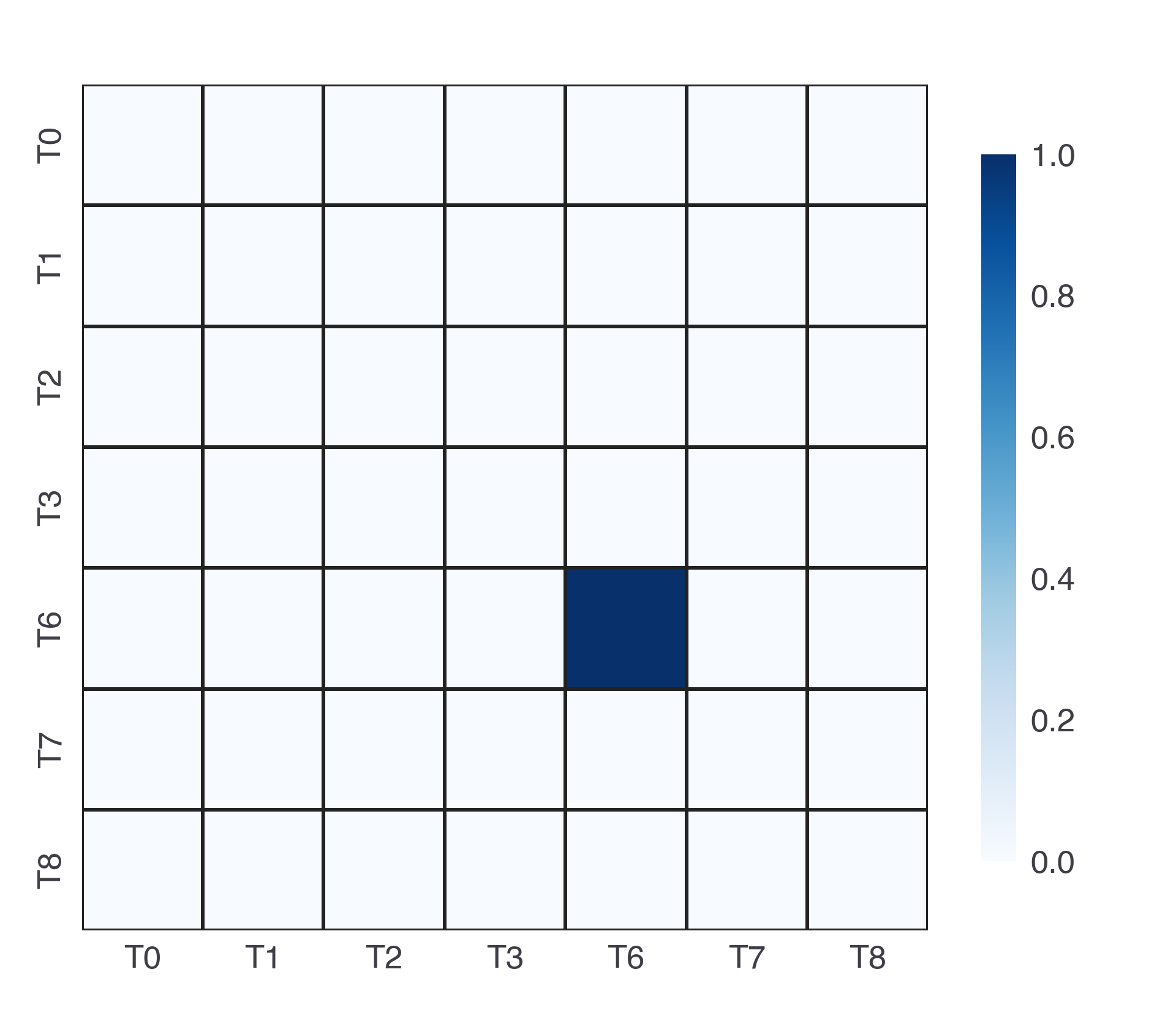}
& \includegraphics[align=c,width=.15\textwidth]{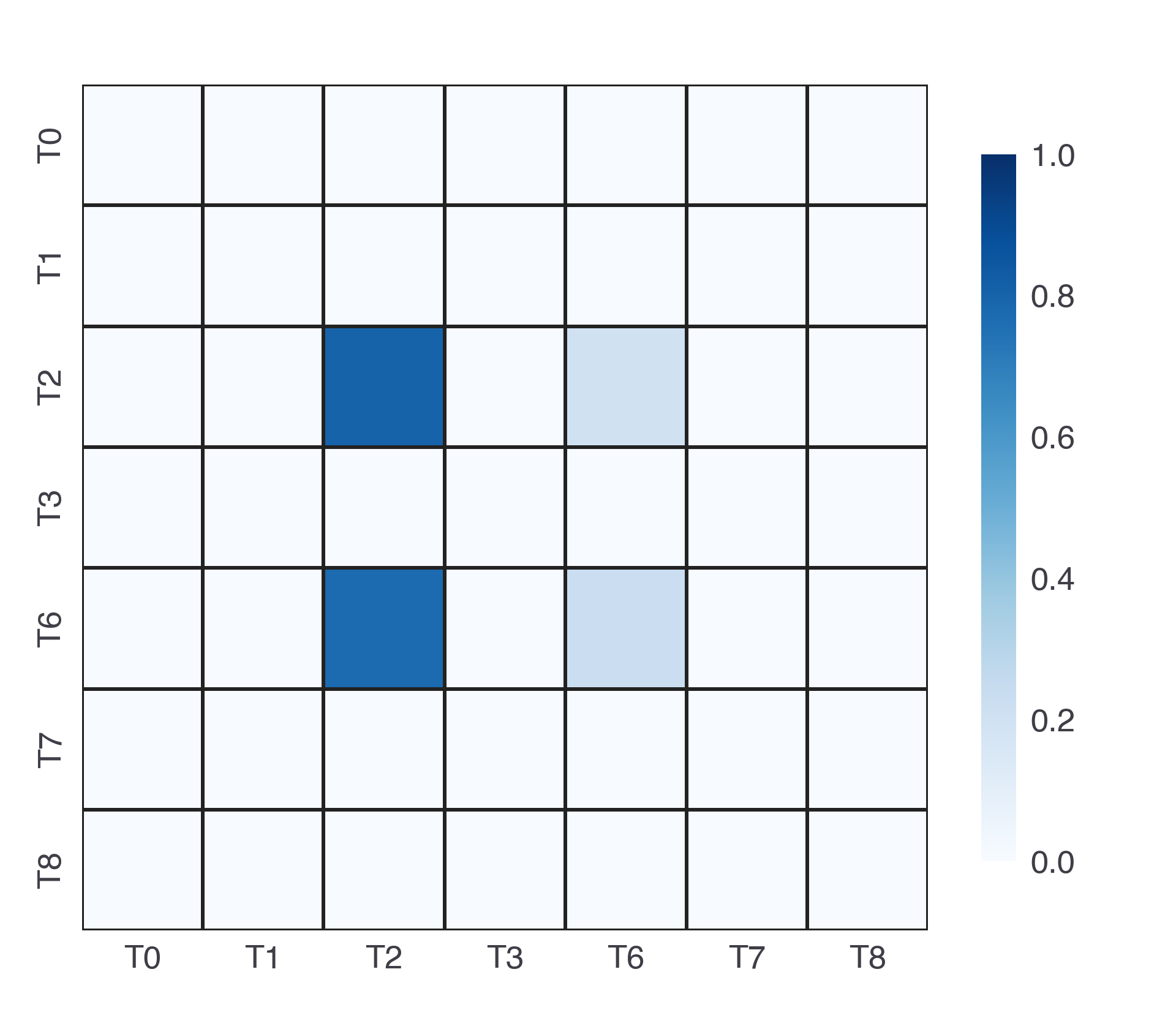}
& \includegraphics[align=c,width=.15\textwidth]{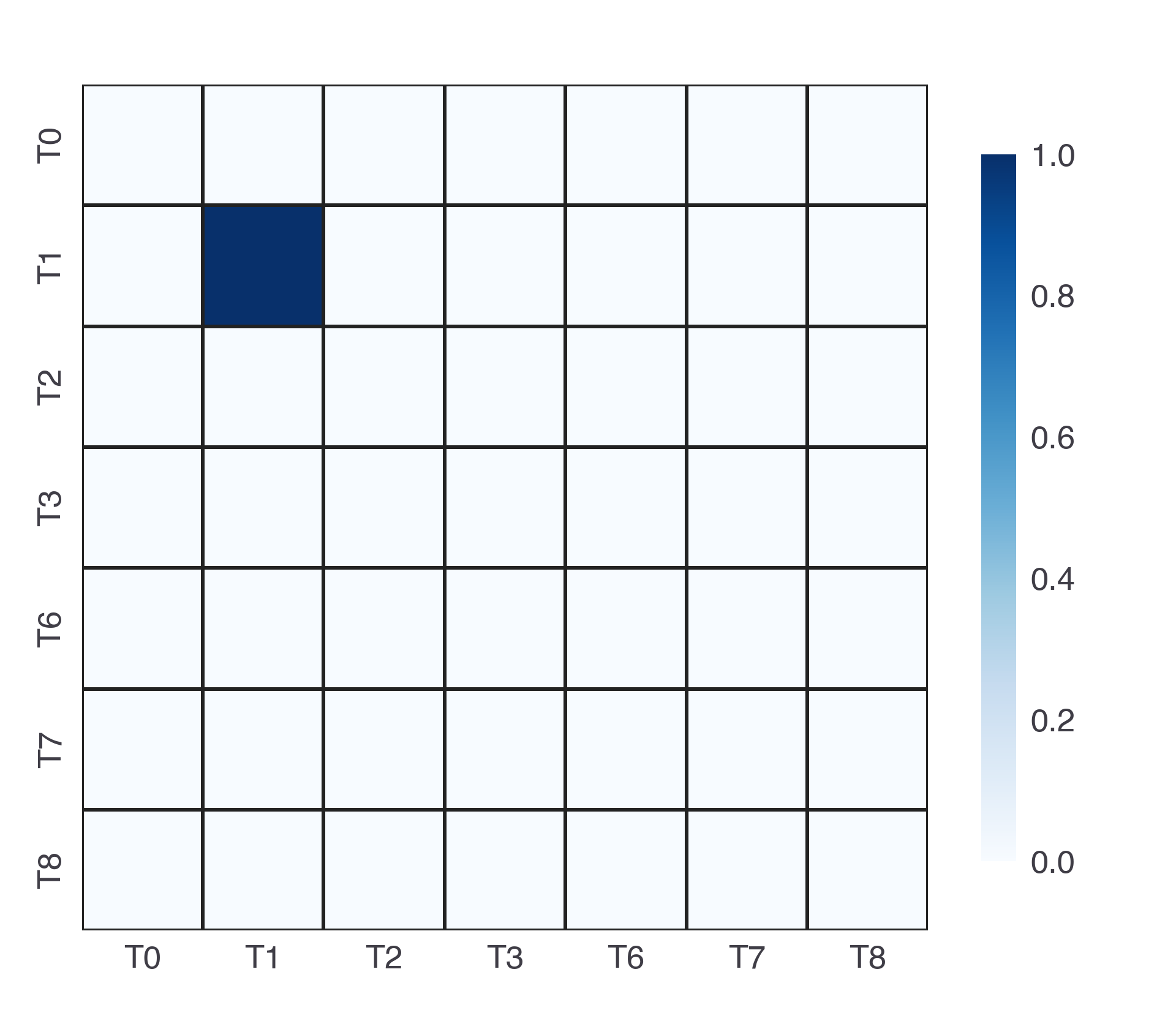}
\\ 
DISMOP-TD3-BOND  
& \includegraphics[align=c,width=.15\textwidth]{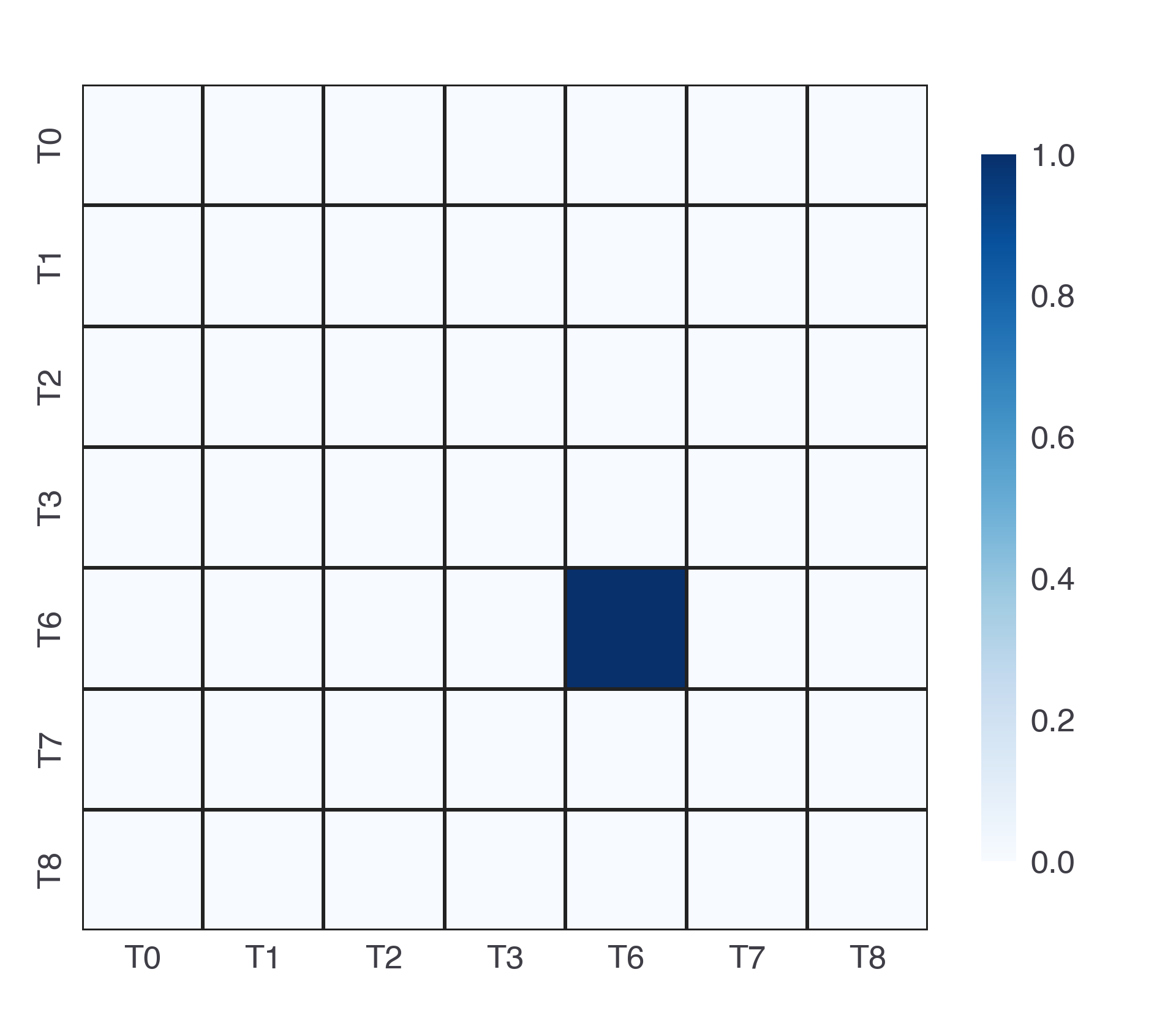}
& \includegraphics[align=c,width=.15\textwidth]{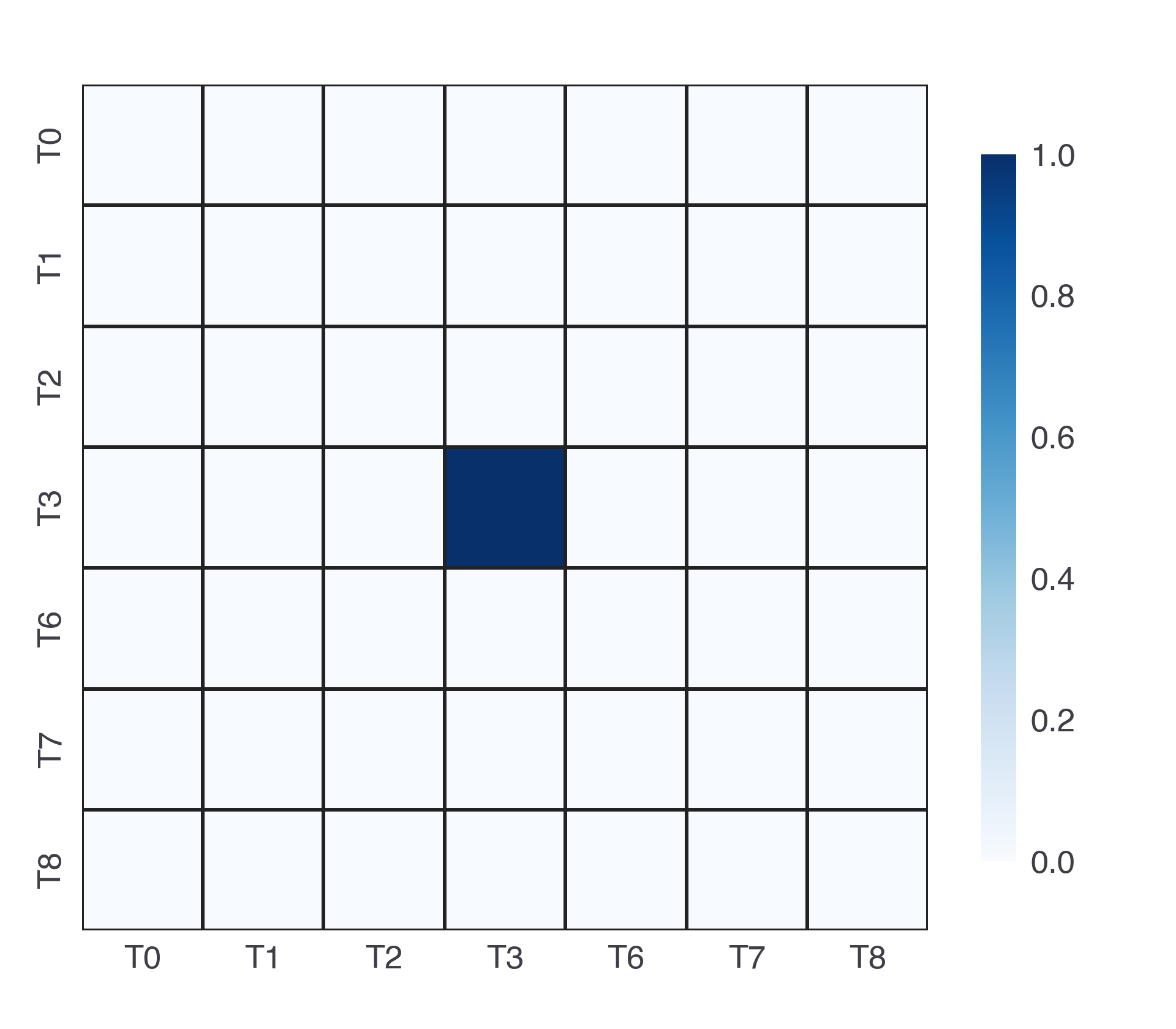}
& \includegraphics[align=c,width=.15\textwidth]{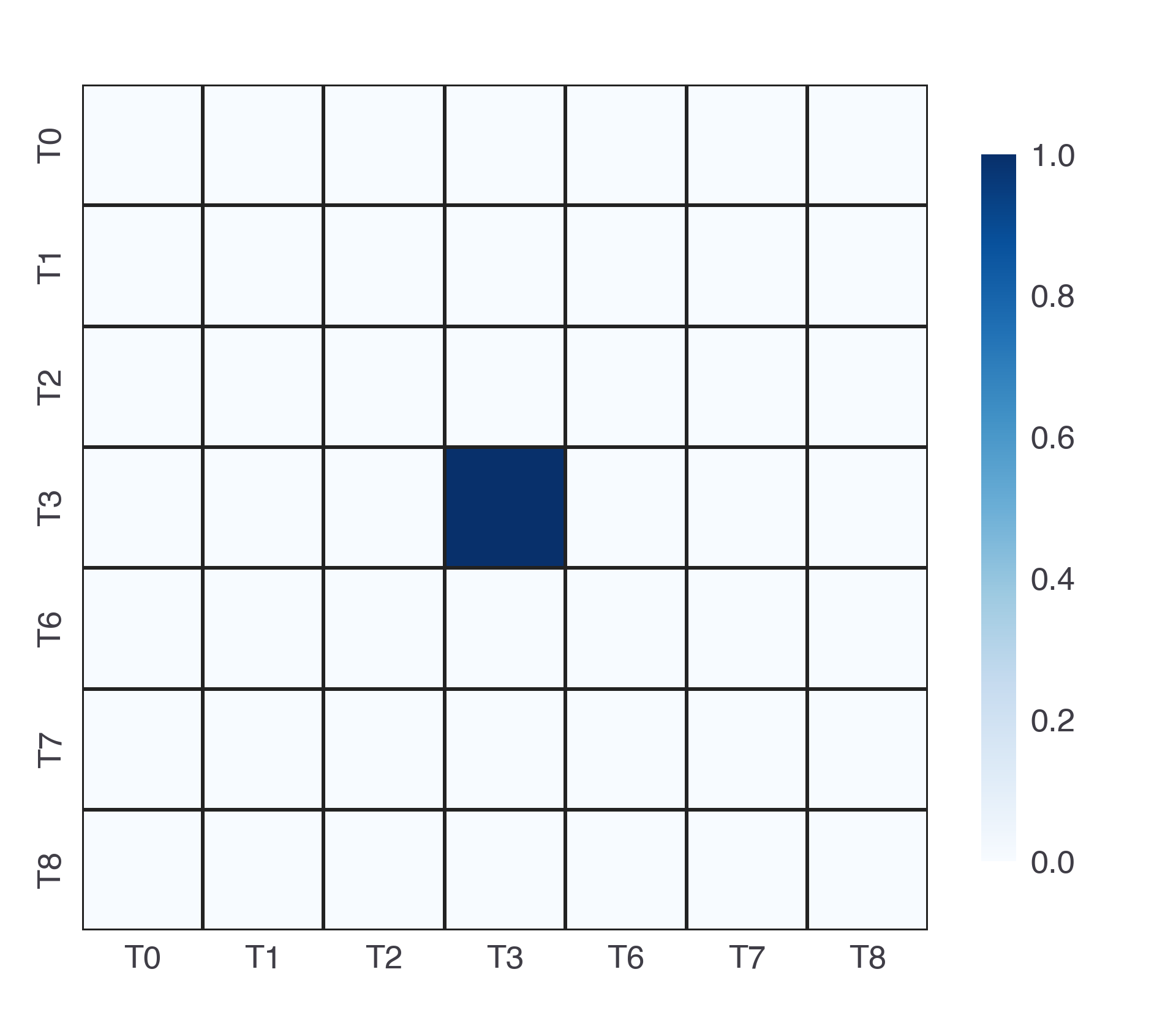}
& \includegraphics[align=c,width=.15\textwidth]{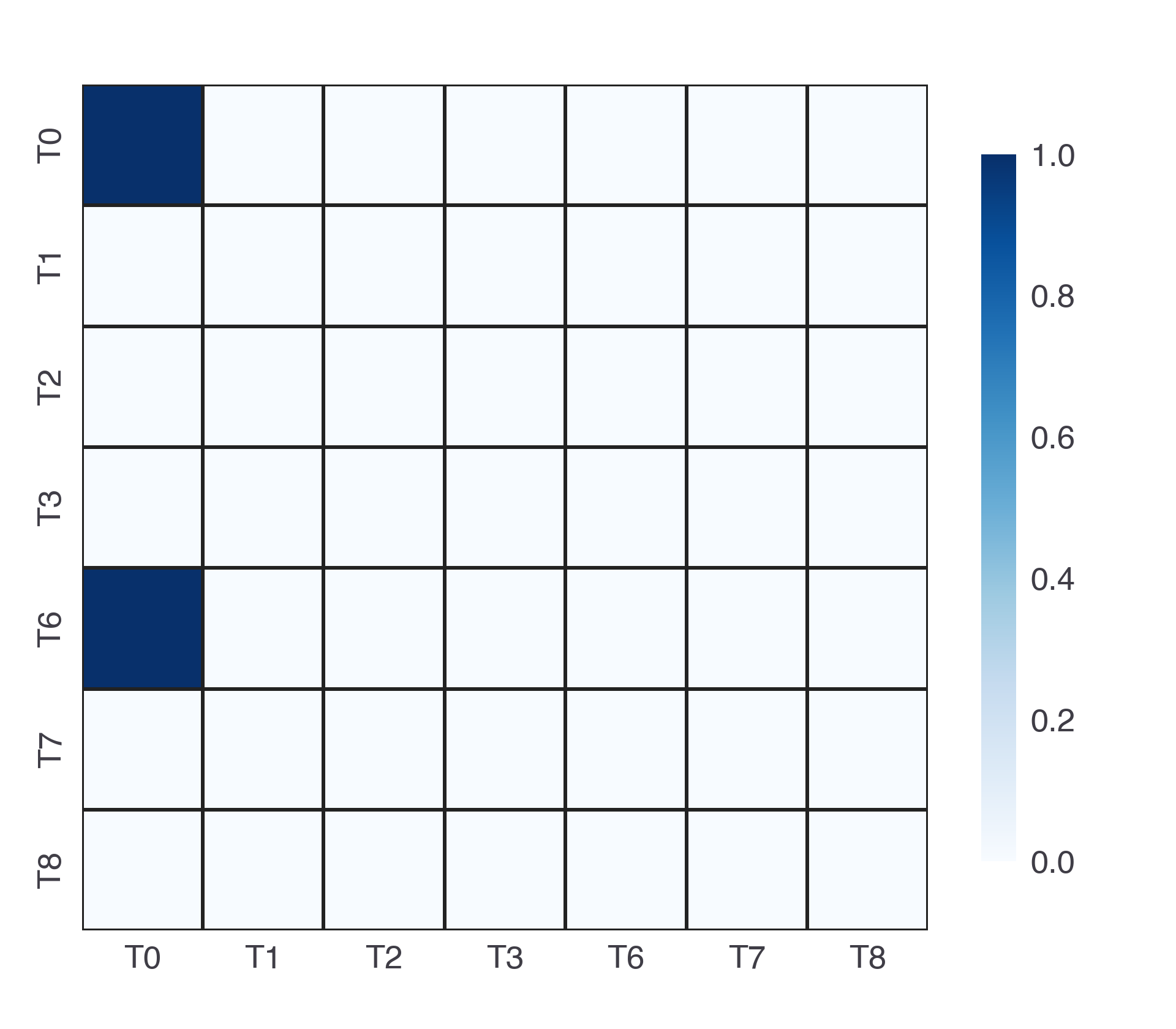}
& \includegraphics[align=c,width=.15\textwidth]{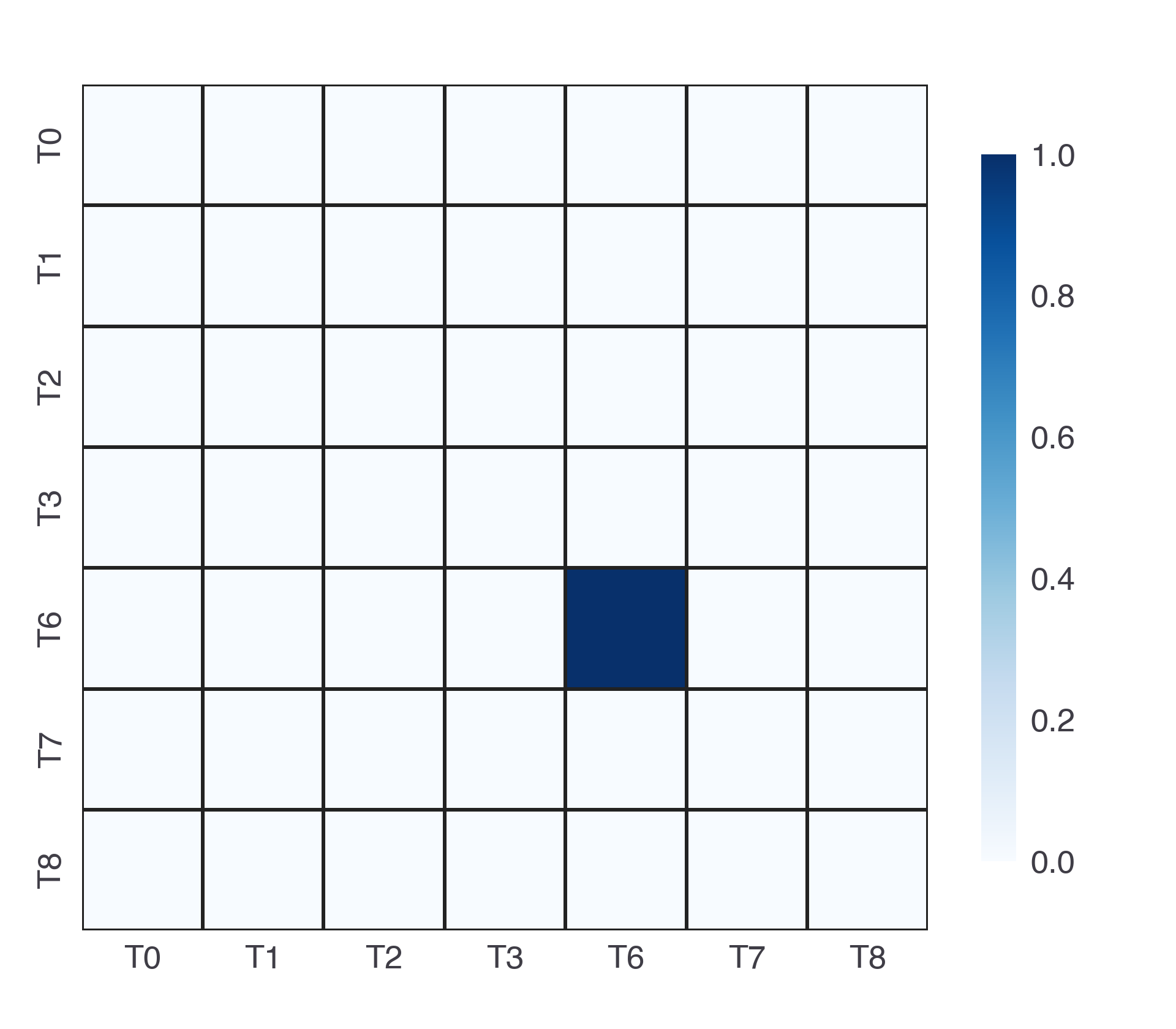} \\ 
DISMOP-TD3-GOAL  
& \includegraphics[align=c,width=.15\textwidth]{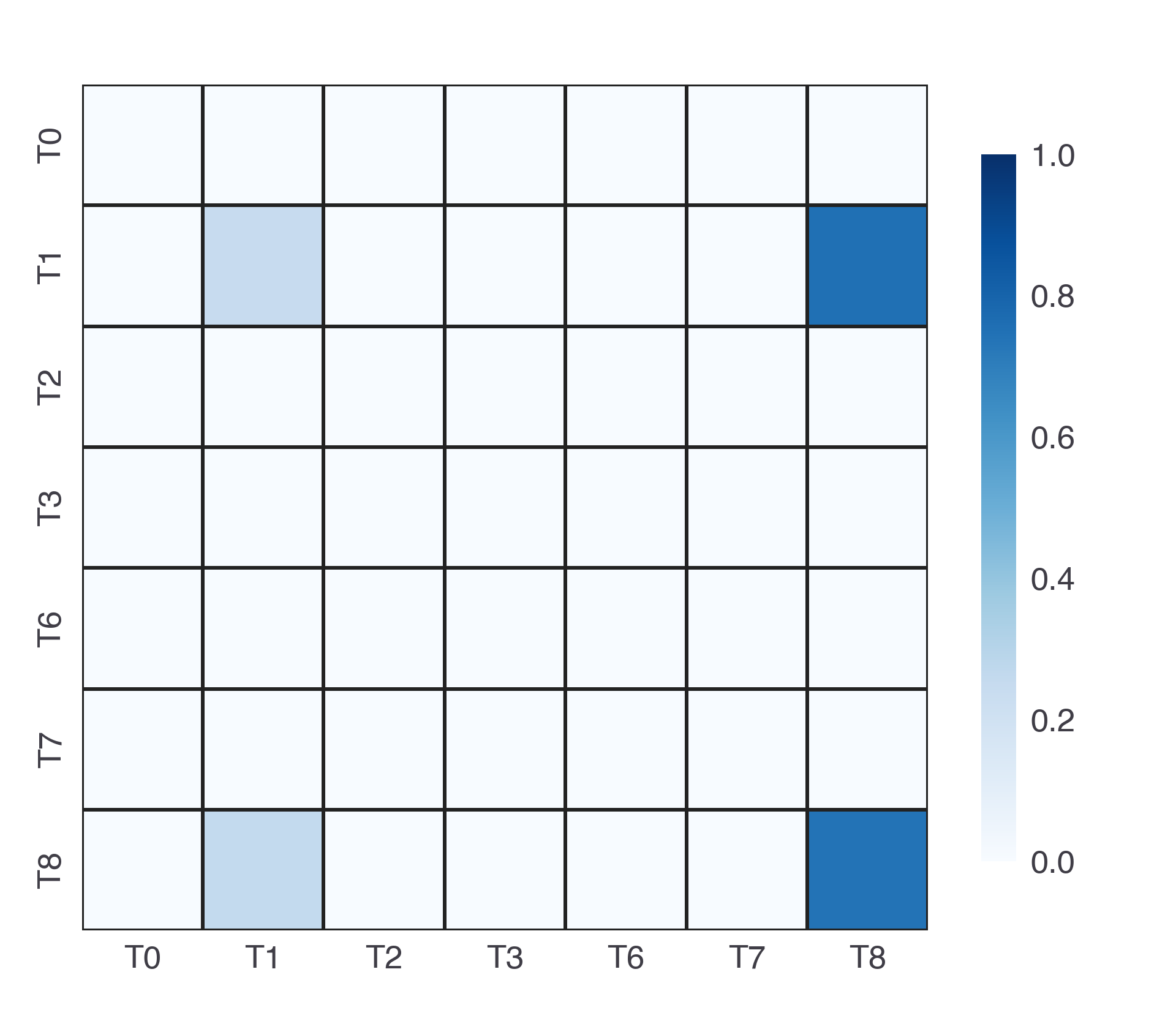}
& \includegraphics[align=c,width=.15\textwidth]{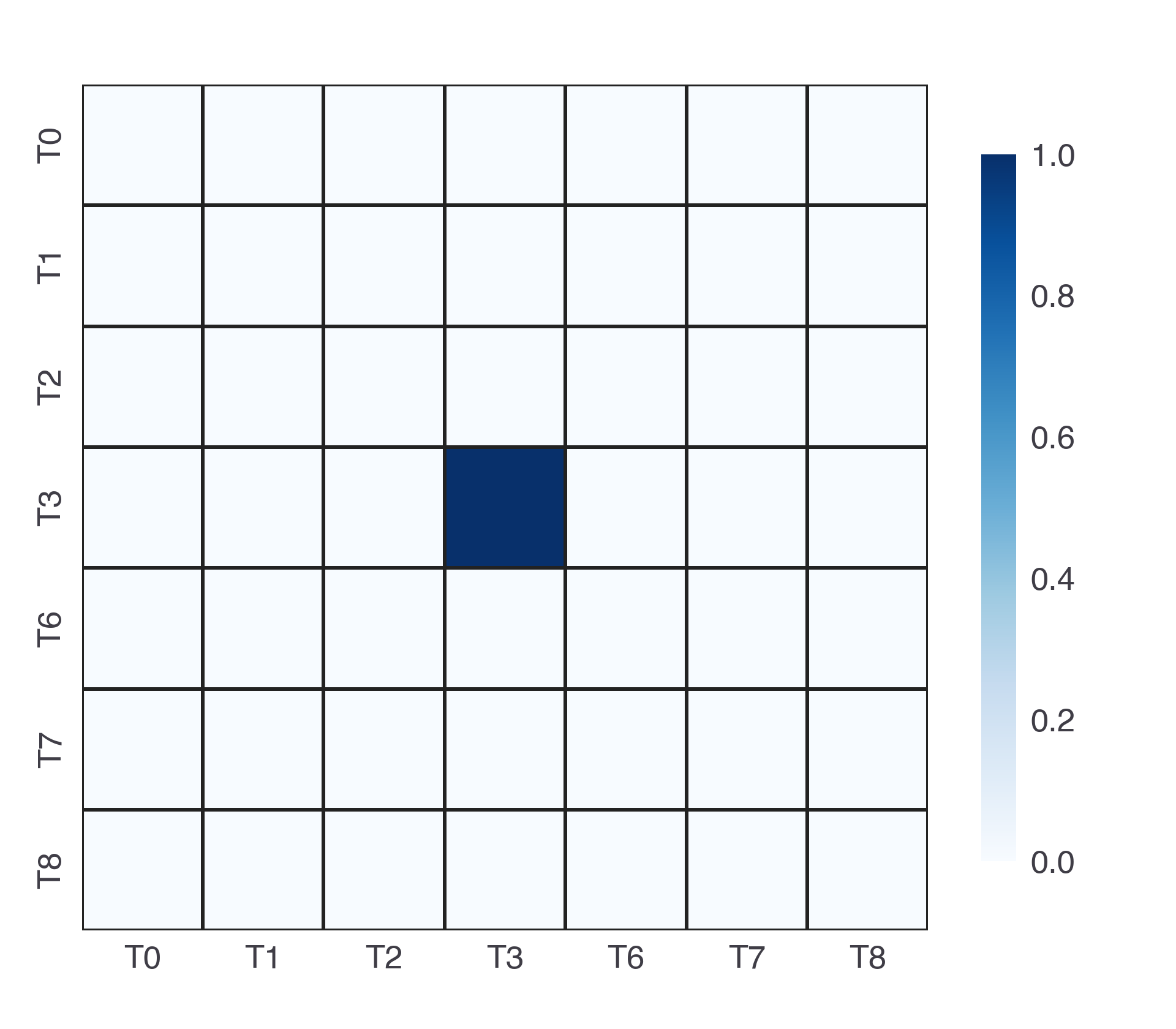}
& \includegraphics[align=c,width=.15\textwidth]{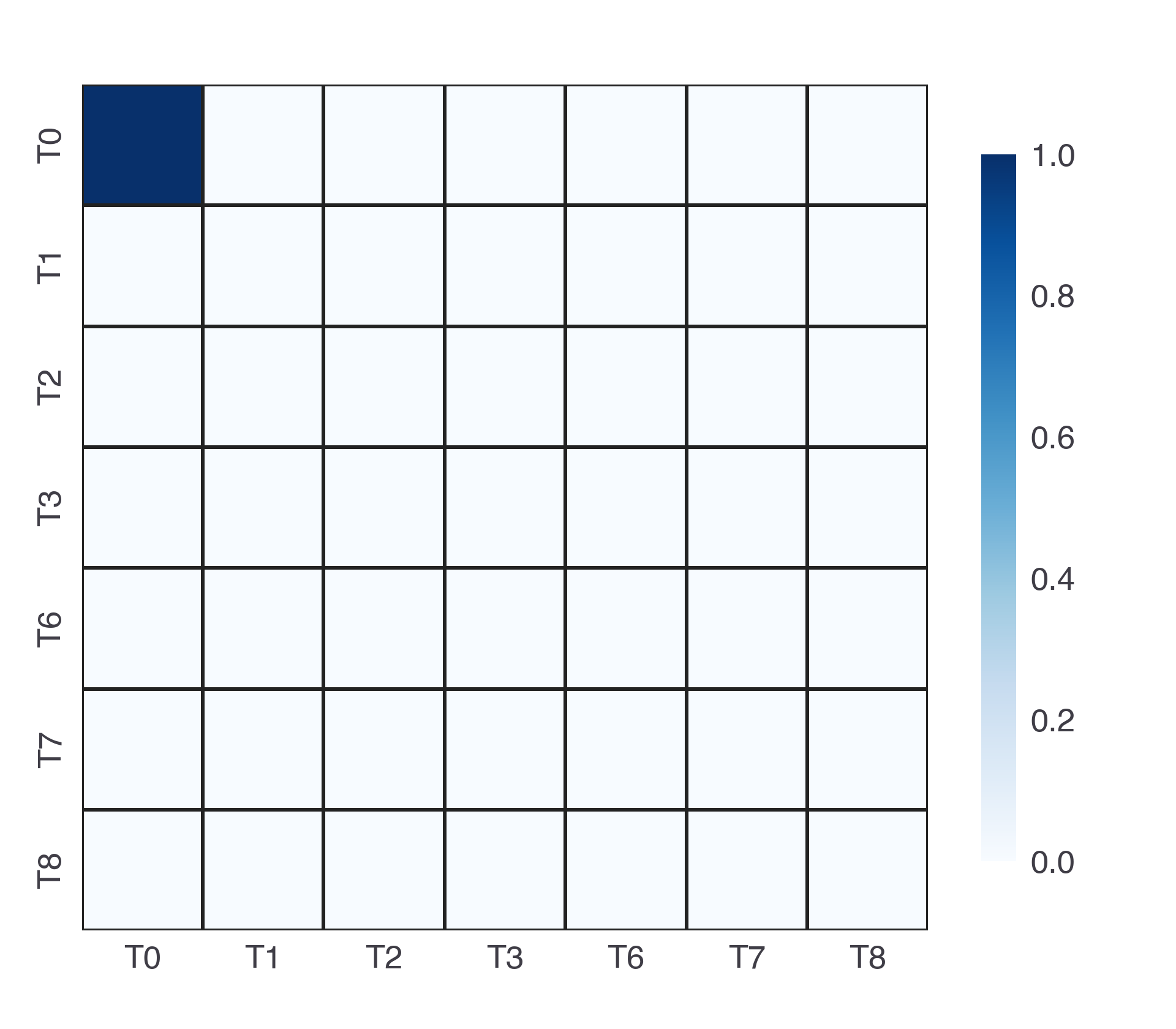}
& \includegraphics[align=c,width=.15\textwidth]{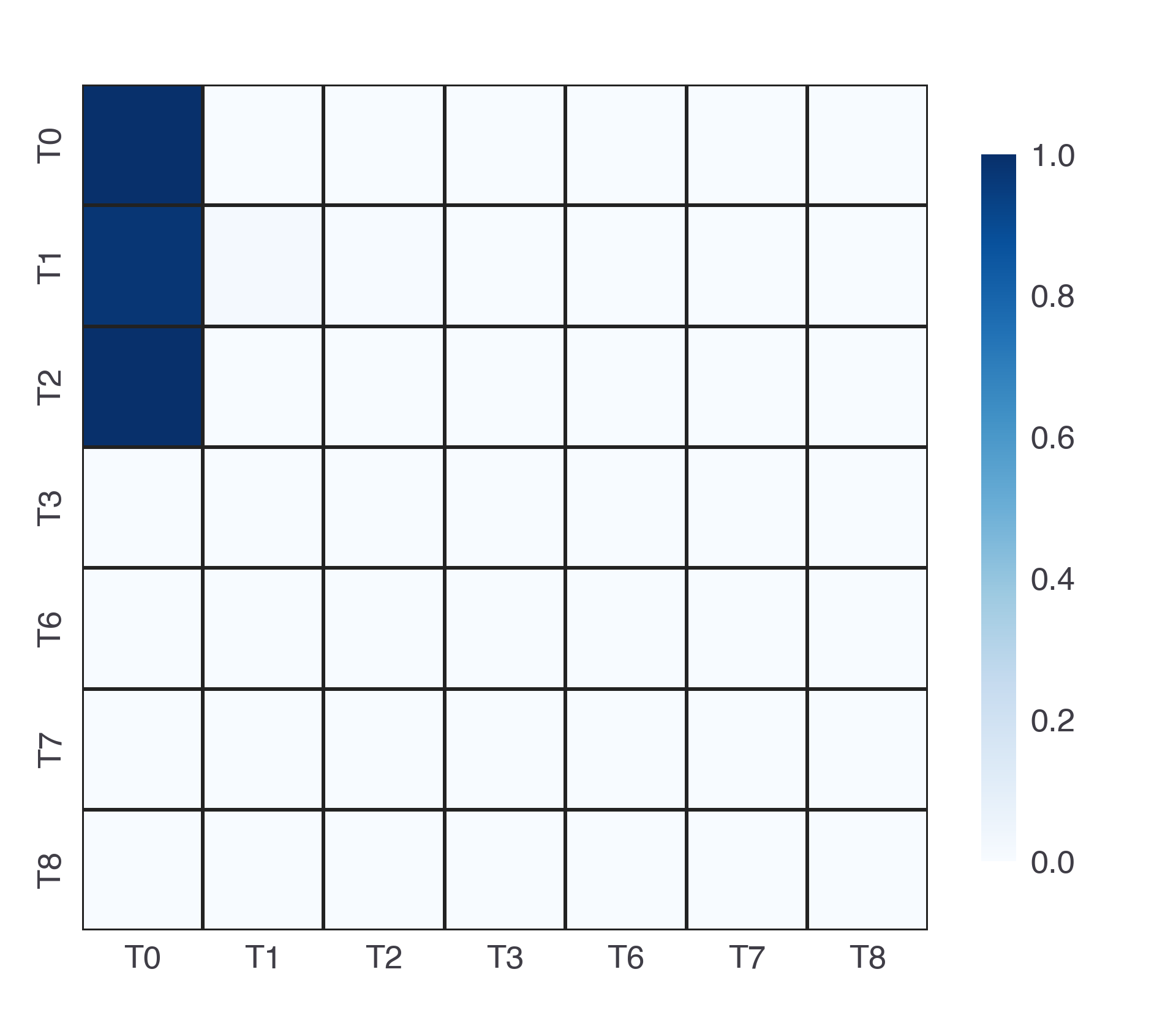}
& \includegraphics[align=c,width=.15\textwidth]{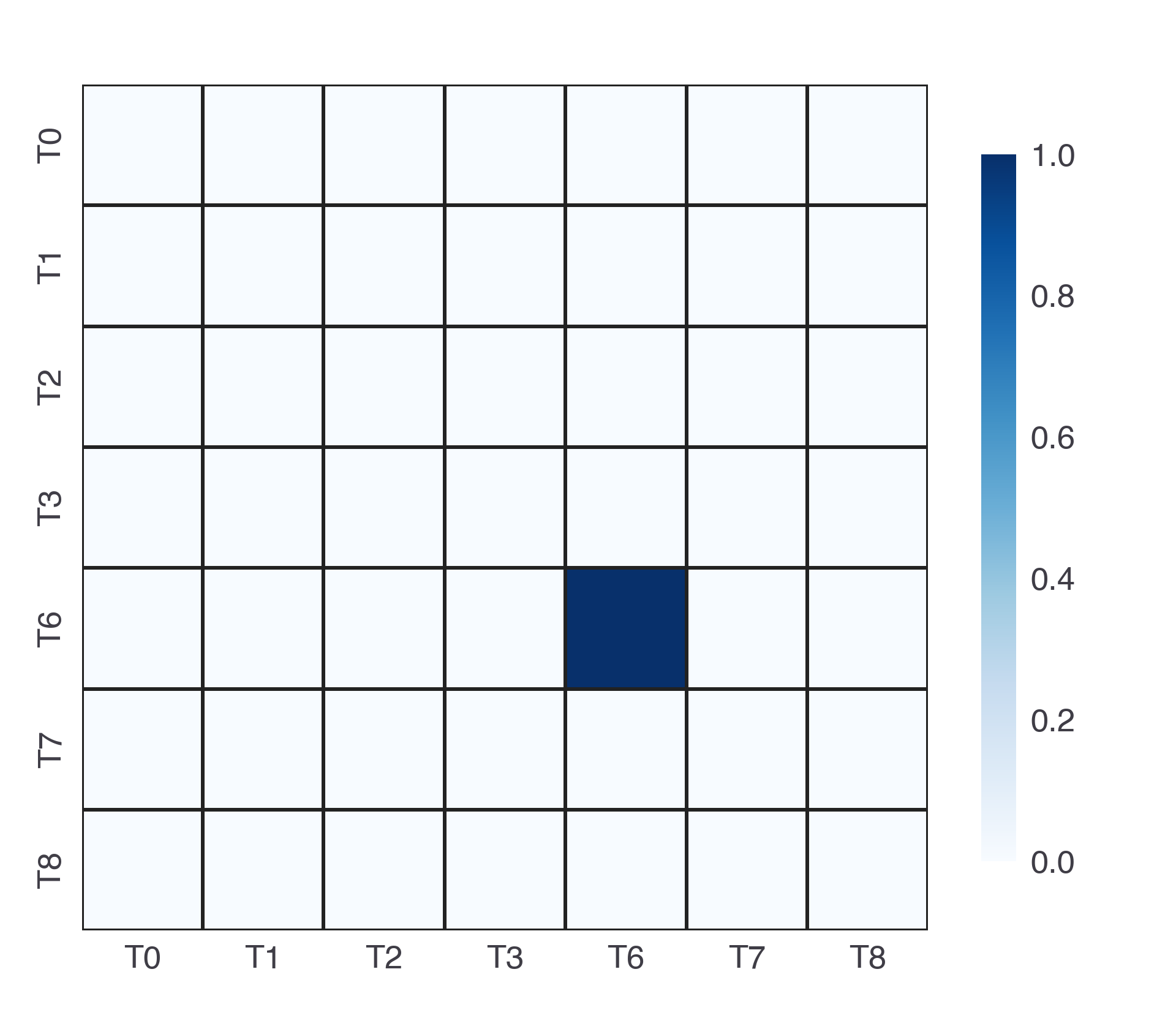}\\ \hline
DISMOP-BCQ-TASK  
& \includegraphics[align=c,width=.15\textwidth]{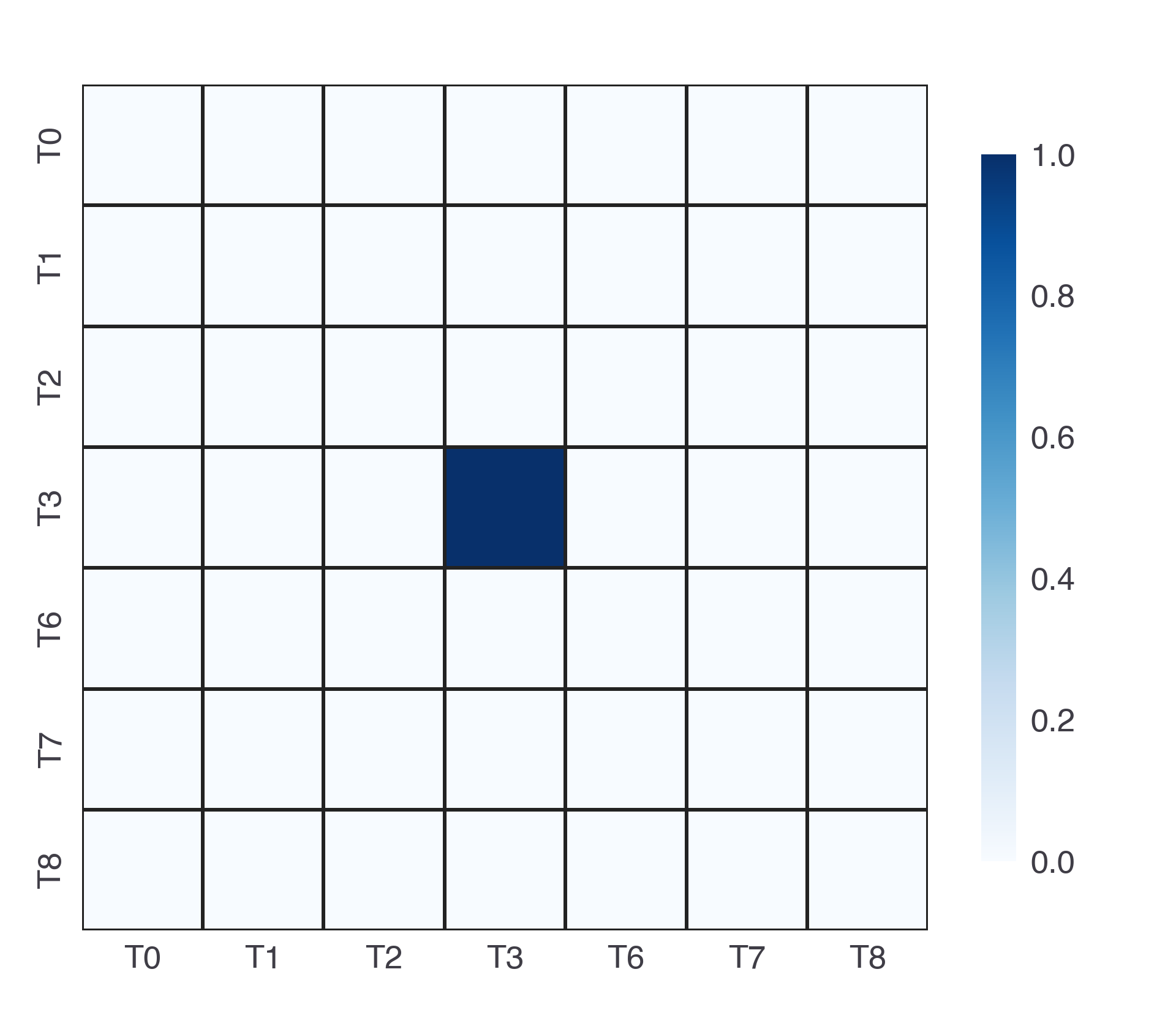}
& \includegraphics[align=c,width=.15\textwidth]{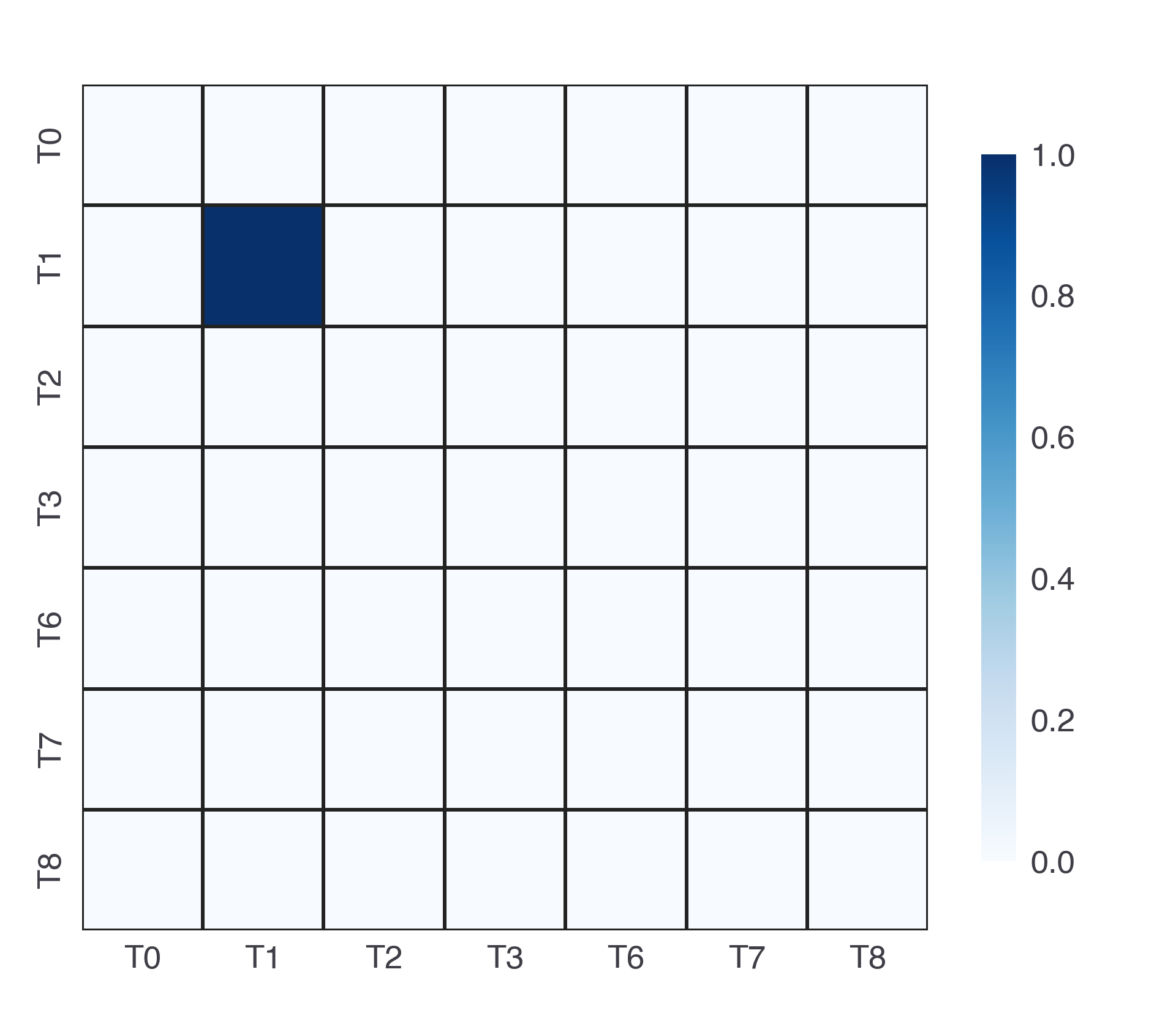}
& \includegraphics[align=c,width=.15\textwidth]{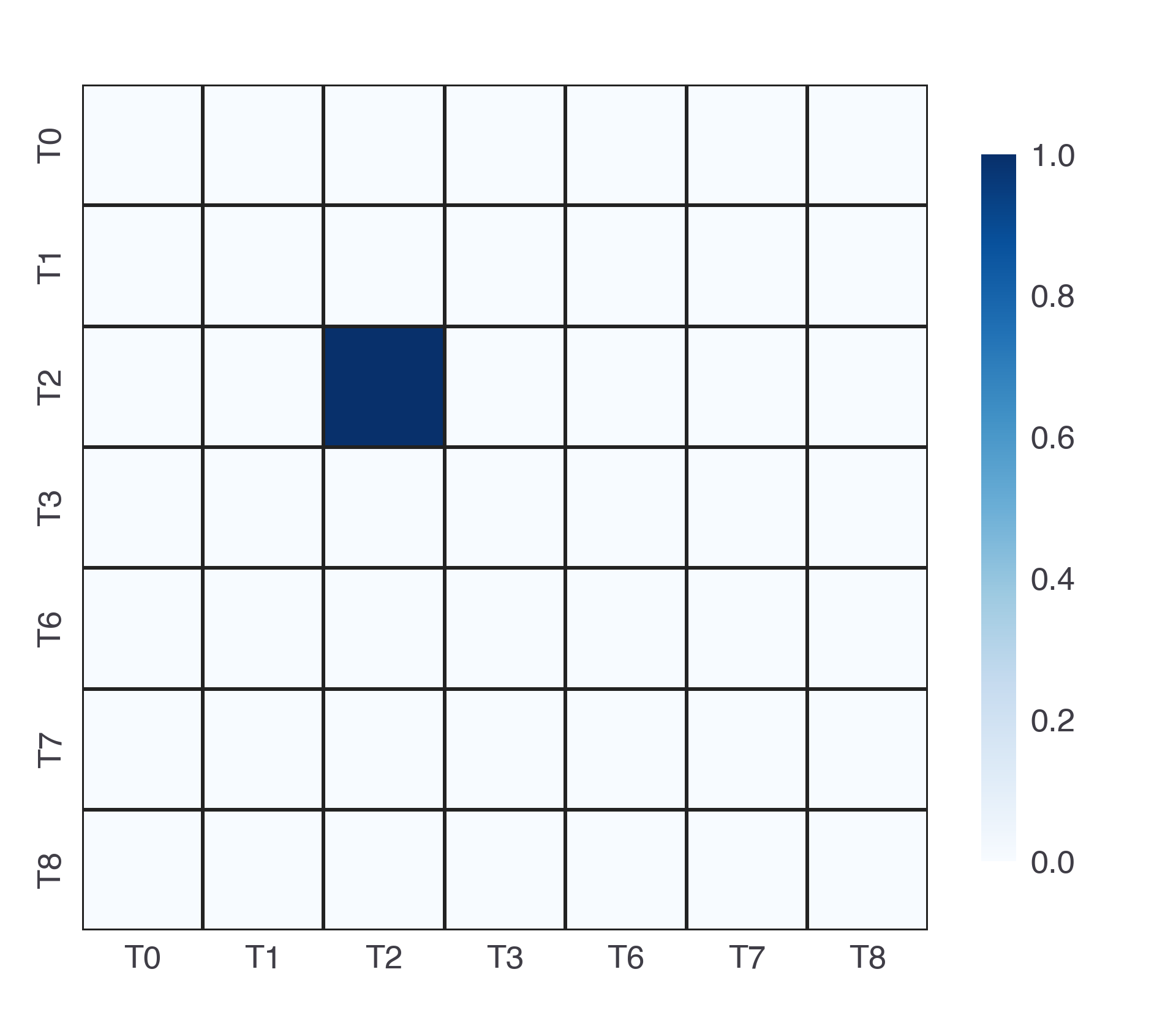}
& \includegraphics[align=c,width=.15\textwidth]{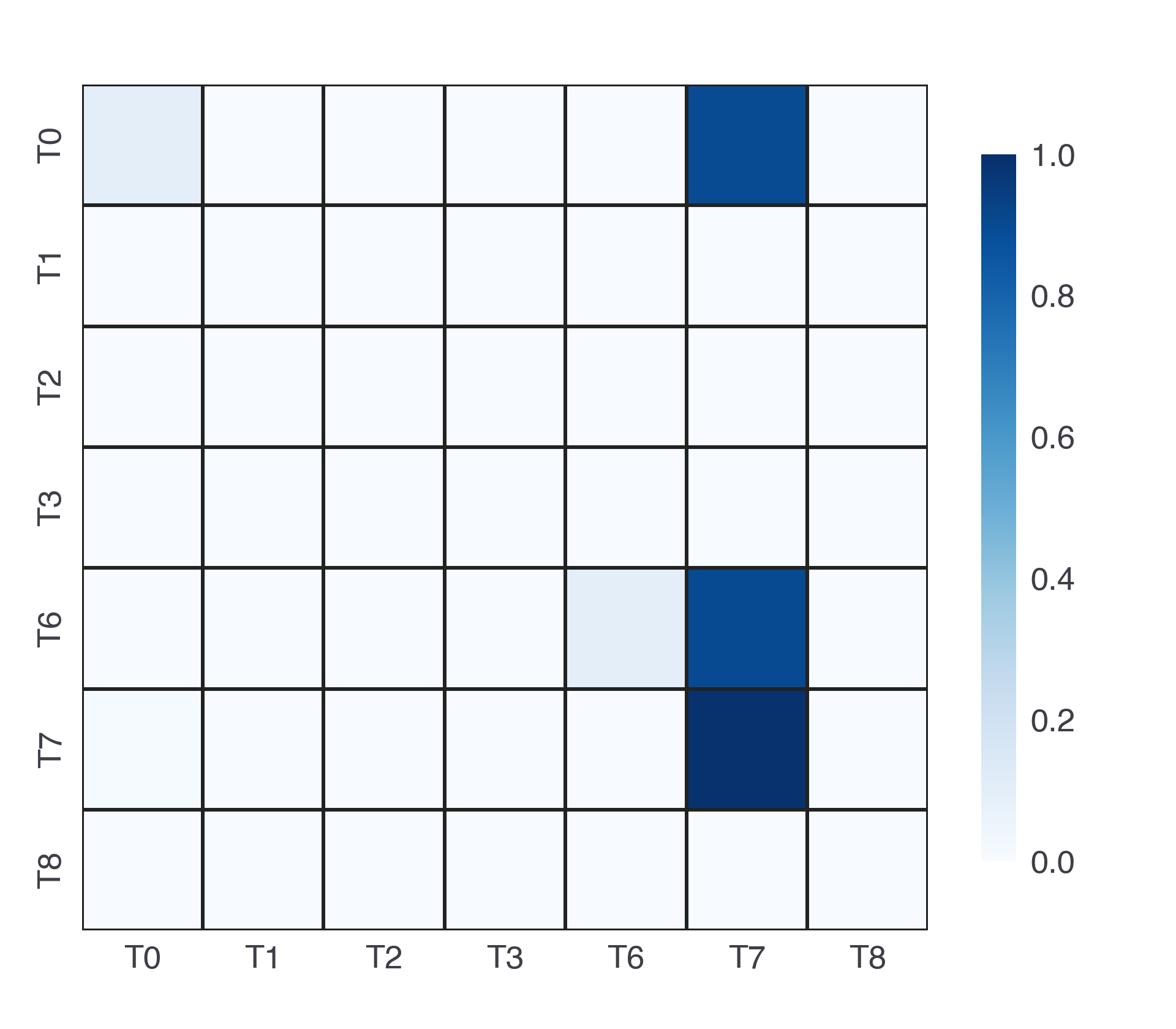}
& \includegraphics[align=c,width=.15\textwidth]{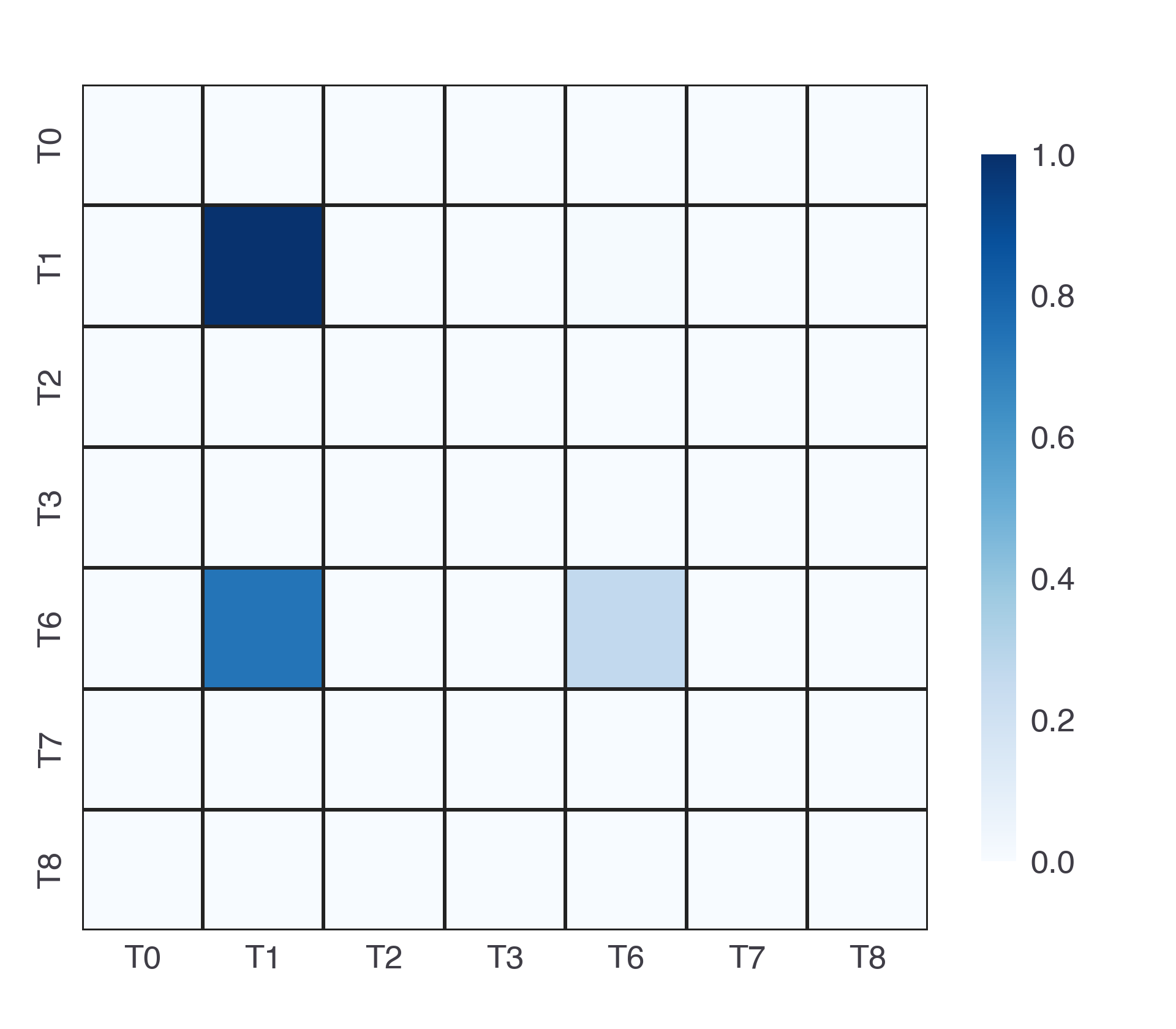} \\ 
DISMOP-BCQ-BOND  
& \includegraphics[align=c,width=.15\textwidth]{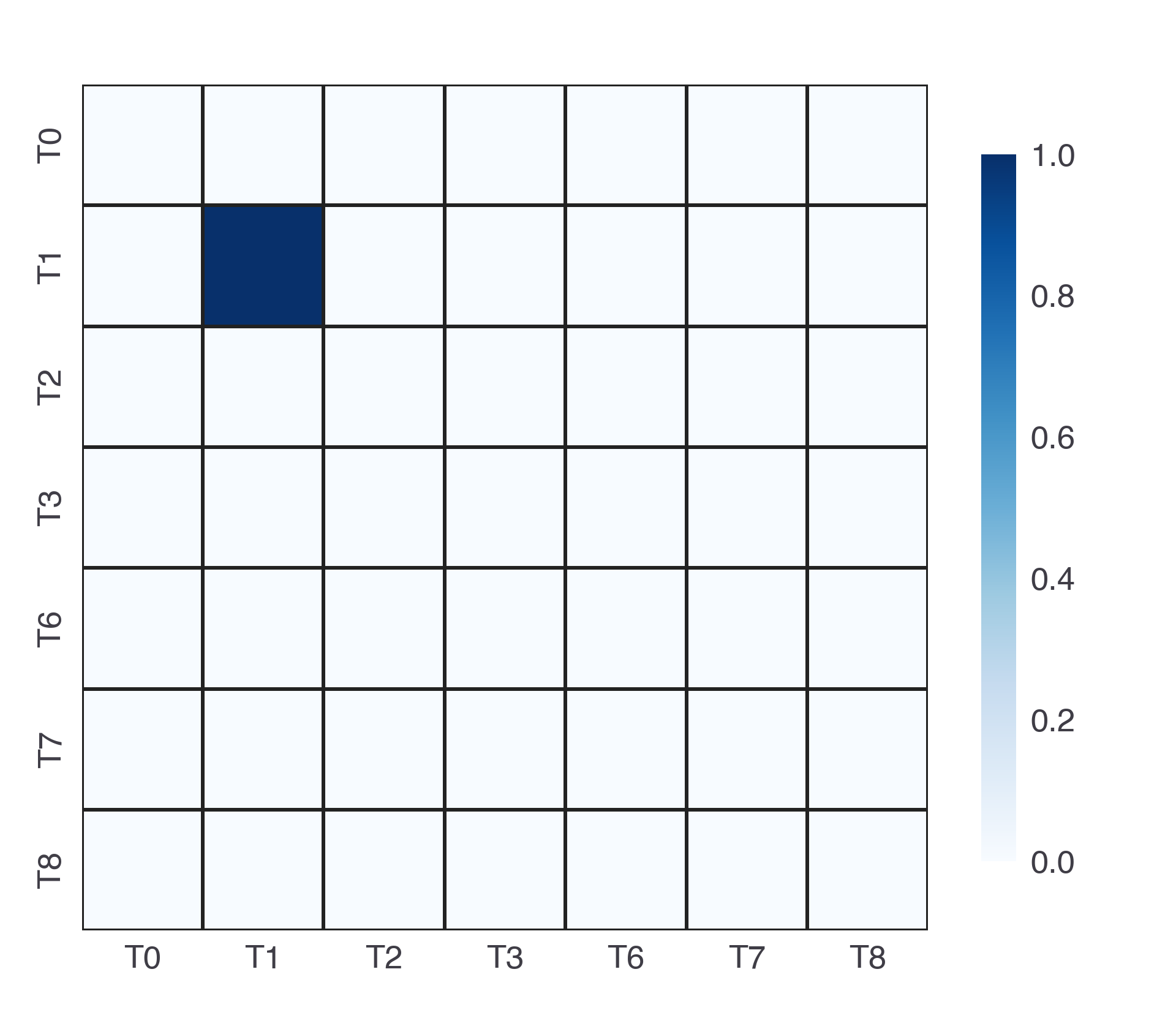}
& \includegraphics[align=c,width=.15\textwidth]{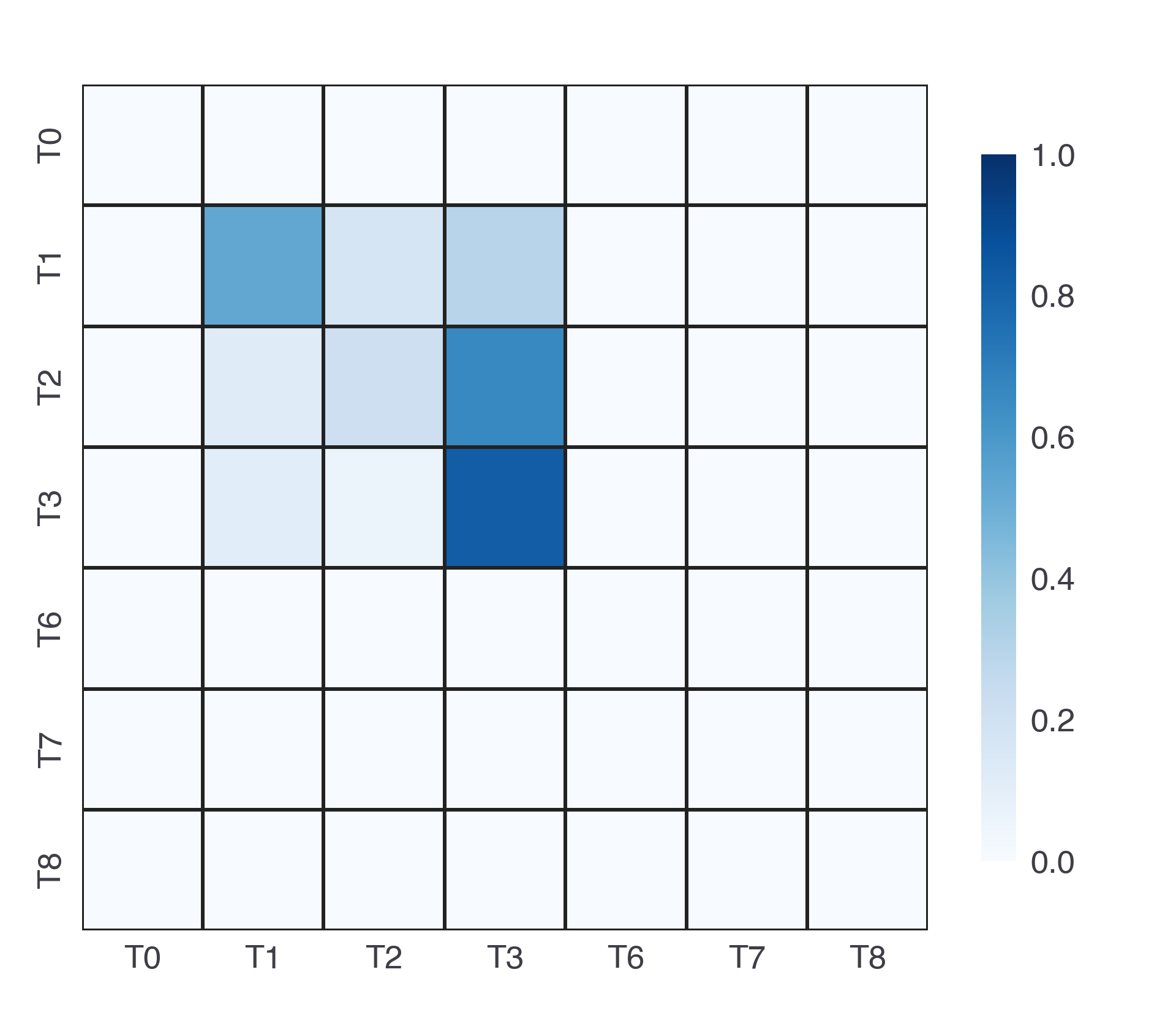}
& \includegraphics[align=c,width=.15\textwidth]{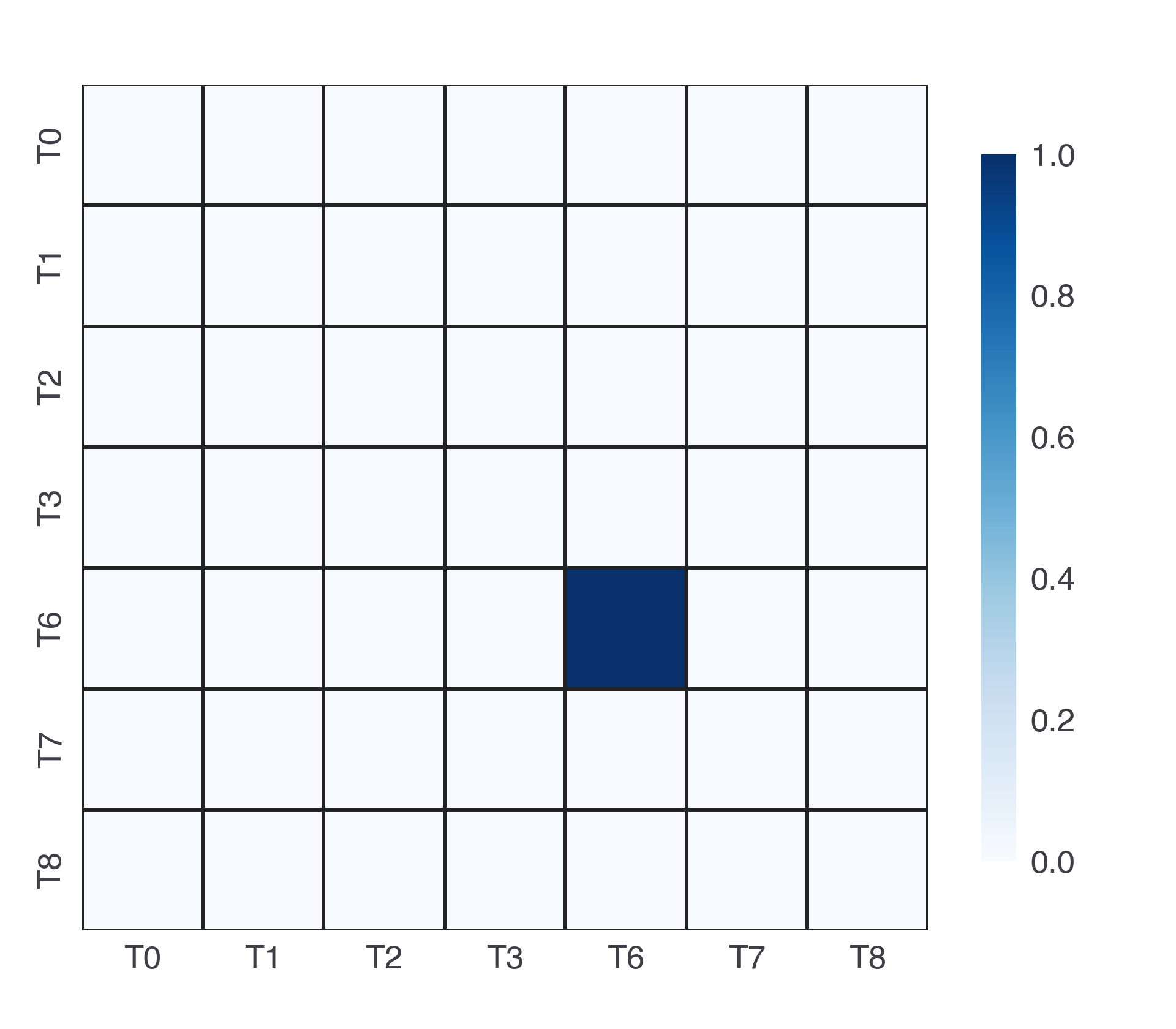}
& \includegraphics[align=c,width=.15\textwidth]{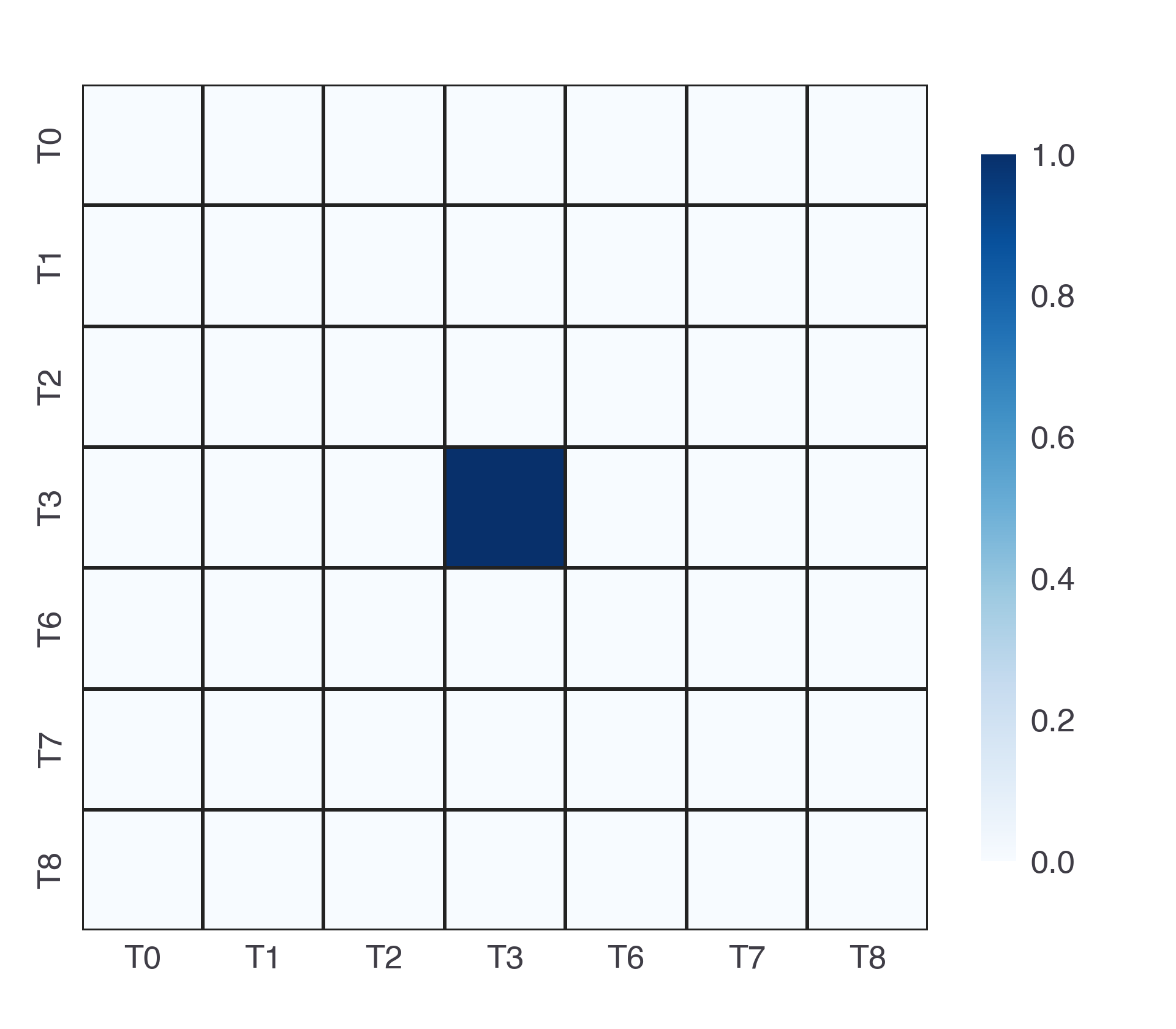}
& \includegraphics[align=c,width=.15\textwidth]{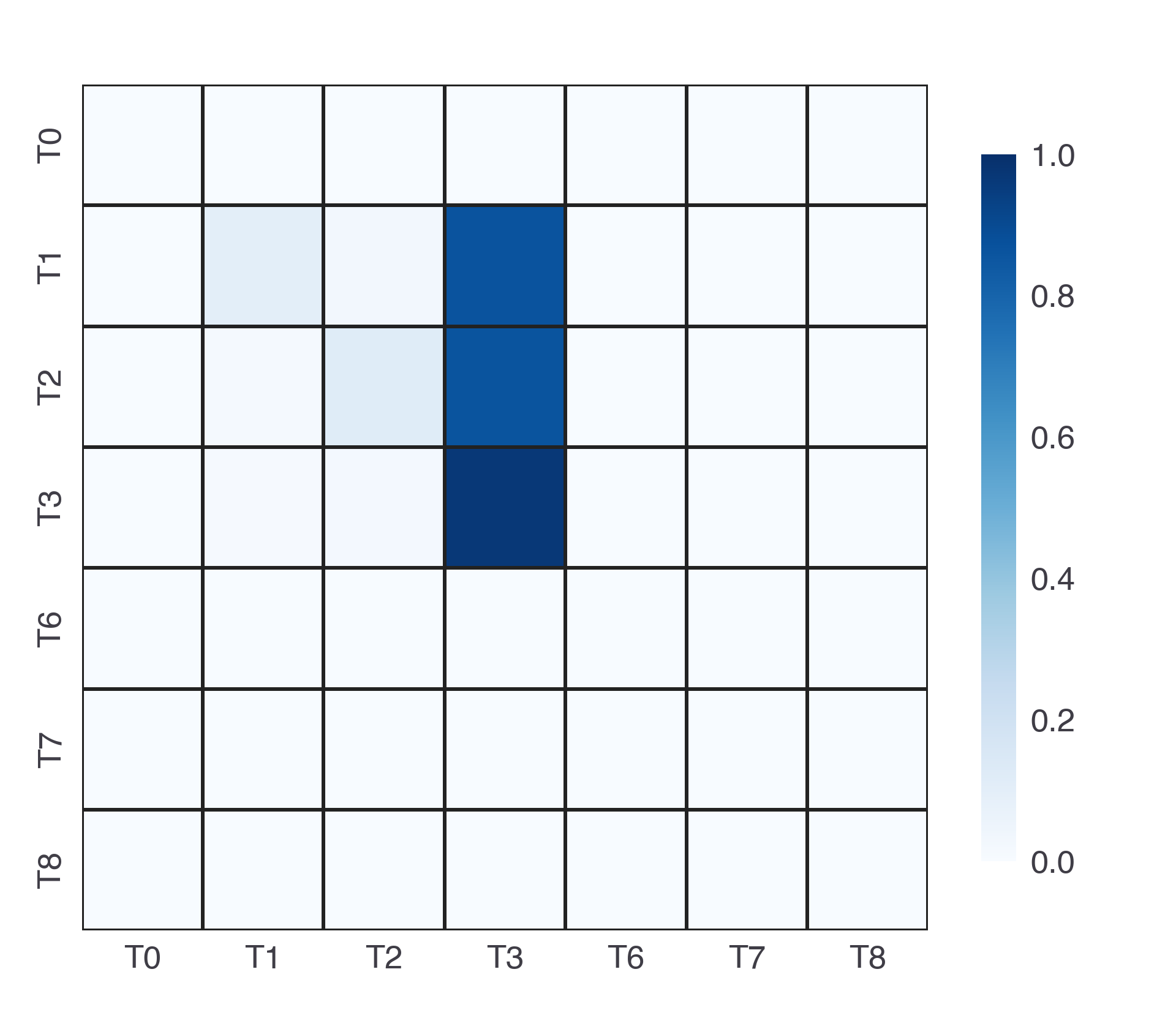} \\ 
DISMOP-BCQ-GOAL 
& \includegraphics[align=c,width=.15\textwidth]{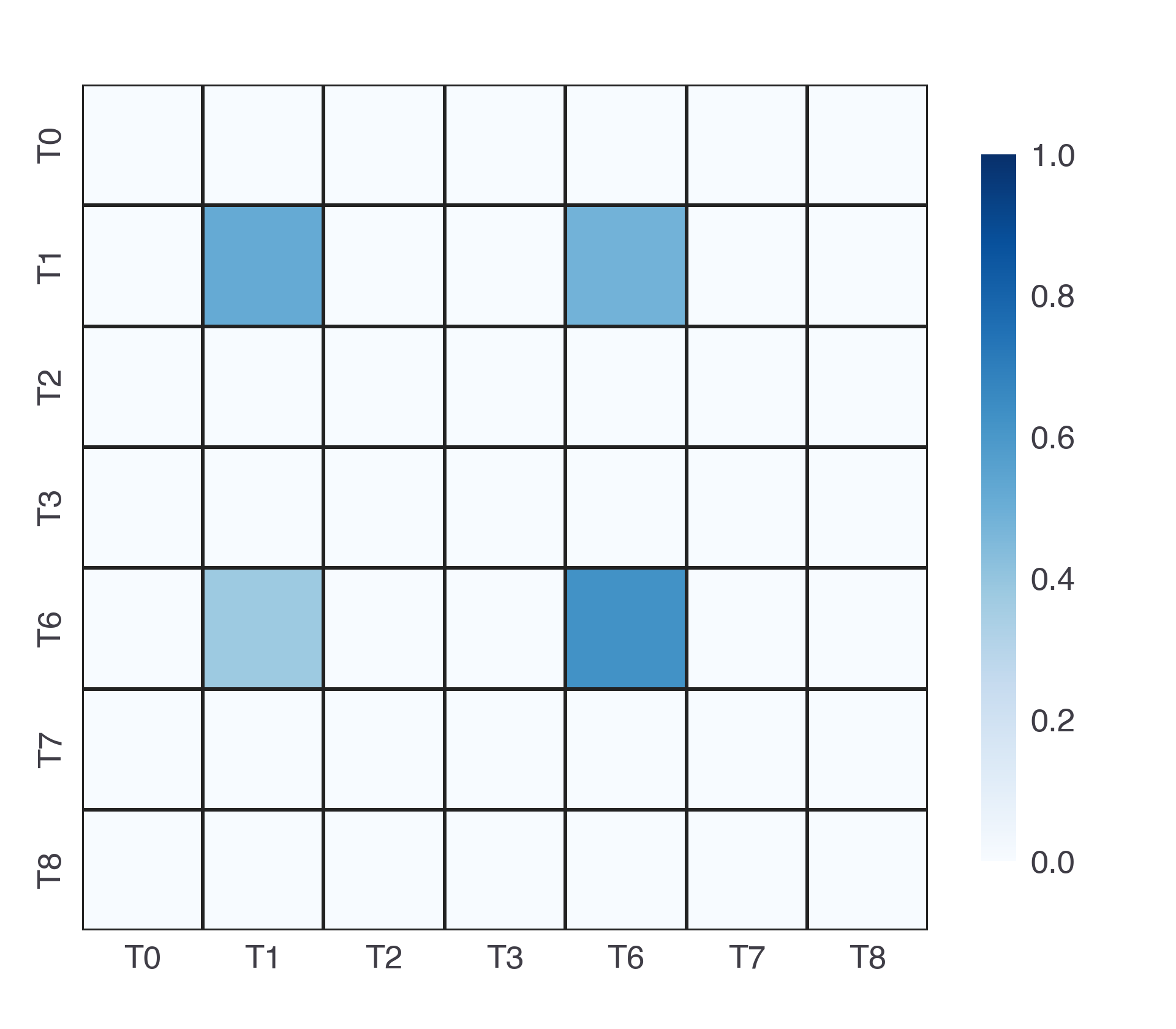}
& \includegraphics[align=c,width=.15\textwidth]{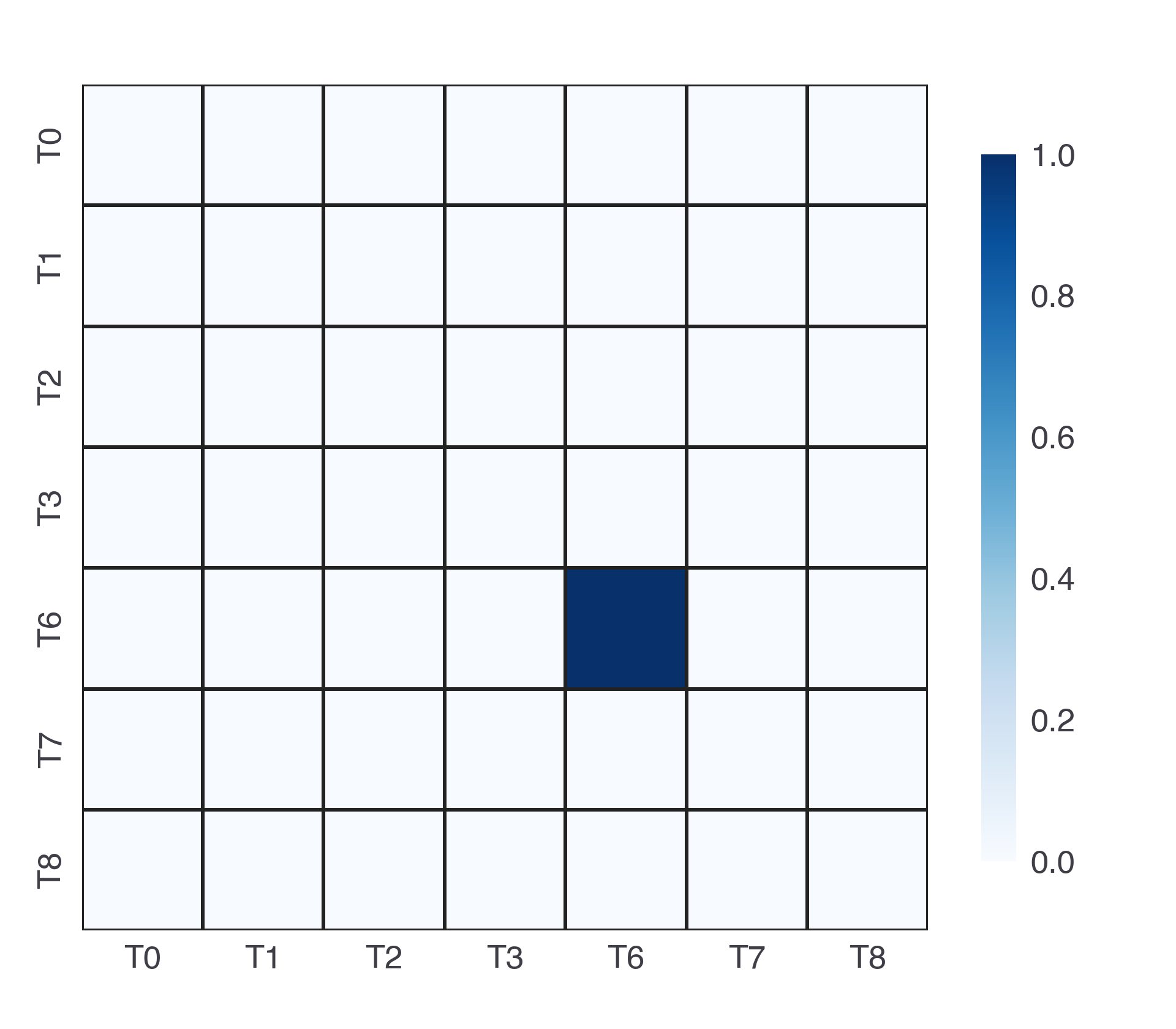}
& \includegraphics[align=c,width=.15\textwidth]{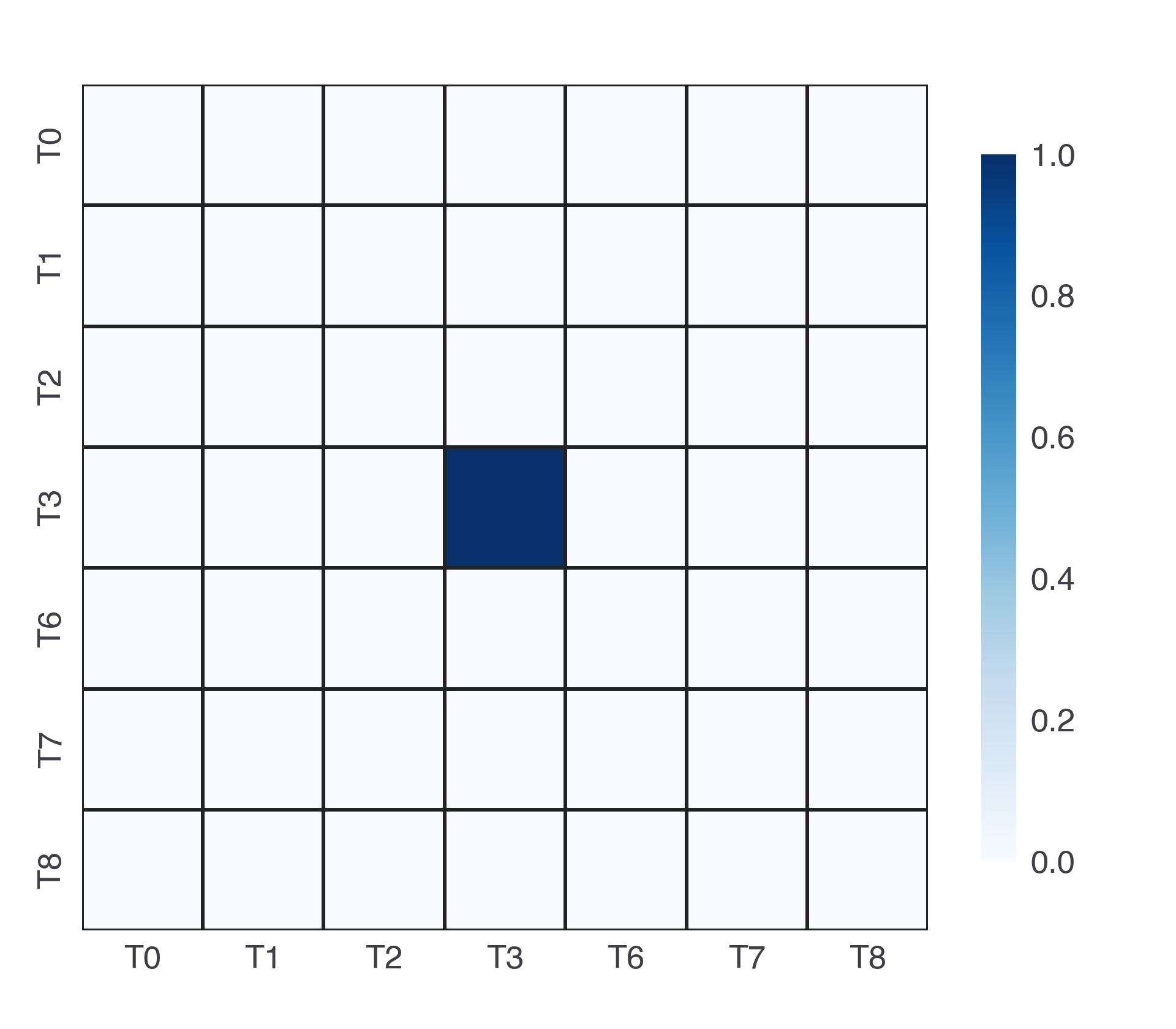}
& \includegraphics[align=c,width=.15\textwidth]{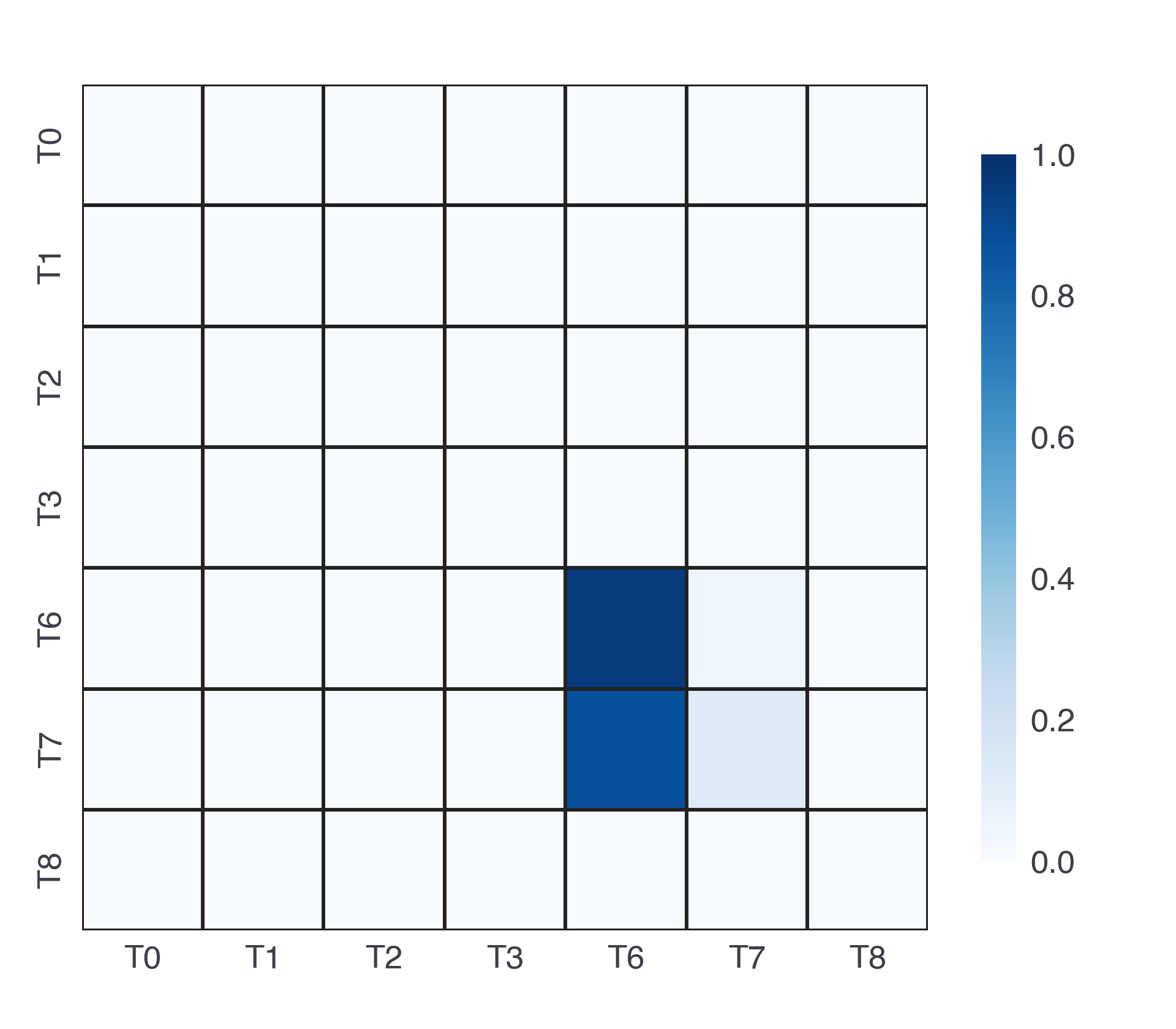}
& \includegraphics[align=c,width=.15\textwidth]{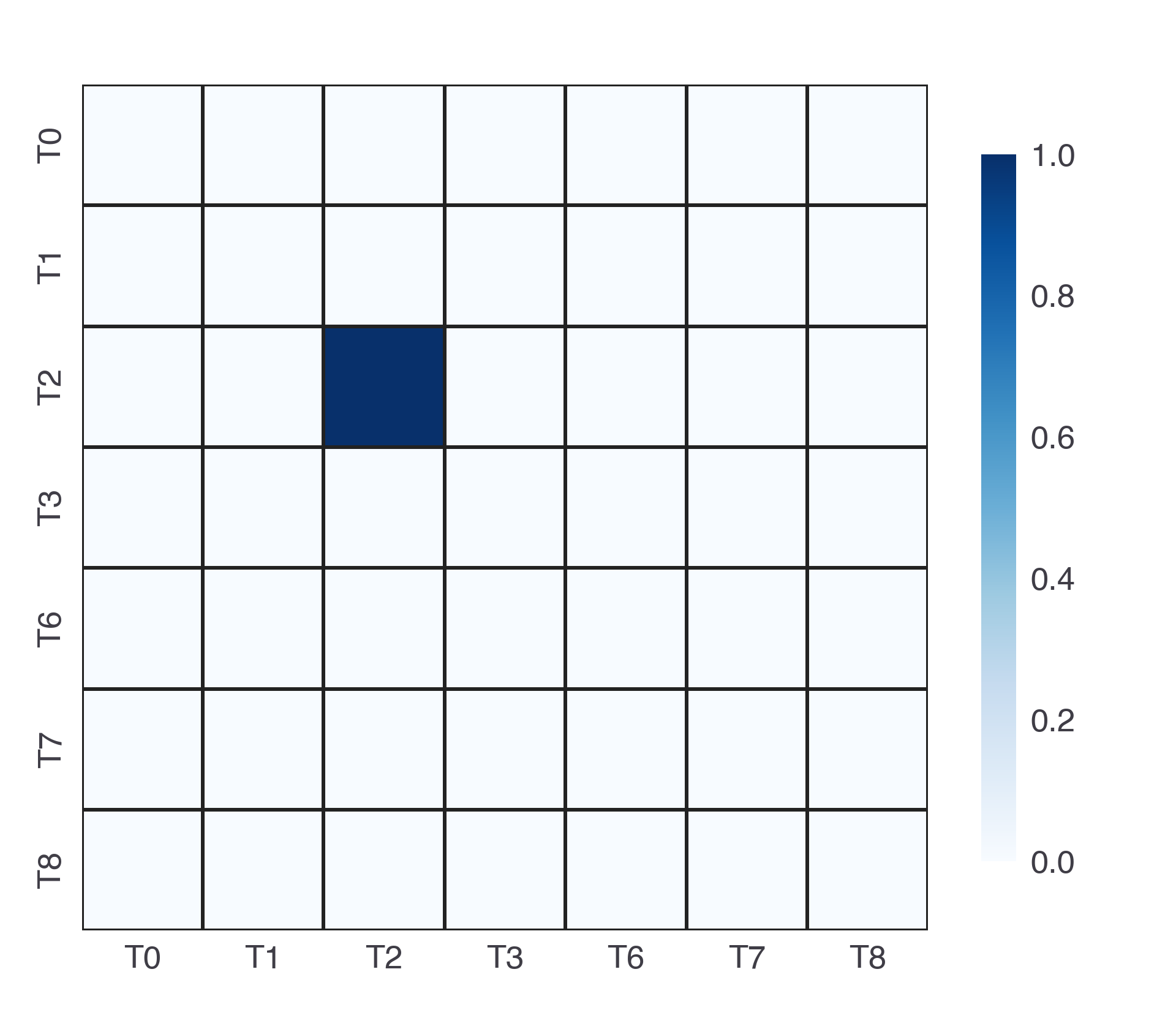} \\ 
\end{tabular}
\end{table*}

\section{Empirical Results}

\subsection{Experimental Setting}

We used the Alex Street psychotherapy dataset \footnote{https://alexanderstreet.com/products/counseling-and-psychotherapy-transcripts-series} to evaluate the recommendation systems. The dataset contains over 950 transcribed therapy sessions between multiple anonymized therapists and patients, covering four psychiatric conditions: anxiety, depression, schizophrenia, and suicidal cases. We preprocessed the dataset into a recommendation system format, which resulted in 219,999 recommendation actions. We split the dataset into 95/5 train-test sets and trained the R2D2-based DISMOP on each of the text corpus, as well as on all four together.

To set up the batch training for reinforcement learning, we cut the turns into frames of 10 turn pairs and used a batch size of 32. We represented the action spaces (the topics to recommend) in three candidate embedding spaces: the averaged 300-dimension Doc2Vec embedding for each topic, the averaged 36-dimension principal component analysis (PCA) embedding, and the averaged 2-dimension Uniform Manifold Approximation and Projection (UMAP) embedding. We presented the results for the first embedding space only. We trained the DISMOP with three reinforcement learning agents (DDPG, TD3, and BCQ), each for 50 epochs, where their losses consistently dropped and converged in a stable way. We did not observe overfitting in any of the model training processes based on the loss curve.

\subsection{Empirical Results}

We evaluated the performance of the three recommendation agents by computing the accuracy of the recommended actions with their corresponding ground truth actions on the test set. We compared the variants of DISMOP, as there were no state-of-the-art or baseline models in this application. We used three different scales of working alliance as our ratings: task, bond, and goal, which measure different aspects of emotional alignments in psychotherapy. Using accuracy to evaluate the recommendation system is a challenging task, as the embedding space can be noisy in our policy generator. Nevertheless, some models using certain therapeutic signals appear to be capturing the real data relatively well.

For instance, DISMOP-BCQ-GOAL (with a test accuracy of 0.6424 for all sessions) and DISMOP-DDPG-TASK (with a test accuracy of 0.6406 for anxiety sessions) were the best-performing models, while others provided trivial solutions. For certain disorders, goal scale and task scale appeared to best capture the human therapists' choices, while other ones favored the models trained with bond scores. For instance, DISMOP-DDPG was the recommender winner for anxiety, while DISMOP-TD3 was the winner for depression and schizophrenia, and DISMOP-BCQ was the winner for schizophrenia and suicidal cases. When pooling the sessions of four disorders together, the recommender winner appeared to be DISMOP-BCQ, which may suggest the offline reinforcement learning's advantage in constraining the possible extrapolation errors by the non-offline methods. This evaluation provides a proof of concept. Future work will focus on systematically comparing a larger spectrum of deep reinforcement learning and model architectures.

\subsection{Interpretable Insights}

We visualized and inspected the policies learned by different policies. Table \ref{tab:traj_viz} presents the standardized average policy trajectories with respect to the action embeddings (marked with topics) projected onto a 2D principal component analysis space. The trajectories are in the length of 10 past actions (as in the frame size). The end of each trajectory is marked with a larger dot. We observed distinct patterns of the policies trained with different reward signals, as well as those trained on sessions with different clinical diagnosis.  

We can further analyze the policies learned by the DISMOPs by inspecting their transition matrices. As shown in Table \ref{tab:policy_viz}, the 1-step transition matrix normalized by rows reveals interesting patterns in the policies. We observe that the transition matrices of the DISMOPs are mostly converged, with different disorders and therapeutic rewards yielding significantly different matrices.

For example, for DISMOP-DDPG trained in depression sessions, if emphasizing the task scale of working alliance, the policy tends to go from talking about sensitive topics like anger, scare, and sadness (topic 2) back to topic 1, which is about play. However, if the policy aims towards the goal scale, it tends to stay focused on discussing anger, scare, and sadness (topic 2). Similarly, for DISMOP-DDPG trained in suicidal sessions, there are recurring discussions about topic 6, which is about explicit ways to deal with stress, such as keeping busy and reaching out for help, which can increase bonding between the doctor and patient and achieve better alliance in their goal scale. However, if the aim is to simply achieve alignment in their tasks during each session, discussing topic 2 is the way to go.

Another interesting example is DISMOP-TD3 trained in schizophrenia patients. We observe that the best topic to achieve the task scale is to continuously discuss topic 6 (dealing with stress), but if the aim is to achieve the bond scale, the focus should be on topic 3 (anger and sadness). If the goal scale is targeted, the policy tends to focus on topic 0 (figuring out and self-discovery).

These insights provide a deeper understanding of the learned policies and how they can be interpreted in the context of psychotherapy. The visualization and interpretation of the DISMOPs' policy dynamics offer valuable insights into the underlying decision-making processes and can help in understanding how the policies are shaped by different disorders and therapeutic rewards. Overall, these vsualization analytics demonstrate that the policies learned by different reinforcement learning agents are distinct and reveal patterns that are consistent with what we know about their underlying therapeutic signals. 

However, it is important to note that these insights are just a starting point for further investigation. While our results suggest that certain policies perform well on the test set and that their trajectories and transition matrices provide interpretable insights, more work is needed to validate these findings and explore their generalizability to other datasets and contexts. Additionally, it is important to consider the ethical implications of using these models in clinical practice and to ensure that they are used responsibly and with proper oversight.

\subsection{Ethical Considerations}

As with any technology used in healthcare, it is essential to consider the ethical implications of our work. One potential concern is the privacy of patient data. We take great care to ensure that all training examples are properly anonymized with pre- and post-processing techniques to protect patient privacy. Additionally, we only use the data for research purposes and do not share it with any third parties. We follow the ethical and operational guidelines 
in \cite{lin2022ethics,matthews2017stories,graham2019artificial}, and ensure that all training examples are properly anonymized with pre- and post-processing techniques. We disclaim that these investigations are proof of concept and require extensive caution to prevent from the pitfall of over-interpretation.

Another concern is the potential for bias in the models. It is well-known that machine learning models can exhibit bias if they are trained on biased data or if the algorithms used to train them are biased themselves. To address this issue, we use a variety of techniques to mitigate bias in our models, including careful selection of training data, use of multiple reinforcement learning algorithms, and regular monitoring of the models' performance on the test set.

We also recognize that the use of machine learning models in psychotherapy raises important ethical and clinical considerations. While these models have the potential to assist clinicians in making recommendations, they should not replace the human judgement and intuition that are critical to effective therapy. It is important to carefully consider the limitations and potential risks of using these models in clinical practice and to ensure that they are used responsibly and ethically. This includes ensuring that patients are fully informed about the use of these models and that their consent is obtained before any data is collected or analyzed.

In conclusion, our study demonstrates the potential of deep reinforcement learning models to assist in making recommendations in psychotherapy. Our results provide a proof of concept that these models can be trained to capture underlying therapeutic signals and provide accurate recommendations. However, further work is needed to validate these findings and explore their generalizability to other datasets and contexts. It is also important to carefully consider the ethical implications of using these models in clinical practice and to ensure that they are used responsibly and with proper oversight.

\section{Conclusions and Future Directions}

In conclusion, we have presented a Psychotherapy AI Companion, a recommendation for psychotherapy topics, and evaluated its performance using reinforcement learning with different reward signals. Our experiments demonstrate the effectiveness of using deep reinforcement learning in this setting, and also provide insights into how different reward signals affect the recommendation policies learned by the system. In addition, we have presented interpretable insights into the recommendation policies through the use of visualizations, such as trajectory and transition matrix plots, and discussed how these insights can inform future work on improving the system and developing new approaches for understanding and interpreting deep reinforcement learning models.

Moving forward, we plan to extend our work in several directions. First, we aim to explore the use of more advanced reinforcement learning algorithms, such as actor-critic methods or proximal policy optimization, and compare their performance to the methods used in this study. We also plan to investigate the use of more sophisticated embeddings, such as contextual embeddings or knowledge graph embeddings, to further improve the quality of the recommendations. In addition, we will explore the use of user feedback to refine the recommendations and personalize them for individual patients.

One of the major contributions of our work is the use of DISMOP to provide interpretable insights into the policies learned by the system. We plan to continue exploring the use of DISMOP in this context, and also investigate the use of other interpretability techniques, such as attention mechanisms or saliency maps. These approaches could help to shed further light on the inner workings of deep reinforcement learning models and provide insights into how they can be improved.

In addition to the three levels of psychotherapy AI companion described in Section \ref{sec:threelevels} and Figure \ref{fig:pipeline1}, our DISMOP approach can also be extended to incorporate interpretable policies that enable more transparent and ethical decision-making. For example, by visualizing the learned policies and analyzing the transition matrices of the DISMOPs, we gain insights into the decision-making process of the AI companion, and thus, provide more safeguarding against potential bias and stereotypes. These insights can provide valuable information to clinicians and researchers for improving the quality of care and advancing the field of psychotherapy in a responsible and safe way. Furthermore, our approach can be extended to incorporate natural language generation capabilities to enable the AI companion to generate responses to patients in real-time, providing more timely and personalized care. Finally, we will also investigate the integration of other types of data, such as physiological signals and behavioral data, to improve the accuracy and effectiveness of our recommendation systems. 

Finally, we aim to continue exploring the universal ethical implications of using recommendation systems in psychotherapy and ensure that our work adheres to ethical guidelines and best practices. This includes addressing concerns around patient privacy and informed consent, as well as considering the potential impact of the recommendations on patients' mental health and well-being. By continuing to develop and refine our approach in a responsible and ethical manner, we hope to contribute to the growing body of work aimed at improving mental health care through the use of artificial intelligence and machine learning.

\clearpage
\bibliographystyle{ACM-Reference-Format}
\bibliography{main}










\end{document}